\documentclass{article}

\usepackage{microtype}
\PassOptionsToPackage{hyphens}{url}
\usepackage{hyperref}

\usepackage[preprint]{icml2027}
\usepackage[T1]{fontenc}

\usepackage{amsmath,amsfonts,bm}

\def\eqref#1{equation~\ref{#1}}
\def\1{\bm{1}}

\DeclareMathAlphabet{\mathsfit}{\encodingdefault}{\sfdefault}{m}{sl}
\SetMathAlphabet{\mathsfit}{bold}{\encodingdefault}{\sfdefault}{bx}{n}

\usepackage{acro}
\usepackage{graphicx}
\usepackage{booktabs}
\usepackage{colortbl}
\usepackage{multirow}

\newcommand{\greytableleaders}{{\color{black!35}\leaders\hrule height 0.3pt\hfill}}
\newcommand{\dottedtableleaders}{{\color{black!50}\cleaders\hbox to 2.4pt{\hss\rule{0.5pt}{0.5pt}\hss}\hfill}}
\newcommand{\greytablerule}[1]{%
  \noalign{\vskip1pt}%
  \multispan{#1}\greytableleaders\cr
  \noalign{\vskip1.5pt}}
\newcommand{\greytableruleafterfirst}[1]{%
  \noalign{\vskip1pt}%
  \omit&\multispan{#1}\greytableleaders\cr
  \noalign{\vskip1.5pt}}
\usepackage{siunitx}
\usepackage{tikz}
\usepackage{listings}
\definecolor{citationnavy}{HTML}{385673}
\usepackage{tcolorbox}
\tcbuselibrary{skins}

\newtcolorbox{policyinstructionbox}{%
    enhanced,
    width=\linewidth,
    colback=citationnavy!3!white,
    colbacklower=white,
    colframe=citationnavy!45!white,
    coltext=black,
    boxrule=0.5pt,
    arc=1.5pt,
    boxsep=0pt,
    left=10pt,
    right=10pt,
    top=8pt,
    bottom=9pt,
    middle=8pt,
    before skip=0pt,
    after skip=0pt,
    fontupper=\small,
    fontlower=\small,
    before upper={\raggedright\setlength{\parindent}{0pt}\setlength{\parskip}{0.4em}},
    before lower={\raggedright\setlength{\parindent}{0pt}\setlength{\parskip}{0.4em}},
    segmentation style={solid, draw=citationnavy!25!white, line width=0.4pt}
}

\newcommand{\instructionheading}[1]{%
    {\sffamily\bfseries\textcolor{citationnavy}{#1}}\par
}

\definecolor{policybackground}{HTML}{F6F8FA}
\definecolor{policycomment}{HTML}{5C6570}
\definecolor{policyrule}{HTML}{D7DFE6}
\lstdefinestyle{generatedpolicy}{
    language=Python,
    basicstyle=\fontencoding{T1}\fontfamily{lmtt}\fontsize{7}{7.6}\selectfont,
    keywordstyle=\color{citationnavy}\bfseries,
    commentstyle=\color{policycomment},
    stringstyle=\color{citationnavy},
    backgroundcolor=\color{policybackground},
    rulecolor=\color{policyrule},
    frame=single,
    framerule=0.3pt,
    framesep=5pt,
    numbers=left,
    numberstyle=\fontencoding{T1}\fontfamily{lmtt}\fontsize{5}{6}\selectfont\color{policycomment},
    numbersep=9pt,
    xleftmargin=25pt,
    xrightmargin=5pt,
    breaklines=true,
    breakatwhitespace=false,
    breakautoindent=true,
    breakindent=12pt,
    postbreak=\mbox{\textcolor{policycomment}{\ensuremath{\hookrightarrow}}\space},
    columns=fullflexible,
    keepspaces=true,
    showstringspaces=false,
    tabsize=4,
    captionpos=b,
    aboveskip=10pt,
    belowskip=6pt
}
\lstdefinestyle{generatedsnippet}{
    style=generatedpolicy,
    basicstyle=\fontencoding{T1}\fontfamily{lmtt}\fontsize{8}{9}\selectfont,
    numberstyle=\fontencoding{T1}\fontfamily{lmtt}\fontsize{6}{7}\selectfont\color{policycomment}
}

\usepackage{url}
\DeclareAcronym{cc}{
  short = CC,
  long = cyclomatic complexity
}

\DeclareAcronym{hv}{
  short = H.V.,
  long = Halstead volume
}

\DeclareAcronym{loc}{
  short = LOC,
  long = lines of code
}

\DeclareAcronym{gpu}{
  short = GPU,
  long = graphics processing unit
}

\DeclareAcronym{api}{
  short = API,
  long = application programming interface,
  short-indefinite = an,
  long-indefinite = an
}

\DeclareAcronym{rl}{
  short = RL,
  long = reinforcement learning
}

\DeclareAcronym{marl}{
  short = MARL,
  long = multi-agent reinforcement learning
}

\DeclareAcronym{zsc}{
  short = ZSC,
  long = zero-shot coordination
}

\DeclareAcronym{sp}{
  short = SP,
  long = self-play
}

\DeclareAcronym{xp}{
  short = XP,
  long = cross-play
}

\DeclareAcronym{llm}{
  short = LLM,
  long = large language model,
  short-indefinite = an
}

\DeclareAcronym{ixp}{
  short = IXP,
  long = inter-seed cross-play
}

\DeclareAcronym{dec-pomdp}{
  short = Dec-POMDP,
  long = decentralised partially observable Markov decision process,
  long-plural-form = decentralised partially observable Markov decision processes
}

\DeclareAcronym{aoh}{
  short = AOH,
  long = action-observation history,
  long-plural-form = action-observation histories,
  short-indefinite = an,
  long-indefinite = an
}

\DeclareAcronym{lstm}{
  short = LSTM,
  long = long short-term memory network,
  short-indefinite = an
}

\ifdefined\ispreprint
    \makeatletter
    \renewcommand{\icmlcorrespondingauthor}[2]{\gdef\icmlcorrespondingauthor@text{#2}}
    \makeatother
\fi

\begin{document}

\twocolumn[
    \icmltitle{Coding Agents Are Zero-Shot Coordinators}
    \begin{icmlauthorlist}
        \icmlauthor{Shashank Reddy Chirra}{oxford}
        \icmlauthor{Ryo Nakahashi}{sony}
        \icmlauthor{Shimon Whiteson}{oxford}
        \icmlauthor{Jakob Foerster}{oxford}
    \end{icmlauthorlist}
    \icmlaffiliation{oxford}{BOLD, University of Oxford}
    \icmlaffiliation{sony}{Sony Interactive Entertainment}
    \icmlcorrespondingauthor{Shashank Reddy Chirra}{shashank@robots.ox.ac.uk}

    \vskip 0.3in
]
\printAffiliationsAndNotice{}

\begin{abstract}
\Ac{zsc} requires independently developed agents to cooperate with unseen partners, including humans. Current methods make progress towards this goal, but rely on restrictive assumptions. On benchmarks such as Overcooked V2 and Yokai, they are either difficult to apply or, even when applicable, perform poorly. This led to an expectation that further progress requires algorithmic innovation. Surprisingly, we find that these challenges are addressed with an increase in the general capability of \acp{llm}. Coding agents, which pair \iac{llm} with an off-the-shelf coding harness, achieve state-of-the-art results on both benchmarks by generating policies as code and refining them using only \acl*{sp} feedback. Further, \acs{zsc} aims to enable coordination with humans, yet its standard formulation overlooks settings where humans pre-agree on conventions without joint training (e.g, Hanabi). Towards this, we extend \acs{zsc} by providing agents with shared conventions in natural language. In Hanabi, given conventions used by human players, coding agents outperform baselines in \acl*{xp} with the Will-Hanabi bot. Finally, we use the generated policies to guide convention selection during \acl*{sp}, achieving the highest \acl*{xp} with both the Will-Hanabi and the AH2AC2 bots trained on human gameplay logs. Code is available at \href{https://shshnkreddy.github.io/coding-agent-zsc}{\nolinkurl{https://shshnkreddy.github.io/coding-agent-zsc}}.

\end{abstract}

\acresetall

\newcommand{\novelLayoutRef}{Appendix~\ref{app:novel-ocv2-layouts}}
\newcommand{\mainDiagramWidth}{0.8\linewidth}
\newcommand{\mainCrossModelPlotWidth}{\mainDiagramWidth}
\begin{figure*}[t]
    \input{Figures/zsc}
\end{figure*}
\section{Introduction}

Building agents that can robustly cooperate with novel partners, particularly humans, is a central challenge in cooperative \ac{marl} \citep{dafoe2020open,hu2020otherplay}, with important implications for real-world applications such as robotics \citep{nikolaidis2017mutual,macalpine2014robocup}, autonomous driving \citep{sadigh2016planning,cusumano2025robust}, and communication \citep{foerster2016learning,bard2020hanabi}. This challenge is formalised by the \ac{zsc} setting, as illustrated in Figure~\ref{fig:lunar-coordination}. \Acl*{zsc} can be viewed as an equilibrium-selection problem: \textit{when multiple high-payoff equilibria exist, how can unseen partners independently select compatible strategies without prior communication?}~\citep{pmlr-v139-treutlein21a}. \Ac{sp}, a widely used \ac{marl} training paradigm~\citep{checkers_sp_samuel,Tesauro1994TDGammonAS}, does not resolve this problem. Since agents train exclusively with copies of themselves, they converge to highly specialised yet arbitrary conventions that achieve high payoffs in \ac{sp} but fail to generalise to unseen partners.

Prior work~\citep{hu2020otherplay,hu2021offbelief,cui2022offteam,forkel2026entropy} has made significant progress on \ac{zsc} in complex environments such as Hanabi~\citep{bard2020hanabi} and Overcooked~\citep{carroll2019utility}. However, many of these approaches rely on restrictive assumptions or computationally expensive procedures, limiting their general applicability. These limitations are evident in benchmarks such as Overcooked V2~\citep{gessler2025overcookedv2} and Yokai~\citep{ruhdorfer2026yokai}, where such methods are either not easily applicable or achieve low \ac{ixp}, i.e., \ac{xp} between policies trained with the same algorithm but different random seeds. Moreover, existing \ac{zsc} methods tend to \textit{under-coordinate}: they adopt fixed strategies and do not support forming conventions with novel partners at test time~\citep{gessler2025overcookedv2}.

\Acs{zsc} aims to enable human-AI coordination, but its current formulation does not capture settings in which humans pre-agree on conventions before the start of the task. This includes Hanabi, where players may agree on how to interpret hints~\citep{hgroupconventions}, and pickup soccer, where teammates may agree on player positions, such as who will play in goal~\citep{rostaraPickupRules}. Existing methods require hand-coded conventions~\citep{bredell2025conventions} or costly LLM inference~\citep{dizdarevic2025adhoc,hu2023instructrl}, leaving this setting an open problem for human-AI collaboration.

In this work, we investigate the use of \acp{llm} to address these challenges. Our motivation comes from their growing ability to tackle difficult scientific problems~\citep{openai2026tenadvances,bubeck2025science} and coordinate in large multi-agent teams~\citep{openai2026huggingface}. The recent breakthrough on the Navier--Stokes Millennium Prize Problem highlights a combination of these capabilities, with approximately 10{,}000 agents collaborating toward a solution~\citep{openai2026navierstokesagents}. In our setting, we use \acp{llm} to generate policies as executable code, motivated by their rapidly advancing coding capabilities~\citep{kwa2025measuring} and recent success in synthesising control policies for robotics~\citep{liang2023codeaspolicies,fu2026capx,lu2026aspire}. Prior work~\citep{hayler2026zeroshot}, using an older generation of models (e.g GPT 5.2) found that \ac{llm}-generated policies perform poorly at \ac{zsc}, even in simplified coordination tasks. We revisit this finding in light of significant advancements in the current frontier \acp{llm}. Whereas prior work generated policies in a single forward pass, we use an off-the-shelf coding harness and runtime environment to enable iterative refinement through rollouts. We refer to these systems as coding agents.

\paragraph{Open challenges in zero-shot coordination.}
We briefly summarise the current state of the field before presenting our contributions. Overcooked V2 and Yokai remain unsolved, as reflected by low \acs{ixp} scores of current ZSC methods. In Hanabi, existing methods achieve high self-play and \acs{ixp} scores, but they do not support settings where partners pre-agree on conventions.
Developing agents that reliably follow human-compatible conventions remains an open problem. Additionally, \citet{hayler2026zeroshot} claim that \ac{llm} agents struggle with \ac{zsc} even in simple tasks. These findings led the field to stipulate that further progress required algorithmic innovation \cite{gessler2025overcookedv2}. 

\paragraph{Our Contributions} Surprisingly, we find that advances in general-purpose LLMs are sufficient to address these open ZSC challenges. We find that off-the-shelf coding agents, given the objective of maximising XP with unseen partners, achieve state-of-the-art IXP returns on Overcooked V2 and Yokai using only SP feedback during training. This compatibility extends across model families and to newly designed Overcooked V2 layouts. Their strong performance on Overcooked V2 tasks that require test-time convention formation provides evidence that generated policies can \textit{mutually co-adapt} at deployment. In Hanabi, given the H-group Level 1 conventions used by human players, coding agents outperform existing \ac{zsc} baselines in cross-play with the Will-Hanabi bot~\citep{willhanabibot}, a community-developed bot that plays by these conventions.
That said, these policies fall short of the strongest baselines in \acs{sp} and \acs{ixp}.
Towards this, we introduce a hybrid method that regularises the \acs{sp} objective towards actions selected by a coding agent's policy.
The hybrid achieves \acs{sp} and \acs{ixp} scores comparable to the strongest baselines and the highest cross-play returns with both the AH2AC2 bots~\citep{dizdarevic2025adhoc} and the Will-Hanabi bot.
We also find that more capable \acp{llm} generate policies with higher \ac{sp} and \ac{ixp} returns.
Additionally, we probe the models' prior knowledge of the benchmarks and find evidence of genuine \ac{zsc} capability.
When asked to name the six benchmark layouts of Overcooked V2, GPT-6 Astra could not name a single one correctly, yet its generated policies outperformed every \ac{zsc} baseline.
Finally, our qualitative analysis identifies several strategies in the generated policies, including adopting complementary roles, modelling partners' responses, and forming conventions at test time.

\begin{figure*}[t]
    \input{Figures/learning_algorithm}
\end{figure*}
\section{Related Work}
\textbf{Learning to coordinate with unseen partners.}
\Ac{sp} can produce strong cooperative policies~\citep{foerster2019bad,hu2020sad}, but independent training runs often converge to incompatible conventions. Several methods aim to address this by regularising the learning process. Other-Play~\citep{hu2020otherplay} uses environment symmetries to discourage arbitrary conventions, while entropy-regularised policy gradients~\citep{forkel2026entropy} can yield symmetry-equivariant policies under sufficiently strong regularisation. Off-Belief Learning~\citep{hu2021offbelief} instead constrains how agents interpret past actions through a specified reference policy, enabling compatible reasoning across independent runs. Population-based methods~\citep{strouse2021collaborating,lupu2021trajectory,sarkar2023diverse,lauffer2025rpg} use diverse training populations to improve coordination with unseen partners. However, these methods rely on restrictive assumptions that prohibit general applicability. Other-Play requires prior specification of the environment's relevant symmetries, limiting its applicability when these symmetries are unknown or difficult to identify~\citep{hu2020otherplay}. OBL is also not model-free: its counterfactual updates require simulator access to generate fictitious rollouts, and its hierarchical structure introduces significant computational overhead~\citep{forkel2026entropy, ruhdorfer2026yokai}. Training against a population of agents is computationally costly, and it remains unclear how to construct a population that captures the full range of partner behaviours an agent is likely to encounter at test time~\citet{strouse2021collaborating}. Finally, \citet{forkel2026entropy} show that there exist environments where entropy regularisation sufficient to ensure symmetry equivariance yields poor returns relative to other symmetry-equivariant policies. The Overcooked V2~\citep{gessler2025overcookedv2} and Yokai~\citep{ruhdorfer2026yokai} benchmarks highlight these challenges where current methods exhibit significant gaps between \ac{sp} and \ac{ixp} returns. Certain Overcooked V2 layouts require establishing conventions online during interaction, a capability not supported by current \acs{zsc} methods~\citep{gessler2025overcookedv2}. Beyond \acs{ixp}, a broader goal is coordination with humans, sometimes evaluated directly in prior work~\citep{hu2020otherplay}. To reduce the cost and improve the reproducibility of such evaluations, \citet{dizdarevic2025adhoc} introduce the AH2AC2 benchmark, which includes bots trained on over 147,000 human-played Hanabi games.

\textbf{Following shared conventions.}
In certain coordination settings, partners may agree to follow a fixed set of conventions without training together; for example, in Hanabi, human players often agree to play according to the H-group set of conventions~\citep{hgroupconventions}. The challenge is then to produce policies compatible with unseen partners that adopt the same conventions. InstructRL~\citep{hu2023instructrl} uses language-model priors to regularise deep \ac{rl} toward simple hint conventions in Hanabi. \citet{bredell2025conventions} incorporate H-group-derived conventions into an augmented action space of a policy trained using deep \ac{rl}. AH2AC2~\citep{dizdarevic2025adhoc} instead prompts an \ac{llm} to select actions directly from observations given descriptions of H-group Level 1 conventions. InstructRL and AH2AC2 both incur \ac{llm} inference costs for action-level guidance, while the approach of \citet{bredell2025conventions} requires manually implementing convention logic. In contrast, coding agents translate convention descriptions directly into executable policies, and the proposed hybrid method uses them to guide SP training.

\textbf{\Acp{llm} for policy generation and optimisation.}
Prior work has explored \ac{llm}-based cooperation using specialised agent scaffolds in Overcooked and Hanabi~\citep{zhang2024proagent,agashe2025llmcoordination,ramesh2026sparks}. \citet{hayler2026zeroshot} study generic setups in which \acp{llm} either select actions directly or generate executable policies, reporting coordination failures even in simple environments. Their code generation setup relies on single-step generation, whereas we use an off-the-shelf coding harness and runtime environment that supports iterative policy refinement. This connects our approach to \ac{llm}-based optimisation in other domains, where candidate solutions are revised using feedback~\citep{madaan2023selfrefine,yang2024opro,novikov2025alphaevolve}. However, these methods evaluate candidates directly against a well-specified objective; in our setting, coding agents optimise a cross-play objective with SP feedback.

\begin{table*}[!t]
    % Baseline source: Gessler et al. (2025), Table 2 (p. 9), official ICLR proceedings.
% Coding-agent source: verified xp_matrix_index.json snapshot 2026-09-20T16:09:39.777886+00:00.
% Fable test-time source: audited six-seed extension 2026-09-23; see overcooked_v2_sources.json.
% Source values and published baseline SDs are retained in overcooked_v2_sources.json.
% Coding-agent SDs describe dispersion across six policy-seed scores (ddof=1).
% IXP scores average partners and both roles per seed; these are not confidence intervals.
    \centering
    \caption{\textbf{Coordination performance on Overcooked V2.} Mean episode return, with $\pm$ SD shown below each mean. Coding-agent results use six seeds and 1,024 games per policy pairing; published baselines, taken directly from \citet{gessler2025overcookedv2}, use ten seeds and 500 games per pairing. We find that coding agents achieve the highest \acs{ixp} return on all six evaluated layouts.}
    \label{tab:overcooked-v2-main}
    \begingroup
    \small
    \setlength{\tabcolsep}{2.6pt}
    \renewcommand{\arraystretch}{1.0}
    \setlength{\heavyrulewidth}{0.6pt}
    \setlength{\lightrulewidth}{0.35pt}
    \sisetup{mode=text, detect-weight=true, table-number-alignment=right}
    \newcommand{\ocvalue}[2]{%
        \shortstack[c]{\strut{\rmfamily\num[detect-family=true]{#1}}\\[-1pt]%
        {\scriptsize\rmfamily\mdseries\color{black!70}\ensuremath{\pm}\,\num[detect-family=true]{#2}}}}
    \colorlet{ocsp}{citationnavy}
    \definecolor{ocixp}{HTML}{9A5B34}
    \newcommand{\ocbestsp}[2]{\cellcolor{ocsp!15}\ocvalue{#1}{#2}}
    \newcommand{\ocbestixp}[2]{\cellcolor{ocixp!15}\ocvalue{#1}{#2}}
    \newcommand{\ocmetric}[1]{{\scriptsize\rmfamily\color{oc#1}\acs{#1}}}
    \begin{tabular}{@{}l@{\hspace{14pt}}cc@{\hspace{14pt}}cc@{\hspace{14pt}}cc@{\hspace{14pt}}cc@{\hspace{14pt}}cc@{\hspace{14pt}}cc@{}}
    \toprule
    & \multicolumn{4}{@{}c@{\hspace{14pt}}}{\footnotesize Demo Cook} & \multicolumn{4}{@{}c@{\hspace{14pt}}}{\footnotesize Grounded Coord.} & \multicolumn{4}{@{}c@{}}{\footnotesize Test-time} \\
    \cmidrule(l{0pt}r{14pt}){2-5}\cmidrule(l{0pt}r{14pt}){6-9}\cmidrule(l{0pt}r{0pt}){10-13}
    & \multicolumn{2}{@{}c@{\hspace{14pt}}}{\footnotesize Simple} & \multicolumn{2}{@{}c@{\hspace{14pt}}}{\footnotesize Wide} & \multicolumn{2}{@{}c@{\hspace{14pt}}}{\footnotesize Simple} & \multicolumn{2}{@{}c@{\hspace{14pt}}}{\footnotesize Ring} & \multicolumn{2}{@{}c@{\hspace{14pt}}}{\footnotesize Simple} & \multicolumn{2}{@{}c@{}}{\footnotesize Wide} \\
    \cmidrule(l{0pt}r{14pt}){2-3}\cmidrule(l{0pt}r{14pt}){4-5}\cmidrule(l{0pt}r{14pt}){6-7}\cmidrule(l{0pt}r{14pt}){8-9}\cmidrule(l{0pt}r{14pt}){10-11}\cmidrule(l{0pt}r{0pt}){12-13}
    {\footnotesize Algorithm} $\mathcal{L}$ & \ocmetric{sp} & \ocmetric{ixp} & \ocmetric{sp} & \ocmetric{ixp} & \ocmetric{sp} & \ocmetric{ixp} & \ocmetric{sp} & \ocmetric{ixp} & \ocmetric{sp} & \ocmetric{ixp} & \ocmetric{sp} & \ocmetric{ixp} \\[2pt]
    \midrule
    {\footnotesize Self-Play} & \ocvalue{172}{19} & \ocvalue{36}{82} & \ocvalue{195}{4} & \ocvalue{40}{61} & \ocvalue{154}{10} & \ocvalue{-55}{124} & \ocvalue{164}{10} & \ocvalue{-12}{75} & \ocvalue{145}{22} & \ocvalue{-81}{99} & \ocvalue{194}{12} & \ocvalue{-30}{96} \\
    \addlinespace[2pt]
    {\footnotesize State-Augmented} & \ocvalue{170}{16} & \ocvalue{112}{50} & \ocvalue{192}{7} & \ocvalue{60}{60} & \ocbestsp{156}{29} & \ocvalue{-17}{115} & \ocbestsp{165}{7} & \ocvalue{-16}{65} & \ocvalue{161}{18} & \ocvalue{-55}{103} & \ocvalue{137}{82} & \ocvalue{-12}{72} \\
    \addlinespace[2pt]
    {\footnotesize Other-Play} & \ocvalue{149}{25} & \ocvalue{47}{48} & \ocvalue{162}{23} & \ocvalue{62}{53} & \ocvalue{122}{24} & \ocvalue{4}{73} & \ocvalue{121}{36} & \ocvalue{-9}{21} & \ocvalue{121}{37} & \ocvalue{-3}{51} & \ocvalue{175}{54} & \ocvalue{-11}{42} \\
    \addlinespace[2pt]
    {\footnotesize Fictitious Co-Play} & \ocvalue{22}{48} & \ocvalue{1}{25} & \ocvalue{45}{43} & \ocvalue{13}{28} & \ocvalue{64}{58} & \ocvalue{46}{47} & \ocvalue{86}{34} & \ocvalue{6}{46} & \ocvalue{35}{44} & \ocvalue{6}{29} & \ocvalue{95}{56} & \ocvalue{23}{40} \\
    \greytablerule{13}
    {\footnotesize GPT-5.6 Sol} & \ocbestsp{207}{16} & \ocvalue{174}{27} & \ocvalue{185}{32} & \ocvalue{162}{29} & \ocvalue{149}{18} & \ocvalue{129}{9} & \ocvalue{152}{17} & \ocvalue{133}{14} & \ocbestsp{180}{1} & \ocvalue{155}{19} & \ocvalue{214}{17} & \ocvalue{202}{11} \\
    \addlinespace[2pt]
    {\footnotesize Claude Opus 5} & \ocvalue{198}{4} & \ocbestixp{197}{1} & \ocvalue{202}{4} & \ocvalue{193}{8} & \ocvalue{139}{8} & \ocvalue{130}{5} & \ocvalue{147}{4} & \ocvalue{130}{4} & \ocvalue{172}{6} & \ocvalue{168}{2} & \ocvalue{209}{11} & \ocvalue{120}{26} \\
    \addlinespace[2pt]
    {\footnotesize GPT-6 Astra} & \ocvalue{198}{3} & \ocvalue{191}{7} & \ocbestsp{213}{9} & \ocbestixp{211}{4} & \ocvalue{138}{4} & \ocbestixp{137}{1} & \ocvalue{152}{3} & \ocbestixp{150}{2} & \ocvalue{161}{12} & \ocvalue{155}{5} & \ocvalue{214}{11} & \ocvalue{177}{26} \\
    \addlinespace[2pt]
    {\footnotesize Claude Fable 5.1} & \ocvalue{200}{11} & \ocvalue{196}{1} & \ocvalue{212}{16} & \ocvalue{203}{5} & \ocvalue{134}{3} & \ocvalue{134}{1} & \ocvalue{158}{16} & \ocvalue{130}{7} & \ocvalue{174}{9} & \ocbestixp{172}{4} & \ocbestsp{231}{10} & \ocbestixp{218}{15} \\
    \bottomrule
    \end{tabular}
    \vspace{-4mm}
    % Pale navy and orange cells mark the highest SP and IXP per task, respectively, ranked before rounding.
    \endgroup

\end{table*}
\section{Background}
\label{sec:background}

We focus on fully cooperative multi-agent settings that can be modelled as a
\ac{dec-pomdp}~\citep{bernstein2002complexity}.
A \ac{dec-pomdp} $\mathcal{M}$ is defined as the tuple
$(\mathcal{S}, n, \{\mathcal{A}^i\}_{i=1}^n,
\{\mathcal{O}^i\}_{i=1}^n, \mathcal{T}, \mathcal{R}, T, \gamma, \rho_0)$,
where $\mathcal{S}$ denotes the state space, $n$ the number of agents,
$\mathcal{A}^i$ and $\mathcal{O}^i$ the action and observation spaces of
agent $i$, respectively, $\mathcal{T}$ the transition function,
$\mathcal{R}$ the shared reward function, $T$ the time horizon, $\gamma$
the discount factor, and $\rho_0$ the initial-state distribution.
The joint action space is given by
$\mathcal{A} = \prod_{i=1}^n \mathcal{A}^i$.
At timestep $t$, the environment is in state $s_t \in \mathcal{S}$ and
the agents jointly select an action
$\mathbf{a}_t = (a_t^1,\ldots,a_t^n) \in \mathcal{A}$,
after which the environment transitions according to
$\mathcal{T}: \mathcal{S} \times \mathcal{A} \rightarrow \Delta(\mathcal{S})$
and the agents receive a shared reward given by
$\mathcal{R}: \mathcal{S} \times \mathcal{A} \rightarrow \mathbb{R}$.
Each agent $i$ receives only a local observation
$o_t^i = \mathcal{O}^i(s_t)$ and acts based on its local
\ac{aoh}
$\tau_t^i = (o_0^i, a_0^i, \ldots, a_{t-1}^i, o_t^i)$.
A local policy $\pi^i(a_t^i \mid \tau_t^i)$ specifies the probability
with which agent $i$ selects action $a_t^i$ given its \ac{aoh}.
We denote the joint policy by
$\boldsymbol{\pi} = (\pi^1,\ldots,\pi^n)$,
such that the joint action distribution factors as
$\boldsymbol{\pi}(\mathbf{a}_t \mid \boldsymbol{\tau}_t)
= \prod_{i=1}^n \pi^i(a_t^i \mid \tau_t^i)$. A learning algorithm $\mathcal{L}$ maps a \ac{dec-pomdp} $\mathcal{M}$ to a distribution over joint policies.
The \ac{sp} return of a joint policy $\boldsymbol{\pi}$ is defined as the expected discounted return obtained when each agent $i$ acts according to its corresponding local policy $\pi^i(\cdot \mid \tau_t^i)$:
\begin{equation}
\label{eq:self-play-return}
J_{\text{\acs*{sp}}}(\boldsymbol{\pi})
=
\mathbb{E}_{\boldsymbol{\pi}}
\left[
\sum_{t=0}^{T-1}
\gamma^t \mathcal{R}(s_t,\mathbf{a}_t)
\right].
\end{equation}

Given a \ac{dec-pomdp} $\mathcal{M}$ with $n$ agents, let $\boldsymbol{\pi}_1,\ldots,\boldsymbol{\pi}_n$ be joint policies sampled as $\boldsymbol{\pi}_j=(\pi_j^1,\ldots,\pi_j^n)\sim\mathcal{L}_j(\mathcal{M})$. The \ac{xp} return is defined as:
\begin{multline*}
J_{\text{\acs*{xp}}}(\boldsymbol{\pi}_1,\ldots,\boldsymbol{\pi}_n) \\
:= \mathbb{E}_{\phi \sim \operatorname{Perm}(n)}
\left[
J_{\text{\acs*{sp}}}
\left(
\pi_{\phi(1)}^1,\ldots,\pi_{\phi(n)}^n
\right)
\right],
\end{multline*}
where $\operatorname{Perm}(n)$ is the set of permutations of $\{1,\ldots,n\}$. 

\paragraph{Evaluation Criteria}

\textit{\Acl{ixp}} (\acs{ixp}) measures \ac{xp} between policies produced by independent runs of the same learning algorithm with different random seeds. High \ac{ixp} is a necessary, but not sufficient, condition for \ac{zsc}: if an algorithm does not produce mutually compatible policies across its own training runs, we cannot expect those policies to coordinate reliably with unseen partners. Despite this limitation, \ac{ixp} is still the primary metric used to compare ZSC methods in the literature~\citep{forkel2026entropy}.

Some prior work evaluates agents against held-out populations~\citep{strouse2021collaborating,liang2024generative}, generating these partners by varying training seeds, architectures, checkpoints, algorithms, etc. Diversity and held-out status alone, however, do not establish that these populations represent the coordination partners an agent is likely to encounter in the real world~\citep{wang2024zsceval}. We prioritise real-world coordination partners (e.g., humans) and therefore limit our held-out population evaluations to the AH2AC2 bots trained on human gameplay~\citep{dizdarevic2025adhoc} and the Will-Hanabi bot, which follows conventions used by human players~\citep{willhanabibot}.

\begin{table}[!t]
    % Baseline source: Ruhdorfer et al. (2026), Tables 2 and 6 (PDF pp. 9--10).
% Exact coding-agent means, six seed aggregates, and SEMs: yokai_sources.json.
% All means and SEMs use one decimal, rounded from full-precision source values.
% Highlighting compares full-precision coding-agent means with published baseline means.
    \centering
    \caption{\textbf{Coordination performance on two-player Yokai.} Mean episode return $\pm$ SEM. Coding-agent results use six seeds and 1,024 games per policy pairing; the 12-seed method baselines are taken directly from \citet{ruhdorfer2026yokai}, were evaluated on 5,000 games. They also report an illustrative human-human baseline of \(5.7 \pm 0.5\) across 25 games in the nine-card imperfect-memory setting. We find that coding agents achieve the highest \acs{ixp} in all three configurations; their best nine-card imperfect-memory \acs{ixp} also exceeds this human reference.}
    \label{tab:yokai-main}
    \begingroup
    \small
    \setlength{\tabcolsep}{1pt}
    \renewcommand{\arraystretch}{1.0}
    \setlength{\heavyrulewidth}{0.6pt}
    \setlength{\lightrulewidth}{0.35pt}
    \sisetup{mode=text, table-number-alignment=right}
    \newcommand{\ykvalue}[2]{%
        \mbox{{\rmfamily\num[detect-family=true]{#1}}%
        {\scriptsize\rmfamily\mdseries\color{black!70}\hspace{0.2em}\ensuremath{\pm}\hspace{0.2em}\num[detect-family=true]{#2}}}}
    \newcommand{\ykbestsp}[2]{\cellcolor{citationnavy!15}\ykvalue{#1}{#2}}
    \newcommand{\ykbestixp}[2]{\cellcolor{citationnavy!15}\ykvalue{#1}{#2}}
    \begin{tabular}{@{}l@{\hspace{6pt}}l@{\hspace{8pt}}r@{\hspace{8pt}}r@{}}
    \toprule
    {\footnotesize Configuration} & {\footnotesize Algorithm} $\mathcal{L}$ & \multicolumn{1}{c}{\footnotesize\acs{sp}} & \multicolumn{1}{c}{\footnotesize\acs{ixp}} \\[2pt]
    \midrule
    \multirow{8}{*}{\footnotesize\shortstack[l]{9 cards\\Perfect memory}} & {\footnotesize IPPO} & \ykbestsp{7.5}{0.3} & \ykvalue{1.6}{0.2} \\
     & {\footnotesize IPPO+High Ent.} & \ykbestsp{7.5}{0.1} & \ykvalue{4.1}{0.8} \\
     & {\footnotesize Other-Play} & \ykvalue{6.6}{0.2} & \ykvalue{4.8}{0.1} \\
     & {\footnotesize OBL L4} & \ykvalue{4.5}{0.2} & \ykvalue{2.8}{0.3} \\
    \greytableruleafterfirst{3}
     & {\footnotesize GPT-5.6 Sol} & \ykvalue{4.9}{0.3} & \ykvalue{4.7}{0.1} \\
     & {\footnotesize Claude Opus 5} & \ykvalue{6.0}{0.5} & \ykvalue{5.6}{0.3} \\
     & {\footnotesize GPT-6 Astra} & \ykvalue{7.5}{0.5} & \ykbestixp{7.3}{0.2} \\
     & {\footnotesize Claude Fable 5.1} & \ykvalue{7.0}{0.3} & \ykvalue{7.0}{0.1} \\
    \midrule
    \multirow{5}{*}{\footnotesize\shortstack[l]{9 cards\\Imperfect memory}} & {\footnotesize Other-Play} & \ykvalue{0.0}{0.0} & \ykvalue{0.0}{0.0} \\
    \greytableruleafterfirst{3}
     & {\footnotesize GPT-5.6 Sol} & \ykvalue{4.9}{0.5} & \ykvalue{4.3}{0.1} \\
     & {\footnotesize Claude Opus 5} & \ykvalue{5.2}{0.5} & \ykvalue{5.2}{0.2} \\
     & {\footnotesize GPT-6 Astra} & \ykvalue{6.3}{0.5} & \ykvalue{6.5}{0.2} \\
     & {\footnotesize Claude Fable 5.1} & \ykbestsp{7.1}{0.3} & \ykbestixp{6.9}{0.2} \\
    \midrule
    \multirow{5}{*}{\footnotesize\shortstack[l]{16 cards\\Perfect memory}} & {\footnotesize IPPO} & \ykvalue{0.0}{0.0} & \ykvalue{0.0}{0.0} \\
    \greytableruleafterfirst{3}
     & {\footnotesize GPT-5.6 Sol} & \ykvalue{6.3}{0.4} & \ykvalue{5.1}{0.2} \\
     & {\footnotesize Claude Opus 5} & \ykvalue{6.1}{0.6} & \ykvalue{4.2}{0.7} \\
     & {\footnotesize GPT-6 Astra} & \ykvalue{8.2}{0.9} & \ykbestixp{8.0}{0.4} \\
     & {\footnotesize Claude Fable 5.1} & \ykbestsp{8.4}{0.2} & \ykvalue{8.0}{0.1} \\
    \bottomrule
    \end{tabular}
    \vspace{-2mm}
    % Pale navy cell fills mark the highest reported scores, not statistical significance.
    \endgroup

\end{table}

\section{Policy Generation Setup}
\label{sec:policy-generation}

We define a coding agent as a pair $\mathcal{C}=(L,\mathcal{H})$, consisting of \iac{llm} $L$ and a harness $\mathcal{H}$. The harness connects the model to a development environment $\mathcal{E}$ by providing tools for interacting with it, including file editing and code execution. The model interacts with the environment through successive tool calls, conditioned on the task instructions and interaction history. The harness executes these calls and returns their outputs to the model. We use each model’s native coding harness, e.g, Codex for OpenAI models and Claude Code for Anthropic models.

In our setting, $\mathcal{E}$ contains a sandboxed workspace for storing policy code and experimental artefacts, together with a runtime for executing and evaluating candidate policies. The runtime provides access to a \acs{gpu} to accelerate policy evaluation. The workspace provides a task description specifying the game, the learning objective, and the policy \acs{api}. The game description includes the rules, configuration, each player's observations and available actions, the shared reward, and episode termination conditions. The objective is to construct policies that coordinate robustly with independently generated partners. The workspace also includes a default policy template implementing the policy and rollout \acp{api} for evaluating candidates. We do not provide any solution strategies, except for the conventions provided in the Hanabi H-group condition, and the agent has no access to the simulator source code or the internet through its tools. The policy \ac{api} enforces decentralised execution: each player acts using only its local observation and private memory. The coding agent determines how this memory is represented and updated, allowing policies to retain information from past observations. To accommodate asymmetric player roles in Overcooked V2, the coding agent implements a separate local policy for each role. In Hanabi and Yokai, where the roles are symmetric, it implements a single local policy, with \ac{sp} pairing two instances that maintain separate private memories.

With $L$ and $\mathcal{H}$ fixed, the coding agent induces a stochastic learning algorithm that takes a task description and a computational budget as inputs and returns an executable joint policy as code. It interacts with $\mathcal{E}$ to implement candidate policies and evaluate them through rollouts, using the resulting feedback to guide further revisions. This iterative experimentation constitutes the learning process, while the language model's parameters remain fixed throughout. To make this process interpretable, we provide an experiment ledger tool whose \ac{api} is documented in the task description. We enforce that the agent use this tool to record the hypothesis behind each revision, its evaluation results, and the stated rationale for retaining or rejecting it.

Each independent run of the learning algorithm, which we sometimes refer to as an independent seed, is initialised in a fresh session and workspace with no access to the policies, interaction histories, or evaluation results of other seeds. Each seed receives a differently worded prompt that preserves the task specification and objective. This reflects our assumption that independent coding agents need not receive identical prompts and makes our XP evaluation sensitive to brittle conventions that rely on identical prompts. We allocate a two-hour wall-clock budget to each policy-generation run and make the elapsed time available to the agent. After each trial, we instruct the agent to update its selection, retaining the candidate it judges most likely to maximise the \ac{xp} objective. The agent may choose to terminate early; in which case its last selected policy constitutes the output. During evaluation, this policy is executed without further \ac{llm} calls. Figure~\ref{fig:coding-agent-learning} illustrates the learning process and initial workspace contents; Appendix~\ref{app:policy-generation-instructions} provides the objective and final policy selection instructions.

\begin{table*}[!t]
    % Exact values, seed aggregates, hashes, and reporting definitions: hanabi_sources.json.
% Existing-method baselines use four checkpoint seeds; coding agents use six independent generations.
    \centering
    \caption{\textbf{Coordination performance on two-player Hanabi.} Mean $\pm$ SEM over six coding-agent runs, six hybrid policies, and four checkpoints per baseline, with 1,024 games per policy pairing. We evaluate R2D2-based OBL checkpoints from \citet{hu2021offbelief} and IPPO, IPPO+HE, and IPPO-based OBL checkpoints from \citet{forkel2026entropy}. We find that coding agents underperform the \acs{zsc} baselines in \acs{sp} and \acs{ixp}, but achieve significantly higher \acs{xp} with the Will-Hanabi bot. Hybrid policies match the baselines in SP and IXP while achieving the highest XP returns with both the AH2AC2 bots and the Will-Hanabi bot.}
    \label{tab:hanabi-main}
    \begingroup
    \small
    \setlength{\tabcolsep}{3pt}
    \renewcommand{\arraystretch}{1.0}
    \setlength{\heavyrulewidth}{0.6pt}
    \setlength{\lightrulewidth}{0.35pt}
    \sisetup{mode=text, table-number-alignment=right}
    \newcommand{\hnvalue}[2]{%
        {\rmfamily\makebox[3.0em][r]{\num[detect-family=true]{#1}}}%
        {\scriptsize\rmfamily\mdseries\hspace{0.1em}\ensuremath{\pm}\hspace{0.1em}\makebox[2.7em][l]{\num[detect-family=true]{#2}}}}
    \newcommand{\hnbestsp}[2]{\cellcolor{citationnavy!15}\hnvalue{#1}{#2}}
    \newcommand{\hnbestixp}[2]{\cellcolor{citationnavy!15}\hnvalue{#1}{#2}}
    \newcommand{\hnbestbc}[2]{\cellcolor{citationnavy!15}\hnvalue{#1}{#2}}
    \newcommand{\hnbestwillhanabibot}[2]{\cellcolor{citationnavy!15}\hnvalue{#1}{#2}}
    \newcommand{\hnmodelrule}{%
        \multispan{6}\dottedtableleaders\cr}
    \begin{tabular}{@{}llrrrr@{}}
    \toprule
    {\footnotesize Algorithm} $\mathcal{L}$ & {\footnotesize Setting} & \multicolumn{1}{c}{\footnotesize\acs{sp}} & \multicolumn{1}{c}{\footnotesize\acs{ixp}} & \multicolumn{1}{c}{\footnotesize\shortstack{AH2AC2 bot \acs{xp}\\(\acs{sp}: 23.80)}} & \multicolumn{1}{c}{\footnotesize\shortstack{Will-Hanabi bot \acs{xp}\\(\acs{sp}: 21.05)}} \\[2pt]
    \midrule
    {\footnotesize IPPO} & {\footnotesize Standard} & \hnbestsp{24.53}{0.02} & \hnvalue{15.63}{2.13} & \hnvalue{13.60}{1.54} & \hnvalue{3.96}{0.48} \\
    {\footnotesize IPPO+HE} & {\footnotesize Standard} & \hnvalue{24.48}{0.01} & \hnbestixp{24.43}{0.02} & \hnvalue{11.64}{0.09} & \hnvalue{4.41}{0.08} \\
    {\footnotesize OBL L4 (R2D2)} & {\footnotesize Standard} & \hnvalue{24.12}{0.03} & \hnvalue{23.76}{0.08} & \hnvalue{19.70}{0.16} & \hnvalue{5.19}{0.04} \\
    {\footnotesize OBL L6 (IPPO)} & {\footnotesize Standard} & \hnvalue{24.35}{0.01} & \hnvalue{24.32}{0.00} & \hnvalue{14.92}{0.15} & \hnvalue{5.56}{0.13} \\
    \greytablerule{6}
    \multirow{2}{*}{\footnotesize GPT-5.6 Sol} & {\footnotesize Standard} & \hnvalue{21.93}{0.19} & \hnvalue{18.74}{0.42} & \hnvalue{15.53}{2.25} & \hnvalue{7.42}{1.15} \\
     & {\footnotesize H-group L1} & \hnvalue{18.59}{0.65} & \hnvalue{18.26}{0.29} & \hnvalue{16.92}{0.57} & \hnvalue{18.54}{0.47} \\
    \hnmodelrule
    \addlinespace[1pt]
    \multirow{2}{*}{\footnotesize Claude Opus 5} & {\footnotesize Standard} & \hnvalue{20.00}{0.36} & \hnvalue{19.35}{0.12} & \hnvalue{18.60}{0.36} & \hnvalue{12.38}{0.63} \\
     & {\footnotesize H-group L1} & \hnvalue{20.05}{0.42} & \hnvalue{19.09}{0.20} & \hnvalue{17.75}{0.24} & \hnvalue{18.21}{0.63} \\
    \hnmodelrule
    \addlinespace[1pt]
    \multirow{2}{*}{\footnotesize GPT-6 Astra} & {\footnotesize Standard} & \hnvalue{19.69}{0.80} & \hnvalue{19.43}{0.35} & \hnvalue{17.02}{1.61} & \hnvalue{12.09}{1.55} \\
     & {\footnotesize H-group L1} & \hnvalue{21.53}{0.34} & \hnvalue{21.11}{0.19} & \hnvalue{17.87}{0.33} & \hnvalue{19.98}{0.25} \\
    \hnmodelrule
    \addlinespace[1pt]
    \multirow{2}{*}{\footnotesize Claude Fable 5.1} & {\footnotesize Standard} & \hnvalue{20.55}{0.26} & \hnvalue{19.50}{0.16} & \hnvalue{19.34}{0.68} & \hnvalue{14.87}{0.99} \\
     & {\footnotesize H-group L1} & \hnvalue{21.81}{0.21} & \hnvalue{20.80}{0.12} & \hnvalue{18.17}{0.40} & \hnvalue{20.25}{0.28} \\
    \greytablerule{6}
    {\footnotesize Hybrid (Astra)} & {\footnotesize H-group L1} & \hnvalue{24.28}{0.01} & \hnvalue{24.14}{0.05} & \hnbestbc{21.75}{0.18} & \hnbestwillhanabibot{21.97}{0.11} \\
    {\footnotesize Hybrid (Fable)} & {\footnotesize H-group L1} & \hnvalue{24.31}{0.02} & \hnvalue{23.98}{0.21} & \hnvalue{20.98}{0.23} & \hnvalue{21.69}{0.23} \\
    \bottomrule
    \end{tabular}
    \vspace{-4mm}

    % {\scriptsize AH2AC2 bot XP averages returns with the two behavioural-cloning bots, BC1 and BC2; all partner scores average both policy seat assignments. The \acs{ixp} aggregates share partners. H-group L1 denotes the policy-generation setting; all policies use standard Hanabi scoring.}
    % Header SP values: mean of BC1 x BC1 and BC2 x BC2, and the Will-Hanabi bot paired with itself, respectively; 1,024 games per pairing.
    % Pale navy cell fills mark the highest mean before rounding, not statistical significance.
    \endgroup

\end{table*}

\section{Experiments}

\label{sec:experiments}
We evaluate coding agents on Overcooked V2, Yokai, and Hanabi, which pose coordination challenges not addressed by current \ac{zsc} methods~\citep{gessler2025overcookedv2,ruhdorfer2026yokai,dizdarevic2025adhoc}. We use their corresponding public JaxMARL-based implementations~\citep{rutherford2024jaxmarl}. Our experiments can be reproduced with ChatGPT Pro and Claude Code Max subscriptions, each costing approximately \$200 USD per month. Subscription accounts are significantly subsidised relative to standard \acs{api} rates~\citep{semianalysis2026subscription}, making large-scale experimentation feasible. In Overcooked V2, we additionally evaluate Kimi K3.1 Pro and DeepSeek V4.1 Flash through their native coding harnesses using the OpenRouter \acs{api}, which costs approximately \$80 USD. The reasoning effort is always set to \texttt{high}. In the Hanabi H-group setting, we provide an offline adaptation of the H-group Level 1 documentation~\citep{hgroupconventions}, including worked examples.

\subsection{Self-Play and Inter-Seed \acs{xp}}
\label{sec:results}
As shown in Tables~\ref{tab:overcooked-v2-main} and~\ref{tab:yokai-main}, coding agents achieve state-of-the-art IXP scores across every tested environment configuration in OvercookedV2 and Yokai. In certain Yokai configurations (9C imperfect recall and 16C perfect recall), existing methods even fail to achieve meaningful SP scores, whereas coding agents remain effective. High \ac{ixp} scores on Overcooked V2's test-time coordination layouts showcase that the generated policies can mutually co-adapt to establish conventions at test-time (Table~\ref{tab:overcooked-v2-main}).

In Yokai's nine-card imperfect-memory setting, Claude Fable 5.1 and GPT 6 Astra achieve \ac{ixp} returns of 6.9 and 6.5, respectively, both exceeding the reported human baseline of 5.7~\citep{ruhdorfer2026yokai}. In Hanabi (Table~\ref{tab:hanabi-main}), coding agents achieve self-play returns of up to 21.93, surpassing previously reported two-player results of 17.5~\citep{ramesh2026sparks} and 13.33~\citep{agashe2025llmcoordination} by \acs{llm} based methods, both of which rely on specialised scaffolds.\footnote{Both works retain accumulated scores when all lives are lost, whereas we assign zero; Agashe et al.'s best-performing configuration reports no bomb-outs.} 

We provide a qualitative analysis of the generated policies in Appendix~\ref{app:generated-policies}, where we highlight code snippets that illustrate task-specific coordination strategies across all three environments. We find that agents model their partners’ behaviour, signal their intended conventions, detect deviations from an established convention, and subsequently revise their behaviour to establish a new one.

We also provide experiment-ledger entries in Appendix~\ref{app:experiment-ledgers}. Interestingly, these entries show Claude Fable 5.1 constructing alternative partner policies to test the cross-play robustness of its candidate policy. We provide learning curves (Appendix Figure~\ref{fig:learning-curves}) that show improvements in \ac{sp} and \ac{ixp} returns over the course of policy generation. Additionally, our ablation in Section~\ref{app:policy-generation-settings} suggests that iterative refinement is essential for consistently achieving high \ac{ixp}. We also report \ac{ixp} returns for GPT-5.6 Luna, Claude Sonnet 5, and GPT-5.5 across all three environments, and for Kimi K3.1 Pro and DeepSeek V4.1 Flash on Overcooked V2 (Appendix Tables~\ref{tab:overcooked-v2-full}, \ref{tab:yokai-full}, and~\ref{tab:hanabi-full}). We find that weaker models achieve lower \ac{sp} and \ac{ixp} returns, indicating that state-of-the-art \ac{ixp} observed in our experiments is currently limited to the most capable models. On Overcooked V2, we also compare coding agents with two population-based baselines, MEP~\citep{zhao2023mep} and GAMMA~\citep{liang2024generative} (Appendix Table~\ref{tab:overcooked-v2-full}).

\begin{figure*}[t]
    \input{Figures/cross_model/main_figure}
\end{figure*}
\subsection{\acs{xp} with Heterogeneous Partners}
\label{sec:heterogeneous-partners}

We next evaluate whether compatibility extends beyond independent runs of the same coding agent.  As shown in Figure~\ref{fig:cross-model-main} and Figure~\ref{fig:cross-model-remaining} in the Appendix, policies generated by different coding agents remain compatible. Across the tested Overcooked V2 and Yokai environment configurations, inter-model \ac{xp} exceeds the best reported \ac{ixp} baseline. Some pairings nevertheless show gaps between \ac{sp} and inter-model \ac{xp}, indicating that there still exists room for improvement. We also find that inter-model \ac{xp} tends to be highest between the most capable models, with the GPT-6 Astra and Claude Fable 5.1 pairing achieving the highest scores across the evaluated environments.

In Hanabi, we evaluate cross-play with two AH2AC2 bots~\citep{dizdarevic2025adhoc} and the Will-Hanabi bot~\citep{willhanabibot}, configured to follow H-group Level 1 conventions. Against the AH2AC2 bots, Claude Fable 5.1 achieves a return of $19.34 \pm 0.68$, comparable to $19.70 \pm 0.16$ for OBL L4 and above IPPO, IPPO+HE, and OBL L6 (Table~\ref{tab:hanabi-main}). Despite similar returns with the AH2AC2 bots, Fable and OBL L4 differ sharply in cross-play with the Will-Hanabi bot: without convention instructions, Fable scores 14.87, compared with 5.19 for OBL L4. Providing H-group Level 1 descriptions to Fable increases its \acs{xp} return to 20.25, close to the Will-Hanabi bot’s \acs{sp} return of 21.05, indicating that the generated policy reliably follows conventions used by human players. The cross-play performance with the Will-Hanabi bot extends to games with three to five players (Appendix Table~\ref{tab:multiplayer-hanabi}). Given the H-group conventions, the generated policies also execute the \textit{finesse} (Appendix Table~\ref{tab:hanabi-finesse}), a move \citet{hu2022selfexplaining} elicited through search tailored to Hanabi.

\subsection{Hybrid Methods}
\label{sec:hybrid-methods}
\Acp{llm} can translate textual descriptions of conventions into policy code, as evidenced by their strong cross-play scores in the H-group setting. However, code policies achieve lower \ac{sp} and \ac{ixp} scores in Hanabi than established \ac{zsc} methods (Section~\ref{sec:results}). We hypothesise that this gap reflects the difficulty of encoding belief updates that depend on a long history of play. On the other hand, recurrent neural policies can track these histories and learn to exploit specialised conventions over billions of timesteps of experience, achieving high \ac{sp} and \ac{ixp} scores. Existing recurrent methods, however, provide no direct way to train policies to follow conventions specified in text.

To combine these strengths, we train a recurrent neural policy with an auxiliary objective that regularises it towards the conventions adopted by the code policy. More specifically, the training objective is given by
\begin{equation}
\label{eq:hybrid-loss}
\begin{aligned}
\mathcal{L}_{\mathrm{hybrid}}(\theta)
&= \mathcal{L}_{\mathrm{SP}}(\theta)
   + \lambda\,\mathcal{L}_{\mathrm{CE}}(\theta), \\
\mathcal{L}_{\mathrm{CE}}(\theta)
&= -\mathbb{E}_{\pi_\theta}\!\left[
   \log \pi_\theta^i\!\left(a_{t,\mathrm{code}}^i \mid \tau_t^i\right)
   \right].
\end{aligned}
\end{equation}
Here, $\mathcal{L}_{\mathrm{SP}}$ is the self-play objective in Equation~\ref{eq:self-play-return}, and $\lambda \geq 0$ controls the strength of the imitation loss. This encourages the neural policy to match the code policy's actions while continuing to learn from self-play. We use IPPO+HE as the policy optimiser with $\lambda=0.2$.

Table~\ref{tab:hanabi-main} shows that the hybrid policies achieve \ac{sp} and \ac{ixp} scores comparable to the baselines, while obtaining the highest cross-play scores with the AH2AC2 bots and the Will-Hanabi bot. Notably, their cross-play score with the Will-Hanabi bot exceeds the bot's own self-play score. Together, these results show that a code policy can guide convention selection during self-play, combining the ability of coding agents to follow conventions specified in text with the ability of recurrent policies to learn from vast amounts of self-play experience.

\begin{figure*}[t]
    \input{Figures/cross_model/ocv2_objective_shared_layouts}
\end{figure*}
\subsection{Evaluating prior knowledge}
\label{sec:evaluating-prior-knowledge}
The strong \acs{zsc} performance of coding agents raises questions of how much they benefit from prior exposure to the evaluated tasks, a concern also raised in other domains~\citep{zhang2024careful,wu2024reasoning,jain2025livecodebench}. Unlike the \ac{zsc} baselines trained from scratch, coding agents can draw on knowledge acquired during pretraining, which may include, for example, information about the Overcooked V2 layouts and their coordination challenges. 

Towards this, we probe the models' prior knowledge of Overcooked V2 and Hanabi using verifiable questions scored by an \acs{llm} judge (Appendix~\ref{app:prior-knowledge}). The Overcooked V2 questions include asking models to name and reconstruct the six benchmark layouts, describe their coordination challenges, and explain how the benchmark differs from the original Overcooked. In Hanabi, the questions include recalling H-group Level 1 conventions and applying them to game scenarios, such as interpreting a partner's hint.

We found that GPT-6 Astra could not name any of the six Overcooked V2 layouts or describe their coordination challenges. Yet, its generated policies outperformed every evaluated \acs{zsc} baseline in \acs{ixp}. Both Claude models named all six layouts, but none of the four models reconstructed a layout exactly. In Hanabi, the four models earned 83\% to 89\% of the available points on convention recall, while their accuracy on the worked examples ranged from 42\% to 75\%. These findings suggest that prior familiarity may contribute to performance, but that memorisation alone is unlikely to account for the strong cross-play results reported in the preceding sections.

We also evaluate coding agents on two novel layouts beyond those introduced in the original Overcooked V2 paper, \textit{Block Grid} and \textit{Pipeline Cook} (\novelLayoutRef). The original Overcooked V2 environments each test a single \acs{zsc} capability, whereas \textit{Block Grid} and \textit{Pipeline Cook} pose more open-ended coordination challenges. Nevertheless, coding agents exhibit strong performance on these layouts, generating policies that remain compatible across independent generations (Appendix Table \ref{tab:overcooked-v2-novel}) and model types (Figure \ref{fig:ocv2-objective-shared-layouts}). These results provide further evidence of coding agents' genuine ability to generate robust policies, rather than relying on leaked knowledge of environments or their solution strategies. That said, disentangling the precise contribution of knowledge acquired during pretraining on \acs{zsc} performance is beyond the scope of this study.

\subsection{Policy Generation Capabilities}

We next examine how general capabilities of the \acs{llm} affect its policy generation performance for ZSC. This helps us study whether the lower \ac{ixp} scores of weaker models stem from limited policy-generation capability or specifically from difficulty generating \acs {zsc}- compatible policies. 

Towards this, we compare two objectives: maximising \acs{sp} return and maximising \acs{xp} return with independently generated partners (as shown in Appendix Figures~\ref{fig:objective-instructions} and~\ref{fig:objective-instructions-sp}). We keep the model, harness and computational budget fixed, changing only the objective and selection rule. Under the \acs{sp} objective, we re-evaluate each proposed policy, selecting the one with the highest \acs{sp} return; under the \acs{xp} objective, the model selects the final policy. 

Figure~\ref{fig:ocv2-objective-shared-layouts} (left) compares the resulting policies across the six Overcooked V2 layouts. We find that more capable models achieve higher \ac{sp} returns under the \ac{sp} objective while exhibiting a larger \ac{sp}--\ac{ixp} gap, indicating a greater capacity to exploit specialised conventions. Under the \ac{xp} objective, they likewise achieve higher \ac{ixp} returns, reflecting improved compatibility across independent generations. At the same time, less capable models still generate policies whose \ac{sp} returns exceed \ac{ixp} returns of more capable models. Together, these results suggest that more capable models can generate a wider range of policies, from highly specialised self-play strategies to robust policies for \acs{zsc}.

\section{Conclusion}
\label{sec:conclusion}
In this work, we investigated coding agents as generators of executable policies for \ac{zsc}. We found that frontier models equipped with off-the-shelf coding harnesses and runtime environments achieve state-of-the-art \ac{ixp} on Overcooked V2 and Yokai, with compatibility extending across model families and novel Overcooked V2 layouts. Their performance on tasks requiring test-time convention formation provides evidence that generated policies can adapt their coordination strategies during interaction. 

In Hanabi, we found that policies generated by coding agents achieved competitive cross-play scores with the AH2AC2 bots. When given the H-group Level 1 conventions, coding agents outperformed current \ac{zsc} baselines in cross-play with the Will-Hanabi bot. Their policies nevertheless scored lower in \ac{sp} and \ac{ixp}. Towards this end, we introduced a hybrid method that used a code policy to guide convention selection during self-play. The hybrid policies achieved \ac{sp} and \ac{ixp} scores comparable to the strongest baselines and the highest cross-play scores with both the AH2AC2 bots and the Will-Hanabi bot. Next, we found that policy generation capability under both \ac{sp} and \ac{xp} objectives improves with model capability, suggesting that further increases in model capability may reduce the performance gap of code policies in Hanabi. 

We note that shared knowledge acquired during pretraining may help independently generated policies coordinate. That said, people also draw on a shared understanding of what actions mean when coordinating with strangers~\citep{pmlr-v202-ma23e}. We probed the models' prior knowledge of Overcooked V2 and Hanabi and found that GPT-6 Astra could not name any of the six Overcooked V2 layouts, yet its policies outperformed all evaluated \acs{zsc} baselines in \acl{ixp}. Together with strong cross-play on novel layouts, this finding suggests that coding agents possess genuine \acs{zsc} capabilities that memorisation of benchmark solutions alone is unlikely to explain.

\ificmlshowauthors
\section*{Acknowledgements}
Shashank Reddy is partially supported by the Cooperative AI PhD Fellowship from the Cooperative AI Foundation.
Jakob N. Foerster is partially funded by the UKRI grant EP/Y028481/1 (originally selected for funding by the ERC).
The authors thank Johannes Forkel, Darius Muglich, Ravi Hammond and Kalyan Nadimpalli for insightful discussions and feedback on the paper.
\fi

\section*{Impact Statement}
We find that coding agents can generate policies that follow conventions specified in natural language. We evaluated these policies in simulated cooperative games, including Hanabi with bots trained on human gameplay or designed to follow human conventions. These results should not be interpreted as evidence of safe or effective coordination with humans in real-world settings.

\section*{AI Disclosure Statement}
In this work, we used generative AI tools to implement methods, namely Codex (w/ GPT 5.6 Sol, GPT 6 Sol, and GPT 6 Astra) to implement the coding agent sandbox for policy generation and to orchestrate, monitor and provide feedback on experiments. We have also used GPT 6 Astra to help interpret the policies generated by the coding agents. We have not used generative AI tools to propose or refine hypotheses. The use of generative AI to generate synthetic data sets, help develop theoretical models or conceptual frameworks, formulate mathematical claims, provide critical ingredients for proving mathematical claims, assist in the writing of proofs, assist with translation, clean and reformat datasets does not apply to this work. Additionally, we used generative AI tools to create or modify scientific figures, including tables and images, sourcing/searching for information, edit the research paper to improve readability and quality of writing, identify relevant literature, and format references.  We have reviewed all AI-assisted work, including text, code and figures; we have also verified the contents of the papers that were suggested by the \acsp{llm} as relevant work. We take responsibility for the final content of this work,
including text, claims or artefacts produced with the aid of generative AI.

\bibliography{iclr2027_conference}
\bibliographystyle{icml2027}

\onecolumn
\appendix
\input{Sections/appendix_contents}

\section{Policy-Generation Instructions}
\label{app:policy-generation-instructions}

\begin{figure}[htbp]
    \centering
    \begin{policyinstructionbox}
        \instructionheading{Objective}
        Maximize expected cooperative episode return when either role of this joint policy is paired with the complementary role from an independently generated policy. Your policy will be paired with independently generated partners. It should coordinate with them without assuming they use the same strategy, role assignments, or communication conventions. Seek competent, convention-robust behavior that can coordinate from locally observable evidence. Self-play return is a useful competence diagnostic, but it is not the optimization target.
        \tcblower
        \instructionheading{Selection rule}
        The primary evaluation metric is \acs*{xp}: cooperative episode return when this policy is paired with independently generated policies. You cannot measure \acs*{xp} within this isolated policy-search run; it is evaluated only after many independent policy-search runs have produced their policies. Use the available self-play rollouts to verify basic task competence and diagnose failures. Leave in \texttt{policy.py} the policy you judge most likely to achieve the highest expected cross-play return with independently generated partners. Use your analysis and available evaluation evidence to inform that judgment; the highest self-play score does not necessarily identify the best cross-play policy. Avoid relying on a private convention that succeeds only when both roles use this same policy.
    \end{policyinstructionbox}
    \caption{Objective and selection instructions for Overcooked V2. Note that these instructions are paraphrased across independent runs while preserving their meaning.}
    \label{fig:objective-instructions}
\end{figure}

\begin{figure}[htbp]
    \centering
    \begin{policyinstructionbox}
        \instructionheading{Objective}
        Maximize expected cooperative episode return when your local policy is paired with an independently generated partner policy, evaluated in both player positions. Your policy will be paired with independently generated partners. It should coordinate with them without assuming they use the same strategy, role assignments, or communication conventions. Seek competent, convention-robust behavior that can coordinate from locally observable evidence. Self-play return is a useful competence diagnostic, but it is not the optimization target.
        \tcblower
        \instructionheading{Selection rule}
        The primary evaluation metric is \acs*{xp}: cooperative episode return when this policy is paired with independently generated policies. You cannot measure \acs*{xp} within this isolated policy-search run; it is evaluated only after many independent policy-search runs have produced their policies. Use the available self-play rollouts to verify basic task competence and diagnose failures. Leave in \texttt{policy.py} the policy you judge most likely to achieve the highest expected cross-play return with independently generated partners. Use your analysis and available evaluation evidence to inform that judgment; the highest self-play score does not necessarily identify the best cross-play policy. Avoid relying on a private convention that succeeds only when both roles use this same policy.
    \end{policyinstructionbox}
    \caption{Objective and selection instructions for regular Hanabi and all Yokai configurations. Note that these instructions are paraphrased across independent runs while preserving their meaning.}
    \label{fig:objective-instructions-shared}
\end{figure}

\begin{figure}[htbp]
    \centering
    \begin{policyinstructionbox}
        \instructionheading{Objective}
        Maximize expected cooperative episode return when your policy is paired with independently generated policies that follow H-group Level 1 conventions, evaluated in both player positions. All partners receive the same pinned convention context. Seek competent play and robust coordination with other convention-following agents. Assume shared convention meanings, while allowing differences in aspects such as implementation, tactical choices, risk thresholds, and tie-breaking wherever the conventions leave choices open. Self-play return is a useful competence diagnostic, but the optimization target is cross-play performance with independently generated convention-following partners.
        \tcblower
        \instructionheading{Selection rule}
        The primary evaluation metric is \acs*{xp}: cooperative episode return when this policy is paired with independently generated policies that follow H-group Level 1 conventions. You cannot measure \acs*{xp} within this isolated policy-search run; it is evaluated only after many independent policy-search runs have produced their policies. Use the available self-play rollouts to verify basic task competence and diagnose failures. Leave in \texttt{policy.py} the policy you judge most likely to achieve the highest expected cross-play return with independently generated policies that follow H-group Level 1 conventions. Use your analysis and available evaluation evidence to inform that judgment; the highest self-play score does not necessarily identify the best cross-play policy. Avoid relying on a private convention that succeeds only when both roles use this same policy.
    \end{policyinstructionbox}
    \caption{Objective and selection instructions for Hanabi with H-group Level 1 conventions. Note that these instructions are paraphrased across independent runs while preserving their meaning.}
    \label{fig:objective-instructions-hgroup}
\end{figure}

% Source: meta_zsc/src/llm_policy/assets/config/llm_policy/objective/SP.yaml.
% Post-run selection: PolicyObjective.final_selection_instruction in context/objectives.py.
\begin{figure}[htbp]
    \centering
    \begin{policyinstructionbox}
        \instructionheading{Objective}
        Maximize only the expected self-play episode return.
        \tcblower
        \instructionheading{Selection rule}
        Leave in \texttt{policy.py} the policy with the highest authoritative development mean self-play episode return. Use minimum return and return variance only to diagnose or break ties between candidates with the same mean return.
    \end{policyinstructionbox}
    \caption{Self-play objective and selection instructions for Overcooked V2. Note that these instructions are paraphrased across independent runs while preserving their meaning.}
    \label{fig:objective-instructions-sp}
\end{figure}

\clearpage
\section{Novel Overcooked V2 Layouts}
\label{app:novel-ocv2-layouts}
\begin{center}
    \includegraphics[width=\mainDiagramWidth]{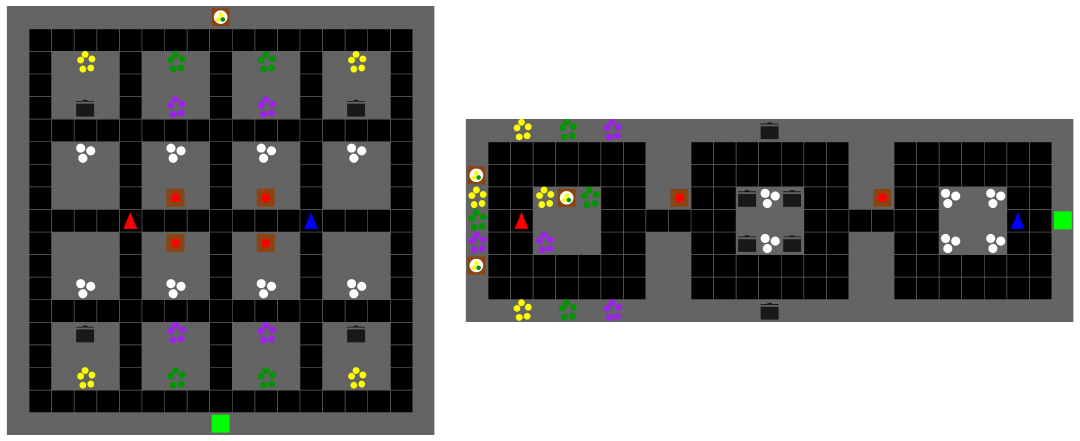}
\end{center}

\section{Full Overcooked V2 Results}
\label{app:overcooked-v2-results}

% Self-Play through Fictitious Co-Play source: Gessler et al. (2025), Table 2 (p. 9), official ICLR proceedings.
% MEP and GAMMA: author-supplied results, transcribed in overcooked_v2_sources.json.
% Coding-agent source: verified xp_matrix_index.json snapshot 2026-09-20T16:09:39.777886+00:00.
% Fable test-time source: audited six-seed extension 2026-09-23; see overcooked_v2_sources.json.
% Kimi/DeepSeek source: audited native/high matrices 2026-11-09; see overcooked_v2_sources.json.
% Source values and published baseline SDs are retained in overcooked_v2_sources.json.
% Coding-agent SDs describe dispersion across six policy-seed scores (ddof=1).
% IXP scores average partners and both roles per seed; these are not confidence intervals.
\begin{table}[htbp]
    \centering
    \caption{\textbf{Coordination performance on Overcooked V2.} Mean episode returns with the reported variation below each mean. For Self-Play through Fictitious Co-Play, \citet{gessler2025overcookedv2} report standard deviations over ten seeds with 500 games per pairing. Coding-agent results report standard deviations over six seeds with 1,024 games per pairing.}
    \label{tab:overcooked-v2-full}
    \begingroup
    \small
    \newcommand{\ocgroupgap}{14pt}
    \setlength{\tabcolsep}{3pt}
    \ifdim\linewidth<420pt
        \renewcommand{\ocgroupgap}{4pt}
        \setlength{\tabcolsep}{2.5pt}
    \fi
    \renewcommand{\arraystretch}{1.0}
    \setlength{\heavyrulewidth}{0.6pt}
    \setlength{\lightrulewidth}{0.35pt}
    \sisetup{mode=text, detect-weight=true, table-number-alignment=right}
    \newcommand{\ocvalue}[2]{%
        \shortstack[c]{\strut{\rmfamily\num[detect-family=true]{#1}}\\[-1pt]%
        {\scriptsize\rmfamily\mdseries\color{black!70}\ensuremath{\pm}\,\num[detect-family=true]{#2}}}}
    \colorlet{ocsp}{citationnavy}
    \definecolor{ocixp}{HTML}{9A5B34}
    \newcommand{\ocbestsp}[2]{\cellcolor{ocsp!15}\ocvalue{#1}{#2}}
    \newcommand{\ocbestixp}[2]{\cellcolor{ocixp!15}\ocvalue{#1}{#2}}
    \newcommand{\ocmetric}[1]{{\scriptsize\rmfamily\color{oc#1}\acs{#1}}}
    \begin{tabular}{@{}l@{\hspace{\ocgroupgap}}cc@{\hspace{\ocgroupgap}}cc@{\hspace{\ocgroupgap}}cc@{\hspace{\ocgroupgap}}cc@{\hspace{\ocgroupgap}}cc@{\hspace{\ocgroupgap}}cc@{}}
    \toprule
    & \multicolumn{4}{@{}c@{\hspace{\ocgroupgap}}}{\footnotesize Demo Cook} & \multicolumn{4}{@{}c@{\hspace{\ocgroupgap}}}{\footnotesize Grounded Coord.} & \multicolumn{4}{@{}c@{}}{\footnotesize Test-time} \\
    \cmidrule(l{0pt}r{\ocgroupgap}){2-5}\cmidrule(l{0pt}r{\ocgroupgap}){6-9}\cmidrule(l{0pt}r{0pt}){10-13}
    & \multicolumn{2}{@{}c@{\hspace{\ocgroupgap}}}{\footnotesize Simple} & \multicolumn{2}{@{}c@{\hspace{\ocgroupgap}}}{\footnotesize Wide} & \multicolumn{2}{@{}c@{\hspace{\ocgroupgap}}}{\footnotesize Simple} & \multicolumn{2}{@{}c@{\hspace{\ocgroupgap}}}{\footnotesize Ring} & \multicolumn{2}{@{}c@{\hspace{\ocgroupgap}}}{\footnotesize Simple} & \multicolumn{2}{@{}c@{}}{\footnotesize Wide} \\
    \cmidrule(l{0pt}r{\ocgroupgap}){2-3}\cmidrule(l{0pt}r{\ocgroupgap}){4-5}\cmidrule(l{0pt}r{\ocgroupgap}){6-7}\cmidrule(l{0pt}r{\ocgroupgap}){8-9}\cmidrule(l{0pt}r{\ocgroupgap}){10-11}\cmidrule(l{0pt}r{0pt}){12-13}
    {\footnotesize Algorithm} $\mathcal{L}$ & \ocmetric{sp} & \ocmetric{ixp} & \ocmetric{sp} & \ocmetric{ixp} & \ocmetric{sp} & \ocmetric{ixp} & \ocmetric{sp} & \ocmetric{ixp} & \ocmetric{sp} & \ocmetric{ixp} & \ocmetric{sp} & \ocmetric{ixp} \\[2pt]
    \midrule
    {\footnotesize Self-Play} & \ocvalue{172}{19} & \ocvalue{36}{82} & \ocvalue{195}{4} & \ocvalue{40}{61} & \ocvalue{154}{10} & \ocvalue{-55}{124} & \ocvalue{164}{10} & \ocvalue{-12}{75} & \ocvalue{145}{22} & \ocvalue{-81}{99} & \ocvalue{194}{12} & \ocvalue{-30}{96} \\
    \addlinespace[2pt]
    {\footnotesize State-Augmented} & \ocvalue{170}{16} & \ocvalue{112}{50} & \ocvalue{192}{7} & \ocvalue{60}{60} & \ocbestsp{156}{29} & \ocvalue{-17}{115} & \ocbestsp{165}{7} & \ocvalue{-16}{65} & \ocvalue{161}{18} & \ocvalue{-55}{103} & \ocvalue{137}{82} & \ocvalue{-12}{72} \\
    \addlinespace[2pt]
    {\footnotesize Other-Play} & \ocvalue{149}{25} & \ocvalue{47}{48} & \ocvalue{162}{23} & \ocvalue{62}{53} & \ocvalue{122}{24} & \ocvalue{4}{73} & \ocvalue{121}{36} & \ocvalue{-9}{21} & \ocvalue{121}{37} & \ocvalue{-3}{51} & \ocvalue{175}{54} & \ocvalue{-11}{42} \\
    \addlinespace[2pt]
    {\footnotesize Fictitious Co-Play} & \ocvalue{22}{48} & \ocvalue{1}{25} & \ocvalue{45}{43} & \ocvalue{13}{28} & \ocvalue{64}{58} & \ocvalue{46}{47} & \ocvalue{86}{34} & \ocvalue{6}{46} & \ocvalue{35}{44} & \ocvalue{6}{29} & \ocvalue{95}{56} & \ocvalue{23}{40} \\
    \addlinespace[2pt]
    {\footnotesize MEP~\citep{zhao2023mep}} & \ocvalue{1}{2} & \ocvalue{1}{1} & \ocvalue{10}{24} & \ocvalue{11}{10} & \ocvalue{2}{9} & \ocvalue{4}{2} & \ocvalue{82}{51} & \ocvalue{83}{20} & \ocvalue{7}{9} & \ocvalue{4}{2} & \ocvalue{12}{27} & \ocvalue{10}{7} \\
    \addlinespace[2pt]
    {\footnotesize GAMMA~\citep{liang2024generative}} & \ocvalue{4}{9} & \ocvalue{4}{3} & \ocvalue{13}{18} & \ocvalue{4}{5} & \ocvalue{9}{17} & \ocvalue{7}{6} & \ocvalue{23}{33} & \ocvalue{17}{16} & \ocvalue{4}{9} & \ocvalue{7}{7} & \ocvalue{24}{43} & \ocvalue{10}{12} \\
    \greytablerule{13}
    {\footnotesize GPT-5.6 Luna} & \ocvalue{77}{41} & \ocvalue{85}{10} & \ocvalue{112}{10} & \ocvalue{110}{6} & \ocvalue{116}{19} & \ocvalue{82}{13} & \ocvalue{105}{15} & \ocvalue{105}{6} & \ocvalue{54}{59} & \ocvalue{8}{27} & \ocvalue{67}{58} & \ocvalue{4}{16} \\
    \addlinespace[2pt]
    {\footnotesize Claude Sonnet 5} & \ocvalue{125}{22} & \ocvalue{101}{19} & \ocvalue{103}{45} & \ocvalue{91}{17} & \ocvalue{132}{13} & \ocvalue{130}{5} & \ocvalue{104}{51} & \ocvalue{101}{20} & \ocvalue{76}{42} & \ocvalue{57}{2} & \ocvalue{113}{31} & \ocvalue{73}{25} \\
    \addlinespace[2pt]
    {\footnotesize GPT-5.5} & \ocvalue{150}{11} & \ocvalue{125}{10} & \ocvalue{153}{17} & \ocvalue{133}{18} & \ocvalue{137}{8} & \ocvalue{101}{3} & \ocvalue{122}{12} & \ocvalue{122}{5} & \ocvalue{127}{77} & \ocvalue{33}{47} & \ocvalue{195}{17} & \ocvalue{195}{7} \\
    \addlinespace[2pt]
    {\footnotesize Kimi K3.1 Pro} & \ocvalue{174}{20} & \ocvalue{150}{9} & \ocvalue{178}{29} & \ocvalue{154}{15} & \ocvalue{140}{12} & \ocvalue{104}{17} & \ocvalue{137}{13} & \ocvalue{112}{23} & \ocvalue{149}{46} & \ocvalue{95}{45} & \ocvalue{188}{12} & \ocvalue{178}{13} \\
    \addlinespace[2pt]
    {\footnotesize DeepSeek V4.1 Flash} & \ocvalue{176}{17} & \ocvalue{174}{7} & \ocvalue{175}{13} & \ocvalue{151}{21} & \ocvalue{137}{14} & \ocvalue{116}{12} & \ocvalue{134}{13} & \ocvalue{130}{8} & \ocvalue{111}{56} & \ocvalue{53}{10} & \ocvalue{102}{55} & \ocvalue{83}{28} \\
    \addlinespace[2pt]
    {\footnotesize GPT-5.6 Sol} & \ocbestsp{207}{16} & \ocvalue{174}{27} & \ocvalue{185}{32} & \ocvalue{162}{29} & \ocvalue{149}{18} & \ocvalue{129}{9} & \ocvalue{152}{17} & \ocvalue{133}{14} & \ocbestsp{180}{1} & \ocvalue{155}{19} & \ocvalue{214}{17} & \ocvalue{202}{11} \\
    \addlinespace[2pt]
    {\footnotesize Claude Opus 5} & \ocvalue{198}{4} & \ocbestixp{197}{1} & \ocvalue{202}{4} & \ocvalue{193}{8} & \ocvalue{139}{8} & \ocvalue{130}{5} & \ocvalue{147}{4} & \ocvalue{130}{4} & \ocvalue{172}{6} & \ocvalue{168}{2} & \ocvalue{209}{11} & \ocvalue{120}{26} \\
    \addlinespace[2pt]
    {\footnotesize GPT-6 Astra} & \ocvalue{198}{3} & \ocvalue{191}{7} & \ocbestsp{213}{9} & \ocbestixp{211}{4} & \ocvalue{138}{4} & \ocbestixp{137}{1} & \ocvalue{152}{3} & \ocbestixp{150}{2} & \ocvalue{161}{12} & \ocvalue{155}{5} & \ocvalue{214}{11} & \ocvalue{177}{26} \\
    \addlinespace[2pt]
    {\footnotesize Claude Fable 5.1} & \ocvalue{200}{11} & \ocvalue{196}{1} & \ocvalue{212}{16} & \ocvalue{203}{5} & \ocvalue{134}{3} & \ocvalue{134}{1} & \ocvalue{158}{16} & \ocvalue{130}{7} & \ocvalue{174}{9} & \ocbestixp{172}{4} & \ocbestsp{231}{10} & \ocbestixp{218}{15} \\
    \bottomrule
    \end{tabular}
    % Pale navy and orange cells mark the highest SP and IXP per task, respectively, ranked before rounding.
    \endgroup
\end{table}

% Source: verified six-seed matrices for Block Grid and Pipeline Cook in
% meta_zsc/paper/RESULTS.md (2026-09-23); full-precision values and index
% checksums are retained in overcooked_v2_novel_sources.json.
% Coding-agent SDs use ddof=1 across six policy-level scores, as in the
% published-layout Overcooked V2 tables. SP uses diagonal cells; for seed i,
% IXP averages both roles with each of the other five seeds.
\begin{table}[htbp]
    \centering
    \caption{\textbf{Coordination performance on novel Overcooked V2 layouts.} Mean episode return, with $\pm$ SD shown below each mean. Coding-agent results use six seeds and 1,024 games per policy pairing.}
    \label{tab:overcooked-v2-novel}
    \begingroup
    \small
    \setlength{\tabcolsep}{3pt}
    \renewcommand{\arraystretch}{1.0}
    \setlength{\heavyrulewidth}{0.6pt}
    \setlength{\lightrulewidth}{0.35pt}
    \sisetup{mode=text, detect-weight=true, table-number-alignment=right}
    \newcommand{\ocvalue}[2]{%
        \shortstack[c]{\strut{\rmfamily\num[detect-family=true]{#1}}\\[-1pt]%
        {\scriptsize\rmfamily\mdseries\color{black!70}\ensuremath{\pm}\,\num[detect-family=true]{#2}}}}
    \colorlet{ocsp}{citationnavy}
    \definecolor{ocixp}{HTML}{9A5B34}
    \newcommand{\ocbestsp}[2]{\cellcolor{ocsp!15}\ocvalue{#1}{#2}}
    \newcommand{\ocbestixp}[2]{\cellcolor{ocixp!15}\ocvalue{#1}{#2}}
    \newcommand{\ocmetric}[1]{{\scriptsize\rmfamily\color{oc#1}\acs{#1}}}
    \begin{tabular}{@{}l@{\hspace{16pt}}>{\centering\arraybackslash}p{25pt}>{\centering\arraybackslash}p{25pt}@{\hspace{16pt}}>{\centering\arraybackslash}p{25pt}>{\centering\arraybackslash}p{25pt}@{}}
    \toprule
    & \multicolumn{2}{@{}c@{\hspace{16pt}}}{\footnotesize Block Grid} & \multicolumn{2}{@{}c@{}}{\footnotesize Pipeline Cook} \\
    \cmidrule(l{0pt}r{16pt}){2-3}\cmidrule(l{0pt}r{0pt}){4-5}
    {\footnotesize Algorithm} $\mathcal{L}$ & \ocmetric{sp} & \ocmetric{ixp} & \ocmetric{sp} & \ocmetric{ixp} \\[2pt]
    \midrule
    {\footnotesize GPT-6 Sol} & \ocvalue{51}{14} & \ocvalue{45}{5} & \ocvalue{64}{6} & \ocvalue{59}{5} \\
    \addlinespace[2pt]
    {\footnotesize Claude Opus 5.5} & \ocvalue{67}{8} & \ocbestixp{69}{3} & \ocvalue{77}{3} & \ocvalue{73}{1} \\
    \addlinespace[2pt]
    {\footnotesize GPT-6 Astra} & \ocvalue{74}{12} & \ocvalue{66}{6} & \ocbestsp{81}{3} & \ocbestixp{77}{3} \\
    \addlinespace[2pt]
    {\footnotesize Claude Fable 5.1} & \ocbestsp{76}{12} & \ocvalue{64}{4} & \ocvalue{80}{2} & \ocvalue{75}{3} \\
    \bottomrule
    \end{tabular}
    % Pale navy and orange cells mark the highest SP and IXP per task, respectively, ranked before rounding.
    \endgroup
\end{table}

\clearpage
\section{Full Yokai Results}
\label{app:yokai-results}

% Baseline source: Ruhdorfer et al. (2026), Tables 2 and 6 (PDF pp. 9--10).
% Exact coding-agent means, six seed aggregates, and SEMs: yokai_sources.json.
% All means and SEMs use one decimal, rounded from full-precision source values.
% Highlighting compares full-precision coding-agent means with published baseline means.
\begin{table}[htbp]
    \centering
    \caption{\textbf{Coordination performance on two-player Yokai.} Mean episode return $\pm$ SEM. Coding-agent results use six seeds and 1,024 games per policy pairing; the 12-seed method baselines are taken directly from \citet{ruhdorfer2026yokai}, who evaluated them on 5,000 games. They also report an illustrative human-human baseline of \(5.7 \pm 0.5\) across 25 games in the nine-card imperfect-memory setting.}
    \label{tab:yokai-full}
    \begingroup
    \small
    \setlength{\tabcolsep}{1pt}
    \renewcommand{\arraystretch}{1.0}
    \setlength{\heavyrulewidth}{0.6pt}
    \setlength{\lightrulewidth}{0.35pt}
    \sisetup{mode=text, table-number-alignment=right}
    \newcommand{\ykvalue}[2]{%
        \mbox{{\rmfamily\num[detect-family=true]{#1}}%
        {\scriptsize\rmfamily\mdseries\color{black!70}\hspace{0.2em}\ensuremath{\pm}\hspace{0.2em}\num[detect-family=true]{#2}}}}
    \newcommand{\ykbestsp}[2]{\cellcolor{citationnavy!15}\ykvalue{#1}{#2}}
    \newcommand{\ykbestixp}[2]{\cellcolor{citationnavy!15}\ykvalue{#1}{#2}}
    \begin{tabular}{@{}l@{\hspace{6pt}}l@{\hspace{8pt}}r@{\hspace{8pt}}r@{}}
    \toprule
    {\footnotesize Configuration} & {\footnotesize Algorithm} $\mathcal{L}$ & \multicolumn{1}{c}{\footnotesize\acs{sp}} & \multicolumn{1}{c}{\footnotesize\acs{ixp}} \\[2pt]
    \midrule
    \multirow{15}{*}{\footnotesize\shortstack[l]{9 cards\\Perfect memory}} & {\footnotesize IPPO} & \ykbestsp{7.5}{0.3} & \ykvalue{1.6}{0.2} \\
     & {\footnotesize IPPO+High Ent.} & \ykbestsp{7.5}{0.1} & \ykvalue{4.1}{0.8} \\
     & {\footnotesize Other-Play} & \ykvalue{6.6}{0.2} & \ykvalue{4.8}{0.1} \\
     & {\footnotesize OBL L1} & \ykvalue{4.0}{0.2} & \ykvalue{2.3}{0.3} \\
     & {\footnotesize OBL L2} & \ykvalue{4.3}{0.2} & \ykvalue{2.1}{0.2} \\
     & {\footnotesize OBL L3} & \ykvalue{4.4}{0.4} & \ykvalue{2.2}{0.3} \\
     & {\footnotesize OBL L4} & \ykvalue{4.5}{0.2} & \ykvalue{2.8}{0.3} \\
     & {\footnotesize OBL L5} & \ykvalue{4.8}{0.2} & \ykvalue{2.5}{0.3} \\
    \greytableruleafterfirst{3}
     & {\footnotesize GPT-5.6 Luna} & \ykvalue{1.5}{0.5} & \ykvalue{0.7}{0.1} \\
     & {\footnotesize Claude Sonnet 5} & \ykvalue{3.5}{0.1} & \ykvalue{3.6}{0.0} \\
     & {\footnotesize GPT-5.5} & \ykvalue{3.4}{0.2} & \ykvalue{2.0}{0.3} \\
     & {\footnotesize GPT-5.6 Sol} & \ykvalue{4.9}{0.3} & \ykvalue{4.7}{0.1} \\
     & {\footnotesize Claude Opus 5} & \ykvalue{6.0}{0.5} & \ykvalue{5.6}{0.3} \\
     & {\footnotesize GPT-6 Astra} & \ykvalue{7.5}{0.5} & \ykbestixp{7.3}{0.2} \\
     & {\footnotesize Claude Fable 5.1} & \ykvalue{7.0}{0.3} & \ykvalue{7.0}{0.1} \\
    \midrule
    \multirow{9}{*}{\footnotesize\shortstack[l]{9 cards\\Imperfect memory}} & {\footnotesize Other-Play} & \ykvalue{0.0}{0.0} & \ykvalue{0.0}{0.0} \\
     & {\footnotesize Other-Play (TrXL)} & \ykvalue{0.3}{0.1} & \ykvalue{0.3}{0.1} \\
    \greytableruleafterfirst{3}
     & {\footnotesize GPT-5.6 Luna} & \ykvalue{0.7}{0.4} & \ykvalue{0.3}{0.1} \\
     & {\footnotesize Claude Sonnet 5} & \ykvalue{3.5}{0.1} & \ykvalue{3.4}{0.1} \\
     & {\footnotesize GPT-5.5} & \ykvalue{3.7}{0.2} & \ykvalue{2.6}{0.1} \\
     & {\footnotesize GPT-5.6 Sol} & \ykvalue{4.9}{0.5} & \ykvalue{4.3}{0.1} \\
     & {\footnotesize Claude Opus 5} & \ykvalue{5.2}{0.5} & \ykvalue{5.2}{0.2} \\
     & {\footnotesize GPT-6 Astra} & \ykvalue{6.3}{0.5} & \ykvalue{6.5}{0.2} \\
     & {\footnotesize Claude Fable 5.1} & \ykbestsp{7.1}{0.3} & \ykbestixp{6.9}{0.2} \\
    \midrule
    \multirow{8}{*}{\footnotesize\shortstack[l]{16 cards\\Perfect memory}} & {\footnotesize IPPO} & \ykvalue{0.0}{0.0} & \ykvalue{0.0}{0.0} \\
    \greytableruleafterfirst{3}
     & {\footnotesize GPT-5.6 Luna} & \ykvalue{0.4}{0.2} & \ykvalue{0.0}{0.0} \\
     & {\footnotesize Claude Sonnet 5} & \ykvalue{1.8}{0.7} & \ykvalue{0.8}{0.1} \\
     & {\footnotesize GPT-5.5} & \ykvalue{2.2}{0.4} & \ykvalue{0.4}{0.1} \\
     & {\footnotesize GPT-5.6 Sol} & \ykvalue{6.3}{0.4} & \ykvalue{5.1}{0.2} \\
     & {\footnotesize Claude Opus 5} & \ykvalue{6.1}{0.6} & \ykvalue{4.2}{0.7} \\
     & {\footnotesize GPT-6 Astra} & \ykvalue{8.2}{0.9} & \ykbestixp{8.0}{0.4} \\
     & {\footnotesize Claude Fable 5.1} & \ykbestsp{8.4}{0.2} & \ykvalue{8.0}{0.1} \\
    \bottomrule
    \end{tabular}
    % Pale navy cell fills mark the highest reported scores, not statistical significance.
    \endgroup
\end{table}

\clearpage
\section{Full Hanabi Results}
\label{app:hanabi-results}

% Exact values, seed aggregates, hashes, and reporting definitions: hanabi_sources.json.
% Existing-method baselines use four checkpoint seeds; coding agents use six independent generations.
\begin{table*}[htbp]
    \centering
    \caption{\textbf{Coordination performance on two-player Hanabi.} Mean $\pm$ SEM over six coding-agent runs, six hybrid policies, and four checkpoints per baseline (hybrid \acs{ixp} uses jackknife SE), with 1,024 games per policy pairing or policy-bot seat assignment. We evaluate R2D2-based OBL checkpoints from \citet{hu2021offbelief} and IPPO, IPPO+HE, and IPPO-based OBL checkpoints from \citet{forkel2026entropy}.}
    \label{tab:hanabi-full}
    \begingroup
    \small
    \setlength{\tabcolsep}{3pt}
    \renewcommand{\arraystretch}{1.0}
    \setlength{\heavyrulewidth}{0.6pt}
    \setlength{\lightrulewidth}{0.35pt}
    \sisetup{mode=text, table-number-alignment=right}
    \newcommand{\hnvalue}[2]{%
        {\rmfamily\makebox[3.0em][r]{\num[detect-family=true]{#1}}}%
        {\scriptsize\rmfamily\mdseries\hspace{0.1em}\ensuremath{\pm}\hspace{0.1em}\makebox[2.7em][l]{\num[detect-family=true]{#2}}}}
    \newcommand{\hnbestsp}[2]{\cellcolor{citationnavy!15}\hnvalue{#1}{#2}}
    \newcommand{\hnbestixp}[2]{\cellcolor{citationnavy!15}\hnvalue{#1}{#2}}
    \newcommand{\hnbestbc}[2]{\cellcolor{citationnavy!15}\hnvalue{#1}{#2}}
    \newcommand{\hnbestwillhanabibot}[2]{\cellcolor{citationnavy!15}\hnvalue{#1}{#2}}
    \begin{tabular}{@{}llrrrr@{}}
    \toprule
    {\footnotesize Algorithm} $\mathcal{L}$ & {\footnotesize Setting} & \multicolumn{1}{c}{\footnotesize\acs{sp}} & \multicolumn{1}{c}{\footnotesize\acs{ixp}} & \multicolumn{1}{c}{\footnotesize\shortstack{AH2AC2 bot \acs{xp}\\(\acs{sp}: 23.80)}} & \multicolumn{1}{c}{\footnotesize\shortstack{Will-Hanabi bot \acs{xp}\\(\acs{sp}: 21.05)}} \\[2pt]
    \midrule
    {\footnotesize IPPO} & {\footnotesize Standard} & \hnbestsp{24.53}{0.02} & \hnvalue{15.63}{2.13} & \hnvalue{13.60}{1.54} & \hnvalue{3.96}{0.48} \\
    {\footnotesize IPPO+HE} & {\footnotesize Standard} & \hnvalue{24.48}{0.01} & \hnbestixp{24.43}{0.02} & \hnvalue{11.64}{0.09} & \hnvalue{4.41}{0.08} \\
    {\footnotesize OBL L4 (R2D2)} & {\footnotesize Standard} & \hnvalue{24.12}{0.03} & \hnvalue{23.76}{0.08} & \hnvalue{19.70}{0.16} & \hnvalue{5.19}{0.04} \\
    {\footnotesize OBL L6 (IPPO)} & {\footnotesize Standard} & \hnvalue{24.35}{0.01} & \hnvalue{24.32}{0.00} & \hnvalue{14.92}{0.15} & \hnvalue{5.56}{0.13} \\
    \greytablerule{6}
    \multirow{2}{*}{\footnotesize GPT-5.6 Luna} & {\footnotesize Standard} & \hnvalue{13.82}{1.05} & \hnvalue{13.19}{0.54} & \hnvalue{4.88}{1.56} & \hnvalue{1.80}{0.56} \\
     & {\footnotesize H-group L1} & \hnvalue{1.54}{1.01} & \hnvalue{0.93}{0.28} & \hnvalue{1.36}{0.56} & \hnvalue{1.28}{0.36} \\
    \addlinespace[1pt]
    \multirow{2}{*}{\footnotesize Claude Sonnet 5} & {\footnotesize Standard} & \hnvalue{16.80}{1.22} & \hnvalue{15.60}{0.61} & \hnvalue{11.08}{3.23} & \hnvalue{6.02}{1.73} \\
     & {\footnotesize H-group L1} & \hnvalue{11.47}{1.40} & \hnvalue{10.49}{0.87} & \hnvalue{13.51}{1.47} & \hnvalue{12.79}{0.81} \\
    \addlinespace[1pt]
    \multirow{2}{*}{\footnotesize GPT-5.5} & {\footnotesize Standard} & \hnvalue{16.76}{0.76} & \hnvalue{12.97}{0.85} & \hnvalue{13.68}{1.79} & \hnvalue{7.13}{1.16} \\
     & {\footnotesize H-group L1} & \hnvalue{11.40}{2.45} & \hnvalue{7.42}{1.09} & \hnvalue{8.88}{1.89} & \hnvalue{10.05}{1.39} \\
    \addlinespace[1pt]
    \multirow{2}{*}{\footnotesize GPT-5.6 Sol} & {\footnotesize Standard} & \hnvalue{21.93}{0.19} & \hnvalue{18.74}{0.42} & \hnvalue{15.53}{2.25} & \hnvalue{7.42}{1.15} \\
     & {\footnotesize H-group L1} & \hnvalue{18.59}{0.65} & \hnvalue{18.26}{0.29} & \hnvalue{16.92}{0.57} & \hnvalue{18.54}{0.47} \\
    \addlinespace[1pt]
    \multirow{2}{*}{\footnotesize Claude Opus 5} & {\footnotesize Standard} & \hnvalue{20.00}{0.36} & \hnvalue{19.35}{0.12} & \hnvalue{18.60}{0.36} & \hnvalue{12.38}{0.63} \\
     & {\footnotesize H-group L1} & \hnvalue{20.05}{0.42} & \hnvalue{19.09}{0.20} & \hnvalue{17.75}{0.24} & \hnvalue{18.21}{0.63} \\
    \addlinespace[1pt]
    \multirow{2}{*}{\footnotesize GPT-6 Astra} & {\footnotesize Standard} & \hnvalue{19.69}{0.80} & \hnvalue{19.43}{0.35} & \hnvalue{17.02}{1.61} & \hnvalue{12.09}{1.55} \\
     & {\footnotesize H-group L1} & \hnvalue{21.53}{0.34} & \hnvalue{21.11}{0.19} & \hnvalue{17.87}{0.33} & \hnvalue{19.98}{0.25} \\
    \addlinespace[1pt]
    \multirow{2}{*}{\footnotesize Claude Fable 5.1} & {\footnotesize Standard} & \hnvalue{20.55}{0.26} & \hnvalue{19.50}{0.16} & \hnvalue{19.34}{0.68} & \hnvalue{14.87}{0.99} \\
     & {\footnotesize H-group L1} & \hnvalue{21.81}{0.21} & \hnvalue{20.80}{0.12} & \hnvalue{18.17}{0.40} & \hnvalue{20.25}{0.28} \\
    \greytablerule{6}
    {\footnotesize Hybrid (Astra)} & {\footnotesize H-group L1} & \hnvalue{24.28}{0.01} & \hnvalue{24.14}{0.05} & \hnbestbc{21.75}{0.18} & \hnbestwillhanabibot{21.97}{0.11} \\
    {\footnotesize Hybrid (Fable)} & {\footnotesize H-group L1} & \hnvalue{24.31}{0.02} & \hnvalue{23.98}{0.21} & \hnvalue{20.98}{0.23} & \hnvalue{21.69}{0.23} \\
    \bottomrule
    \end{tabular}
    \vspace{2pt}

    % {\scriptsize AH2AC2 bot XP averages returns with the two behavioural-cloning bots, BC1 and BC2; all partner scores average both policy seat assignments. The \acs{ixp} aggregates share partners. H-group L1 denotes the policy-generation setting; all policies use standard Hanabi scoring.}
    % Header SP values: mean of BC1 x BC1 and BC2 x BC2, and the Will-Hanabi bot paired with itself, respectively; 1,024 games per pairing.
    % Pale navy cell fills mark the highest mean before rounding, not statistical significance.
    \endgroup
\end{table*}

\clearpage
\section{3--5 Player Hanabi}
\label{app:multiplayer-hanabi}

\begin{table}[htbp]
\centering
\caption{\textbf{Coordination performance on three- to five-player Hanabi.} Mean $\pm$ SEM over six coding-agent runs, six hybrid policies, and four IPPO checkpoints, with 1,024 games per policy pairing. Across all player counts, IPPO achieves the highest \acs{sp} and \acs{ixp} returns, while coding agents given H-group Level 1 conventions and the resulting hybrid policies achieve higher \acs{xp} returns with the Will-Hanabi bot.}
\label{tab:multiplayer-hanabi}
\begingroup
\footnotesize
\setlength{\tabcolsep}{3pt}
\newcommand{\mhscore}[2]{\ensuremath{#1\,{\scriptstyle\pm\,#2}}}
\begin{tabular}{@{}llrrr@{}}
\toprule
Algorithm $\mathcal{L}$ & Setting & \acs{sp} & \acs{ixp} & Will-Hanabi bot \acs{xp} \\
\midrule
\multicolumn{5}{@{}l}{\textit{3 players}} \\[3pt]
IPPO & Standard & \cellcolor{citationnavy!15}\mhscore{24.66}{0.00} & \cellcolor{citationnavy!15}\mhscore{24.59}{0.02} & \mhscore{7.18}{0.12} \\
\greytablerule{5}
\multirow{2}{*}{GPT-6 Astra} & Standard & \mhscore{20.23}{0.12} & \mhscore{20.19}{0.03} & \mhscore{11.94}{0.92} \\
 & H-group L1 & \mhscore{22.50}{0.11} & \mhscore{22.41}{0.03} & \mhscore{22.72}{0.08} \\
\addlinespace[1pt]
\multirow{2}{*}{GPT-6.1 Sol} & Standard & \mhscore{20.22}{0.21} & \mhscore{20.08}{0.02} & \mhscore{12.17}{1.18} \\
 & H-group L1 & \mhscore{22.28}{0.08} & \mhscore{22.20}{0.02} & \mhscore{22.54}{0.11} \\
\greytablerule{5}
Hybrid (Astra) & H-group L1 & \mhscore{24.38}{0.01} & \mhscore{24.34}{0.01} & \cellcolor{citationnavy!15}\mhscore{22.91}{0.08} \\
Hybrid (Sol 6.1) & H-group L1 & \mhscore{24.34}{0.02} & \mhscore{24.30}{0.02} & \mhscore{22.86}{0.11} \\
\midrule
\multicolumn{5}{@{}l}{\textit{4 players}} \\[3pt]
IPPO & Standard & \cellcolor{citationnavy!15}\mhscore{24.54}{0.01} & \cellcolor{citationnavy!15}\mhscore{24.43}{0.03} & \mhscore{8.85}{0.22} \\
\greytablerule{5}
\multirow{2}{*}{GPT-6 Astra} & Standard & \mhscore{19.84}{0.19} & \mhscore{19.71}{0.02} & \mhscore{14.34}{0.85} \\
 & H-group L1 & \mhscore{22.55}{0.06} & \mhscore{22.45}{0.02} & \cellcolor{citationnavy!15}\mhscore{22.40}{0.13} \\
\addlinespace[1pt]
\multirow{2}{*}{GPT-6.1 Sol} & Standard & \mhscore{20.00}{0.23} & \mhscore{19.77}{0.02} & \mhscore{13.72}{0.69} \\
 & H-group L1 & \mhscore{22.39}{0.10} & \mhscore{22.32}{0.02} & \mhscore{22.29}{0.08} \\
\greytablerule{5}
Hybrid (Astra) & H-group L1 & \mhscore{24.14}{0.02} & \mhscore{24.10}{0.02} & \mhscore{22.28}{0.10} \\
Hybrid (Sol 6.1) & H-group L1 & \mhscore{24.12}{0.02} & \mhscore{24.06}{0.01} & \mhscore{22.30}{0.05} \\
\midrule
\multicolumn{5}{@{}l}{\textit{5 players}} \\[3pt]
IPPO & Standard & \cellcolor{citationnavy!15}\mhscore{23.70}{0.03} & \cellcolor{citationnavy!15}\mhscore{23.60}{0.03} & \mhscore{11.94}{0.11} \\
\greytablerule{5}
\multirow{2}{*}{GPT-6 Astra} & Standard & \mhscore{18.99}{0.23} & \mhscore{18.92}{0.02} & \mhscore{15.23}{0.29} \\
 & H-group L1 & \mhscore{21.75}{0.10} & \mhscore{21.64}{0.06} & \cellcolor{citationnavy!15}\mhscore{20.91}{0.17} \\
\addlinespace[1pt]
\multirow{2}{*}{GPT-6.1 Sol} & Standard & \mhscore{19.16}{0.21} & \mhscore{19.09}{0.01} & \mhscore{15.49}{0.31} \\
 & H-group L1 & \mhscore{21.60}{0.11} & \mhscore{21.56}{0.01} & \mhscore{20.90}{0.06} \\
\greytablerule{5}
Hybrid (Astra) & H-group L1 & \mhscore{23.16}{0.04} & \mhscore{23.03}{0.06} & \mhscore{20.03}{0.20} \\
Hybrid (Sol 6.1) & H-group L1 & \mhscore{23.14}{0.04} & \mhscore{23.04}{0.03} & \mhscore{20.06}{0.18} \\
\bottomrule
\end{tabular}
\endgroup
\end{table}

\begin{table}[htbp]
\centering
\caption{\textbf{Finesse play in three-player Hanabi.} Mean $\pm$ SEM across policies, with 1,000 deals per policy. \textit{Eligible turns} is the percentage of all turns on which a single-touch finesse was available. Of those opportunities, \textit{Initiated} is the percentage in which the acting player gave the required colour hint, and \textit{Executed} is the percentage in which the hint led the next player to play the connecting card and the hinted player to play the newly playable card on the following turn.}
\label{tab:hanabi-finesse}
\begingroup
\footnotesize
\setlength{\tabcolsep}{3pt}
\newcommand{\finesserate}[2]{\ensuremath{#1\,{\scriptstyle\pm\,#2}}}
\begin{tabular}{@{}llrrr@{}}
\toprule
Algorithm $\mathcal{L}$ & Setting & Eligible turns (\%) & Initiated (\%) & Executed (\%) \\
\midrule
IPPO & Standard & \finesserate{0.18}{0.00} & \finesserate{0.00}{0.00} & \finesserate{0.00}{0.00} \\
\greytablerule{5}
\multirow{2}{*}{GPT-6 Astra} & Standard & \finesserate{0.25}{0.02} & \finesserate{0.35}{0.16} & \finesserate{0.00}{0.00} \\
 & H-group L1 & \finesserate{0.39}{0.03} & \finesserate{43.04}{5.06} & \finesserate{32.42}{3.03} \\
\addlinespace[1pt]
\multirow{2}{*}{GPT-6.1 Sol} & Standard & \finesserate{0.27}{0.02} & \finesserate{0.00}{0.00} & \finesserate{0.00}{0.00} \\
 & H-group L1 & \finesserate{0.38}{0.03} & \finesserate{57.77}{8.87} & \finesserate{43.19}{7.27} \\
\greytablerule{5}
Hybrid (Astra) & H-group L1 & \finesserate{0.39}{0.02} & \finesserate{58.50}{3.52} & \finesserate{44.58}{2.39} \\
Hybrid (Sol 6.1) & H-group L1 & \finesserate{0.39}{0.02} & \finesserate{71.18}{6.89} & \finesserate{54.56}{6.02} \\
\midrule
Will-Hanabi bot & H-group L1 & 0.35 & 56.50 & 48.00 \\
\bottomrule
\end{tabular}
\endgroup
\end{table}

\clearpage
\section{Policy-Generation Settings}
\label{app:policy-generation-settings}
We compare three policy generation settings to assess the contributions of the coding harness and iterative feedback. Following \citet{hayler2026zeroshot}, we evaluate single-shot generation, where all information available to the coding agent at the start of generation (Figure~\ref{fig:coding-agent-learning}) is concatenated into a single prompt followed by a single policy generation step. We compare this with the first policy submitted for evaluation by the coding agent, which may use multiple tool calls to read and write files before submission but has not yet received evaluation feedback. Finally, we also compare against the policy submitted after iterative refinement using feedback from evaluations.

From Table~\ref{tab:policy-generation-settings}, we find that iterative refinement improves mean \acs{ixp} over the first attempt in all 30 model--environment comparisons and over single-shot generation in 25 of 30 comparisons. By contrast, the first attempt outperforms single-shot generation in only eight comparisons, suggesting that access to the coding harness alone does not consistently improve coordination.

In the single-shot generation setting, GPT-6 Astra achieves higher mean \acs{ixp} than the strongest deep \ac{rl} baseline in nine of ten environments. Together, these results suggest that the stronger \ac{zsc} performance of coding agents relative to \citet{hayler2026zeroshot} reflects both advances in the capabilities of the underlying \acp{llm} and the use of a coding harness that enables iterative refinement with rollout feedback.

% Generated by meta_zsc/paper/evaluations/generation_settings_20261012/build_table.py.
% Exact scores, policy identities, audit paths and hashes: generation_settings_sources.json.
% Below each mean: SEM (sample SD / sqrt(6)) across six policy seeds; IXP averages partners and both roles.
\begin{table}[!htbp]
    \centering
    \caption{\textbf{Coordination performance across different generation settings.} Each cell represents the mean $\pm$ SEM across  six seeds.}
    \label{tab:policy-generation-settings}
    \begingroup
    \footnotesize
    \setlength{\tabcolsep}{3pt}
    \renewcommand{\arraystretch}{1.05}
    \colorlet{gssp}{citationnavy}
    \definecolor{gsixp}{HTML}{9A5B34}
    \newcommand{\gsvalue}[2]{%
        \shortstack[c]{{\small\rmfamily\num[detect-family=true]{#1}}\\[-1pt]%
        {\scriptsize\rmfamily\mdseries\color{black!70}\ensuremath{\pm}\,\num[detect-family=true]{#2}}}}
    \newcommand{\gsbestsp}[2]{\cellcolor{gssp!15}\gsvalue{#1}{#2}}
    \newcommand{\gsbestixp}[2]{\cellcolor{gsixp!15}\gsvalue{#1}{#2}}
    \newcommand{\gsmetric}[1]{{\color{gs#1}\acs{#1}}}
    \newcommand{\gsenvironmentrule}{\noalign{\vskip1pt
        \hbox to \linewidth{\greytableleaders}%
        \vskip1.5pt}}
    \sisetup{mode=text, detect-weight=true}
    \begin{tabular*}{\linewidth}{@{\extracolsep{\fill}}llcc@{\hspace{14pt}}cc@{\hspace{14pt}}cc@{}}
    \toprule
    & & \multicolumn{2}{c@{\hspace{14pt}}}{GPT-5.6 Luna} & \multicolumn{2}{c@{\hspace{14pt}}}{GPT-5.6 Sol} & \multicolumn{2}{c}{GPT-6 Astra} \\
    \cmidrule(l{3pt}r{14pt}){3-4}\cmidrule(l{3pt}r{14pt}){5-6}\cmidrule(l{3pt}){7-8}
    Environment & Setting & \multicolumn{1}{c}{\gsmetric{sp}} & \multicolumn{1}{c@{\hspace{14pt}}}{\gsmetric{ixp}} & \multicolumn{1}{c}{\gsmetric{sp}} & \multicolumn{1}{c@{\hspace{14pt}}}{\gsmetric{ixp}} & \multicolumn{1}{c}{\gsmetric{sp}} & \multicolumn{1}{c@{}}{\gsmetric{ixp}} \\
    \midrule
    Demo Cook Simple & Single Shot & \gsvalue{73.24}{23.44} & \gsvalue{55.60}{12.95} & \gsvalue{95.30}{25.51} & \gsvalue{87.75}{10.44} & \gsvalue{146.41}{30.42} & \gsvalue{151.43}{11.90} \\[0.5pt]
     & First Attempt & \gsvalue{10.22}{7.17} & \gsvalue{8.64}{2.81} & \gsvalue{130.80}{8.87} & \gsvalue{104.59}{2.03} & \gsvalue{148.27}{16.29} & \gsvalue{148.00}{6.57} \\[0.5pt]
     & Refined & \gsbestsp{76.69}{16.68} & \gsbestixp{85.09}{3.88} & \gsbestsp{206.67}{6.67} & \gsbestixp{173.92}{11.08} & \gsbestsp{197.97}{1.08} & \gsbestixp{190.76}{2.68} \\[0.5pt]
    \gsenvironmentrule
    Demo Cook Wide & Single Shot & \gsvalue{-0.16}{16.16} & \gsvalue{6.65}{4.29} & \gsvalue{91.30}{31.84} & \gsvalue{76.32}{12.90} & \gsvalue{193.61}{5.11} & \gsvalue{185.66}{3.48} \\[0.5pt]
     & First Attempt & \gsvalue{23.33}{19.61} & \gsvalue{14.43}{5.65} & \gsvalue{110.09}{22.51} & \gsvalue{80.58}{7.36} & \gsvalue{145.19}{21.05} & \gsvalue{149.47}{7.55} \\[0.5pt]
     & Refined & \gsbestsp{112.40}{4.02} & \gsbestixp{109.79}{2.39} & \gsbestsp{184.65}{13.20} & \gsbestixp{162.37}{11.77} & \gsbestsp{212.96}{3.67} & \gsbestixp{210.51}{1.82} \\[0.5pt]
    \gsenvironmentrule
    Grounded Coord. Simple & Single Shot & \gsvalue{7.43}{25.53} & \gsvalue{3.01}{8.25} & \gsvalue{130.00}{0.00} & \gsbestixp{130.00}{0.00} & \gsvalue{131.08}{1.70} & \gsvalue{132.40}{0.52} \\[0.5pt]
     & First Attempt & \gsvalue{-294.15}{308.05} & \gsvalue{-301.60}{127.18} & \gsvalue{126.21}{3.79} & \gsvalue{122.04}{3.52} & \gsvalue{128.33}{1.67} & \gsvalue{125.67}{1.91} \\[0.5pt]
     & Refined & \gsbestsp{116.00}{7.90} & \gsbestixp{81.51}{5.50} & \gsbestsp{148.73}{7.29} & \gsvalue{128.69}{3.72} & \gsbestsp{137.94}{1.43} & \gsbestixp{137.06}{0.56} \\[0.5pt]
    \gsenvironmentrule
    Grounded Coord. Ring & Single Shot & \gsvalue{-346.93}{329.61} & \gsvalue{-346.04}{132.00} & \gsvalue{125.89}{2.62} & \gsvalue{111.30}{6.63} & \gsvalue{138.97}{6.39} & \gsvalue{137.60}{2.40} \\[0.5pt]
     & First Attempt & \gsvalue{31.31}{27.26} & \gsvalue{32.67}{10.76} & \gsvalue{122.57}{2.53} & \gsvalue{122.72}{1.08} & \gsvalue{103.11}{21.22} & \gsvalue{103.11}{8.49} \\[0.5pt]
     & Refined & \gsbestsp{105.37}{6.21} & \gsbestixp{105.37}{2.48} & \gsbestsp{151.63}{6.86} & \gsbestixp{132.51}{5.86} & \gsbestsp{151.94}{1.26} & \gsbestixp{149.96}{0.73} \\[0.5pt]
    \gsenvironmentrule
    Test-time Simple & Single Shot & \gsvalue{26.25}{11.80} & \gsbestixp{26.25}{4.72} & \gsvalue{93.19}{49.68} & \gsvalue{69.49}{18.65} & \gsvalue{116.81}{26.66} & \gsvalue{118.54}{8.39} \\[0.5pt]
     & First Attempt & \gsvalue{-5.97}{17.77} & \gsvalue{-5.97}{7.11} & \gsvalue{134.50}{15.41} & \gsvalue{101.36}{20.31} & \gsvalue{153.33}{4.22} & \gsvalue{153.22}{1.65} \\[0.5pt]
     & Refined & \gsbestsp{54.28}{23.89} & \gsvalue{7.58}{11.15} & \gsbestsp{180.21}{0.21} & \gsbestixp{155.48}{7.63} & \gsbestsp{160.75}{5.08} & \gsbestixp{154.60}{2.23} \\[0.5pt]
    \gsenvironmentrule
    Test-time Wide & Single Shot & \gsvalue{35.62}{21.03} & \gsbestixp{13.63}{3.55} & \gsvalue{101.20}{38.49} & \gsvalue{101.07}{15.41} & \gsvalue{189.67}{16.40} & \gsbestixp{185.32}{8.77} \\[0.5pt]
     & First Attempt & \gsvalue{19.79}{35.91} & \gsvalue{-8.34}{9.83} & \gsvalue{171.23}{4.25} & \gsvalue{146.62}{11.05} & \gsvalue{176.47}{3.96} & \gsvalue{35.54}{19.04} \\[0.5pt]
     & Refined & \gsbestsp{66.51}{23.77} & \gsvalue{4.23}{6.51} & \gsbestsp{214.13}{6.93} & \gsbestixp{201.80}{4.31} & \gsbestsp{213.72}{4.44} & \gsvalue{177.08}{10.61} \\[0.5pt]
    \gsenvironmentrule
    Hanabi & Single Shot & \gsvalue{9.33}{2.28} & \gsvalue{6.98}{1.29} & \gsvalue{17.45}{0.94} & \gsvalue{15.66}{0.80} & \gsvalue{19.23}{0.33} & \gsvalue{18.91}{0.15} \\[0.5pt]
     & First Attempt & \gsvalue{6.62}{2.41} & \gsvalue{5.51}{1.15} & \gsvalue{14.91}{1.96} & \gsvalue{13.72}{0.44} & \gsvalue{16.08}{0.58} & \gsvalue{15.97}{0.23} \\[0.5pt]
     & Refined & \gsbestsp{13.82}{1.05} & \gsbestixp{13.19}{0.54} & \gsbestsp{21.93}{0.19} & \gsbestixp{18.74}{0.42} & \gsbestsp{19.69}{0.80} & \gsbestixp{19.43}{0.35} \\[0.5pt]
    \gsenvironmentrule
    Yokai 9C, perfect recall & Single Shot & \gsvalue{1.38}{0.42} & \gsvalue{0.32}{0.18} & \gsvalue{3.74}{0.25} & \gsvalue{3.62}{0.11} & \gsvalue{5.44}{0.40} & \gsvalue{5.69}{0.16} \\[0.5pt]
     & First Attempt & \gsvalue{0.31}{0.21} & \gsvalue{0.18}{0.04} & \gsvalue{3.52}{0.11} & \gsvalue{3.43}{0.05} & \gsvalue{4.01}{0.05} & \gsvalue{4.08}{0.03} \\[0.5pt]
     & Refined & \gsbestsp{1.54}{0.53} & \gsbestixp{0.69}{0.14} & \gsbestsp{4.88}{0.27} & \gsbestixp{4.71}{0.13} & \gsbestsp{7.48}{0.45} & \gsbestixp{7.33}{0.25} \\[0.5pt]
    \gsenvironmentrule
    Yokai 9C, imperfect recall & Single Shot & \gsbestsp{1.27}{0.47} & \gsbestixp{0.99}{0.22} & \gsvalue{3.72}{0.18} & \gsvalue{3.72}{0.09} & \gsvalue{4.82}{0.20} & \gsvalue{5.05}{0.11} \\[0.5pt]
     & First Attempt & \gsvalue{0.06}{0.03} & \gsvalue{0.05}{0.01} & \gsvalue{3.11}{0.31} & \gsvalue{3.27}{0.11} & \gsvalue{4.06}{0.05} & \gsvalue{4.07}{0.02} \\[0.5pt]
     & Refined & \gsvalue{0.68}{0.39} & \gsvalue{0.32}{0.08} & \gsbestsp{4.89}{0.46} & \gsbestixp{4.28}{0.10} & \gsbestsp{6.33}{0.52} & \gsbestixp{6.45}{0.17} \\[0.5pt]
    \gsenvironmentrule
    Yokai 16C, perfect recall & Single Shot & \gsvalue{0.24}{0.21} & \gsvalue{0.00}{0.00} & \gsvalue{2.46}{0.64} & \gsvalue{2.40}{0.25} & \gsvalue{7.54}{0.44} & \gsvalue{7.85}{0.18} \\[0.5pt]
     & First Attempt & \gsvalue{0.01}{0.00} & \gsvalue{0.00}{0.00} & \gsvalue{2.19}{0.26} & \gsvalue{1.88}{0.23} & \gsvalue{4.89}{0.27} & \gsvalue{5.18}{0.05} \\[0.5pt]
     & Refined & \gsbestsp{0.44}{0.17} & \gsbestixp{0.05}{0.02} & \gsbestsp{6.29}{0.39} & \gsbestixp{5.13}{0.20} & \gsbestsp{8.21}{0.94} & \gsbestixp{8.03}{0.35} \\[0.5pt]
    \bottomrule
    \end{tabular*}
    \endgroup
\end{table}

\clearpage
\section{Inter-Model Cross-Play Matrices}
\label{app:cross-model-matrices}

\begin{figure}[!ht]
    \input{Figures/cross_model/appendix_figure}
\end{figure}

\clearpage
\section{Prior Knowledge}
\label{app:prior-knowledge}
As discussed in Section~\ref{sec:evaluating-prior-knowledge}, language models acquire prior knowledge during training that may influence their coordination performance. To probe their familiarity with the evaluation environments, we ask a series of questions about these environments, detailed below. The questions are designed to elicit readily verifiable answers, which we evaluate using an \ac{llm} judge and the rubric provided below. Our rubric awards partial credit for incomplete or partly correct answers.

\subsection{Overcooked V2}
\label{app:prior-knowledge-ocv2}
\begingroup\setlength{\parskip}{3pt}
\paragraph{Grader instruction} For all questions, award \textbf{1} for a correct explanation, \textbf{0.5} for partial understanding, and \textbf{0} for absent or fundamentally incorrect answers, unless specified otherwise. Credit equivalent explanations and assess errors only against the criterion they affect. Merely naming a mechanism earns no credit.

\begin{samepage}
\medskip\noindent\textbf{Q1: Benchmark differences}\par\nobreak
\textbf{Question.} What is the Overcooked V2 benchmark? Identify the main environment-design components that differ from the original Overcooked benchmark.
\end{samepage}

\textbf{Rubric.} Assess three components:
\begin{list}{$\bullet$}{\setlength{\leftmargin}{1.5em}\setlength{\topsep}{3pt}\setlength{\partopsep}{0pt}\setlength{\parsep}{0pt}\setlength{\itemsep}{1pt}}
    \item \textbf{Partial observability:} restricted local observations and asymmetric access to recipe information.
    \item \textbf{Stochasticity:} randomly sampled target recipes that require agents to adapt.
    \item \textbf{Communication:} arbitrary behavioural signals versus a recipe-reveal button whose meaning requires no prior agreement.
\end{list}
Q1 is the mean of the three component scores.

\begin{samepage}
\medskip\noindent\textbf{Q2: Benchmark tasks}\par\nobreak
\textbf{Question.} Name the six benchmark tasks evaluated in the Overcooked V2 paper and describe their mechanics and layouts.
\end{samepage}

\textbf{Rubric.} Award one point for each correctly identified environment: \texttt{grounded\_coord\_simple}, \texttt{grounded\_coord\_ring}, \texttt{test\_time\_simple}, \texttt{test\_time\_wide}, \texttt{demo\_cook\_simple}, and \texttt{demo\_cook\_wide}. Accept unambiguous naming equivalents and documented paper-table aliases. Count each environment once, with no partial credit or deductions for additional incorrect names. Report Q2 out of six; mechanics and layouts are assessed separately below.

\begin{samepage}
\medskip\noindent\textbf{Q3: Observations, actions, and rewards}\par\nobreak
\textbf{Question.} Specify the exact observation tensor shapes and feature encodings, available actions, and numerical rewards and penalties in Overcooked V2, including differences across the six tasks.
\end{samepage}

\textbf{Rubric.} Assess three equally weighted subscales:
\begin{list}{$\bullet$}{\setlength{\leftmargin}{1.5em}\setlength{\topsep}{3pt}\setlength{\partopsep}{0pt}\setlength{\parsep}{0pt}\setlength{\itemsep}{1pt}}
    \item \textbf{Observations:} score tensor shapes, agent features, static features, object/recipe encodings, and timer/visibility/delivery-feedback features separately. 
    \item \textbf{Actions:} score identification of the six actions and explanation of grid-relative movement and facing-tile interaction separately.
    \item \textbf{Rewards:} score correct delivery ($+20$), incorrect delivery ($-20$), recipe reveal ($-5$), and separate shaping bonuses for ingredient placement ($+3$), plate pickup ($+3$), and correct dish pickup ($+5$).
\end{list}
Average criteria within each subscale, then average the three subscales to obtain Q3.

\begin{samepage}
\medskip\noindent\textbf{Q4: Coordination challenges and strategies}\par\nobreak
\textbf{Question.} What is the key zero-shot coordination challenge in each of the six environments, and what strategy addresses it?
\end{samepage}

\textbf{Rubric.} For each environment, separately assess recognition of the information problem and a feasible coordination strategy:
\begin{list}{$\bullet$}{\setlength{\leftmargin}{1.5em}\setlength{\topsep}{3pt}\setlength{\partopsep}{0pt}\setlength{\parsep}{0pt}\setlength{\itemsep}{1pt}}
    \item \textbf{Grounded coordination:} asymmetric recipe access; use the recipe-reveal button.
    \item \textbf{Test-time coordination:} potentially incompatible signals; use delivery feedback to establish and adapt a protocol.
    \item \textbf{Demo Cook:} asymmetric recipe access; infer the recipe from demonstrated ingredient placement in the shared pot.
\end{list}
Accept feasible alternatives that do not assume an already shared arbitrary convention. Q4 averages both criteria across all six environments.

\begin{samepage}
\medskip\noindent\textbf{Q5: Layout reconstruction}\par\nobreak
\textbf{Question.} Draw an ASCII grid of each environment, giving its dimensions and identifying walls, floor, ingredient and plate dispensers, pots, serving stations, recipe indicators/buttons, and template agent positions.
\end{samepage}

\textbf{Rubric.} Assess four criteria per environment:
\begin{list}{$\bullet$}{\setlength{\leftmargin}{1.5em}\setlength{\topsep}{3pt}\setlength{\partopsep}{0pt}\setlength{\parsep}{0pt}\setlength{\itemsep}{1pt}}
    \item Grid dimensions.
    \item Essential stations and ingredient identities.
    \item Floor connectivity, agent separation, and bottlenecks.
    \item Role-specific access to information, facilities, and communication channels.
\end{list}
Q5 averages these criteria across the six environments. Allow consistent rotations, reflections, legend conversions, and agent-label swaps. Report exact grid matches separately as a binary diagnostic, excluded from Q5.
\endgroup

\subsection{Hanabi}
\label{app:prior-knowledge-hanabi}
We assess knowledge of H-group Level 1 conventions using convention-recall questions and worked examples.

\textbf{Convention questions.} Each question begins: ``In standard two-player Hanabi using H-group Level 1 conventions\ldots''

\begin{center}
\small
\renewcommand{\arraystretch}{1.1}
\begin{tabular}{@{}p{0.37\linewidth}@{\hspace{1em}}p{0.60\linewidth}@{}}
\toprule
\textbf{Question} & \textbf{Full-credit criterion} \\
\midrule
What is a Play Clue? & A clue signals that the recipient should play its focused card. \\
How is the focus of a clue determined? & Prioritise chop if touched; otherwise the leftmost newly clued card, or the leftmost re-touched card if none are new. \\
What is a 5 Save? & A number-5 clue protects an endangered chop 5 from discard. \\
What is a 2 Save? & A number-2 clue protects an endangered chop 2 from discard. \\
What is a Critical Save? & A clue protects a needed last remaining copy from impending discard. \\
What is a Self-Prompt? & The recipient plays a previously clued connecting card in their own hand before the newly indicated card. \\
What is the chop? & The oldest unclued card, which is the predictable default discard. \\
What is a Delayed Play Clue? & A clue indicates a card intended for play after prerequisite connecting cards are played. \\
What is the Good Touch Principle? & All cards touched by a clue should be useful for eventual play. \\
What is the Save Principle? & Important cards in danger of discard must be protected. \\
What is the Minimum Clue Value Principle? & A normal clue must obtain at least one previously unclued card for play or saving. \\
What are the Early Game rules? & Before the first chop discard, exhaust available valid Play Clues and Save Clues. \\
\bottomrule
\end{tabular}
\end{center}

\textbf{Recall rubric.} Award \textbf{1} for the correct defining mechanism, \textbf{0.5} for substantive but incomplete or partly mistaken understanding, and \textbf{0} for absent or fundamentally incorrect understanding. Accept equivalent wording and clearly implied information; optional exceptions, examples, and exhaustive explanations are not required.

\textbf{Worked examples.} We take the worked examples from the H-group convention documentation~\citep{hgroupconventions}. We give one example below.

Alice and Bob are playing using H-group Level 1 conventions. The current stacks are Red 3, Yellow 4, Green 3, Blue 3, and Purple 2. Neither player has previously clued cards. Alice gives Bob a number-4 clue touching only slot 4 of his five-card hand, numbered from newest to oldest. Bob interprets this as a Play Clue. Which possible card identities should he note?

\textbf{Expected answer.} Red 4, Green 4, or Blue 4, the only 4s currently playable.

\textbf{Worked-example rubric.} Award \textbf{1} if the response clearly gives the correct requested action, card, clue, or inference, and \textbf{0} otherwise, including when incompatible alternatives remain unresolved. Accept equivalent wording and consistent card-position descriptions. Grade the conclusion rather than the explanation; incidental reasoning errors do not change the score when the answer is clear. Inconsistent or underspecified questions are marked unscorable.

\begin{table}[!t]
% Values rounded directly from the three-run aggregate; provenance: prior_knowledge_sources.json.
    \centering
    \caption{\textbf{Prior knowledge of Overcooked V2.} Entries report the mean $\pm$ standard error of the mean across three independent answer-generation and grading runs, scored by GPT-6.1 Sol (High) using the rubric. Q1 assesses benchmark differences; Q2, environment names; Q3, observations, actions, and rewards; Q4, coordination challenges and strategies; and Q5, layout reconstruction. Q2 counts correctly identified environments out of six; other scores range from zero to one. Overall scores weight the five questions equally after normalising Q2 by six.}
    \label{tab:ocv2-prior-knowledge}
    \begingroup
    \footnotesize
    \setlength{\tabcolsep}{2pt}
    \renewcommand{\arraystretch}{1.15}
    \begin{tabular*}{\linewidth}{@{\extracolsep{\fill}}lcccccc@{}}
    \toprule
    Model & Q1 & Q2 (/6) & Q3 & Q4 & Q5 & Overall \\
    \midrule
    GPT-5.6 Sol & $0.33\,\pm\,0.10$ & $0.00\,\pm\,0.00$ & $0.52\,\pm\,0.03$ & $0.06\,\pm\,0.06$ & $0.14\,\pm\,0.01$ & $0.21\,\pm\,0.03$ \\
    GPT-6 Astra & $0.28\,\pm\,0.06$ & $0.00\,\pm\,0.00$ & $0.22\,\pm\,0.11$ & $0.00\,\pm\,0.00$ & $0.00\,\pm\,0.00$ & $0.10\,\pm\,0.02$ \\
    Claude Fable 5.1 & $0.83\,\pm\,0.00$ & $6.00\,\pm\,0.00$ & $0.73\,\pm\,0.04$ & $0.35\,\pm\,0.06$ & $0.38\,\pm\,0.05$ & $0.66\,\pm\,0.02$ \\
    Claude Opus 5.5 & $0.83\,\pm\,0.00$ & $6.00\,\pm\,0.00$ & $0.80\,\pm\,0.01$ & $0.56\,\pm\,0.12$ & $0.37\,\pm\,0.01$ & $0.71\,\pm\,0.02$ \\
    \bottomrule
    \end{tabular*}
    \endgroup

\vspace{1em}
% Values rounded from per-round scores; provenance: prior_knowledge_hanabi_sources.json.
    \centering
    \caption{\textbf{Prior knowledge of H-group Level 1 conventions in Hanabi.} Entries report the mean $\pm$ standard error of the mean across three independent answer-generation rounds, scored by GPT-6.1 Sol (High). Each round includes 12 convention-recall questions and 12 worked examples. Scores range from zero to one.}
    \label{tab:hanabi-prior-knowledge}
    \begingroup
    \footnotesize
    \setlength{\tabcolsep}{8pt}
    \renewcommand{\arraystretch}{1.15}
    \begin{tabular}{lcc}
    \toprule
    Model & Recall & Worked examples \\
    \midrule
    GPT-5.6 Sol & $0.83\,\pm\,0.04$ & $0.42\,\pm\,0.10$ \\
    GPT-6 Astra & $0.85\,\pm\,0.01$ & $0.75\,\pm\,0.05$ \\
    Claude Fable 5.1 & $0.86\,\pm\,0.05$ & $0.69\,\pm\,0.06$ \\
    Claude Opus 5.5 & $0.89\,\pm\,0.01$ & $0.75\,\pm\,0.00$ \\
    \bottomrule
    \end{tabular}
    \endgroup

\end{table}

\subsection{Results}
Table~\ref{tab:ocv2-prior-knowledge} shows that models' knowledge of Overcooked V2 is uneven, with the largest gaps in coordination strategies (Q4) and layout reconstruction (Q5). While the Claude models demonstrate familiarity with the benchmark, the OpenAI models show little benchmark-specific knowledge: neither identifies any of the six tasks, and both score poorly on coordination strategies and layout reconstruction. Despite this, GPT-6 Astra outperforms all evaluated existing \ac{zsc} methods in \ac{ixp} on Overcooked V2. This provides evidence that the models possess genuine \ac{zsc} capabilities, rather than solely relying on prior knowledge of benchmark-specific solutions.

In Hanabi, all models achieve high convention-recall scores (Table~\ref{tab:hanabi-prior-knowledge}), indicating familiarity with H-group Level 1 conventions. Performance on the worked examples is lower. These results suggest that recalling conventions does not always translate into correctly applying them. Access to documentation and simulator feedback may therefore help translate this knowledge into effective policies.

\clearpage
\section{Coordination Mechanisms in Generated Policies}
\label{app:generated-policies}
We highlight snippets from the generated policies that demonstrate task-specific capabilities, including the ability to adapt to their partner's behaviour. We also provide examples of complete code policies in Appendix~\ref{app:full-generated-policies}.

\subsection{Overcooked V2: Grounded Communication}
\label{app:grounded-communication-examples}
\subsubsection{Pressing the Recipe Button When the Cook Needs Information}
\textbf{GPT-6 Astra, Grounded Simple, generation seed 5.}
\begin{lstlisting}[style=generatedsnippet,firstline=44,lastline=54,firstnumber=44]
    retry_base = jnp.maximum(20, (3 * fill_gap) // 2)
    retry_interval = retry_base * jnp.left_shift(jnp.asarray(1, jnp.int32), jnp.minimum(jnp.maximum(retries - 1, 0), 3))
    signal_delay = jnp.maximum(3, (2 * fill_gap) // 3 - 5)
    next_cycle_steps = 20 + (11 * fill_gap) // 3
    cycle_signal = (count == 0) & (((empty_age >= signal_delay) & (since_signal >= retry_interval)) | changed)
    correction = (count > 0) & (jnp.argmax(pot[2:]) != recipe) & (warned == 0) & (since_signal >= 10)
    ready = (cycle_signal | correction) & ~display_active & (time + next_cycle_steps < 400)
    action = approach(row, col, facing, 2, 2, 2)
    at_button = (row == 2) & (col == 2) & (facing == 2)
    action = jnp.where(at_button & ~ready, 4, action).astype(jnp.int32)
    activated = at_button & ready
\end{lstlisting}
The informed agent watches the shared pot. It presses the button to display the recipe when a new batch needs to start, or when the cook is adding the wrong ingredient. It avoids pressing while the display is already active. The waiting times depend on \texttt{fill\_gap}, an estimate of how quickly the partner adds ingredients, updated in source lines 34--36. This makes the timing responsive to the partner's pace and avoids unnecessary costly button presses.

\begin{lstlisting}[style=generatedsnippet,firstline=101,lastline=105,firstnumber=101]
    displayed = jnp.sum(obs[:, :, 36:39], axis=(0, 1))
    visible = jnp.sum(displayed) > 0
    recipe = jnp.where(visible, jnp.argmax(displayed), recipe)
    valid = jnp.where(visible, 1, valid)
    informed = jnp.maximum(informed, visible.astype(jnp.int32))
\end{lstlisting}
The cook reads the recipe revealed by the button and remembers it after the display disappears. The meaning of this message is supplied by the environment: the agents do not need to agree on a private code for ingredient types.

\begin{lstlisting}[style=generatedsnippet,firstline=130,lastline=138,firstnumber=130]
    mismatched = jnp.argmax(inv[2:]) != recipe
    park_dish = dish & mismatched & (valid == 1) & free_counter
    park_raw = raw & mismatched & (valid == 1) & (count == 0) & free_counter

    action = jnp.where(dish, jnp.where(park_dish, store_action, deliver),
             jnp.where(plate, jnp.where(cooked, pot_action, jnp.where(pot_action == 5, 4, pot_action)),
             jnp.where(raw, jnp.where(park_raw, store_action, pot_action),
             jnp.where(count >= 3, get_plate,
             jnp.where((valid > 0) | (count > 0), acquire, wait_display)))))
\end{lstlisting}
If a new display contradicts the food the cook is carrying, it can put that food on a free counter. A wrong raw ingredient is set aside before starting an empty pot; a wrong finished dish is stored instead of delivered. This lets the cook respond to information that arrives after it has already begun working.

\subsubsection{Helping When an Ingredient Cue Does Not Produce a Response}
\textbf{GPT-6 Astra, Grounded Ring, generation seed 5.}
\begin{lstlisting}[style=generatedsnippet,firstline=123,lastline=139,firstnumber=123]
    # While displaying an ingredient, remain beside its supply. This keeps the
    # always-on recipe display visible, so a recipe change is noticed promptly.
    correct = (inv[2 + kind] > 0) & (jnp.sum(recipe) > 0)
    cue_route = jnp.asarray(ROUTES, jnp.int32)[jnp.asarray(_SOURCES)[kind], state]
    cue_route = jnp.where(cue_route == 5, 4, cue_route)
    action = jnp.where(correct, cue_route, action).astype(jnp.int32)
    # A semantic cue may not be understood by a separately generated cook.
    # Offer a direct public recipe after a prolonged lack of visible work.
    activity = jnp.any(obs[:, :, 15:20] > 0)
    quiet = jnp.where(activity, 0, m['quiet'] + 1).astype(jnp.int32)
    eligible = m['eligible'] | activity
    responsive = m['responsive'] | (activity & (m['t'] - m['help_time'] <= 16))
    patience = jnp.where(responsive, 28, 40)
    enter = (quiet >= patience) & eligible & (m['t'] < 350) & ~m['helping']
    helping = (m['helping'] | enter) & ~activity
    preferred_button = jnp.where(responsive, m['help_button'], _nearest(_BUTTONS, state))
    button = jnp.where(enter, preferred_button, m['help_button'])
\end{lstlisting}
The outer agent first holds the requested ingredient beside its supply, where it can keep watching the recipe display. If the partner shows no visible inventory activity for long enough, it switches to helping through a button. It remembers whether activity followed a previous attempt and adjusts its patience. These are simple observations of the partner's response, rather than access to the partner's policy.

\begin{lstlisting}[style=generatedsnippet,firstline=64,lastline=75,firstnumber=64]
    shown = jnp.max(obs[:, :, 36:39], axis=(0, 1))
    delivered = m['last_cooked'] & (inv[1] == 0) & (m['last_action'] == 5) & ((m['last_target'] == 38) | (m['last_target'] == 42))
    recipe = jnp.where(delivered, jnp.zeros(3, jnp.int32), m['recipe'])
    recipe = jnp.where(jnp.sum(shown) > 0, shown, recipe)
    armed = m['armed'] & ~delivered
    known = jnp.sum(recipe) > 0
    ingredient = jnp.argmax(recipe - pot[31:34])
    full = (jnp.sum(pot[31:34]) >= 3) | (pot[30] > 0) | (pot[39] > 0)
    target = jnp.where(known, jnp.asarray(_SOURCES)[ingredient], _nearest(_BUTTONS, state))
    target = jnp.where(full, _nearest(_PLATES, state), target)
    prearm = full & ~armed & (jnp.sum(inv) == 0) & (m['t'] < 354)
    target = jnp.where(prearm, _nearest(_BUTTONS, state), target)
\end{lstlisting}
The inner cook also has a fallback: when it does not know the recipe, it approaches a button itself. It can prepare another button read while a batch finishes. Thus, this policy supports both partner assistance and direct information gathering; successful cooking need not always depend on the outer agent's cue.

\clearpage
\subsection{Overcooked V2: Demonstration Cooking}
\label{app:demonstration-cooking-examples}
\subsubsection{Starting a Dish to Show the Partner What to Cook}
\textbf{Claude Opus 5, Demo Simple, generation seed 5.}
\begin{lstlisting}[style=generatedsnippet,firstline=253,lastline=260,firstnumber=253]
    # ======================= left-room seat =============================
    want = _i32(jnp.where(pot_n > 0, pot_type, recipe))
    pot_open = (
        pot_idle
        & (want < 3)
        & ((pot_n > 0) | fresh_recipe)
        & ((~wrong_batch) | may_recover)
    )
\end{lstlisting}
Only the left-room agent can see the recipe display. When the pot is empty, it chooses the displayed recipe; once ingredients are present, it continues that batch. The \texttt{fresh\_recipe} check delays starting a new batch while a delivery may still change the requested dish. This agent's first ingredient both advances cooking and shows its partner what to make.

\begin{lstlisting}[style=generatedsnippet,firstline=322,lastline=328,firstnumber=322]
    reachable_in_pot = live_counts[0:2]
    fill_type = _i32(jnp.where(
        pot_mixed > 0,
        jnp.where(reachable_in_pot.sum() > 0, jnp.argmax(reachable_in_pot), 2),
        pot_type))
    fill_possible = pot_idle & (pot_n > 0) & (pot_n < 3) & (fill_type < 2)
    take_fill = jnp.where(fill_type == 0, M_ING0_R, M_ING1_R)
\end{lstlisting}
The right-room agent reads the ingredient already in the shared pot and fetches more of that type. It does not need to see the recipe display to help complete the dish. In normal operation it waits for the informed partner to start; separate fallback code permits solo cooking after prolonged inactivity (source lines 330--345).

\begin{lstlisting}[style=generatedsnippet,firstline=370,lastline=372,firstnumber=370]
    is_right = pos[1] > _POT_C
    mode = _i32(jnp.where(is_right, right_mode, left_mode))
    action = _navigate(pos, face_act, mode)
\end{lstlisting}
Both agents use the same decision function, but choose their role from their location. The left room provides recipe information and ingredients; the right room has the plates and delivery station. The division of labour follows the task's geometry.

\clearpage
\subsubsection{Reading Held Ingredients and Avoiding Duplicate Work}
\textbf{GPT-6 Astra, Demo Wide, generation seed 4.}
\begin{lstlisting}[style=generatedsnippet,firstline=76,lastline=86,firstnumber=76]
    cue_stable = jnp.where(other_raw & (other_type == memory[6]) & (other_x == 5), memory[7]+1, 0)
    stalled = jnp.where(n == memory[3], memory[4]+1, 0)
    carrier = (memory[8] > 0) | ((n == 2) & other_raw & (cue_stable >= 6) & (stalled >= 6))
    # A persistent literal ingredient can be a signal rather than a promised
    # deposit. Allow it again after a delivery guard instead of waiting forever.
    persistent_cue = other_raw & (cue_stable >= 4) & (memory[5] >= 8)
    fresh = (memory[2] > 0) | (visible_partner & (jnp.sum(other) == 0)) | persistent_cue
    ingredient = jnp.where((memory[1] < 0) & fresh & other_raw, jnp.argmax(other[2:]), memory[1])
    ingredient = jnp.where(n > 0, jnp.argmax(pot[2:]), ingredient)
    invalid_pot = (jnp.sum(pot[4:]) > 0) | ((pot[2] > 0) & (pot[3] > 0))
    ingredient = jnp.where((n > 0) & invalid_pot, jnp.where(other_raw, other_type, 0), ingredient)
\end{lstlisting}
The upper cook can infer the recipe from an ingredient in its partner's hand, before that ingredient reaches the pot. If the partner keeps holding the same ingredient without filling the pot, the cook can treat it as a persistent demonstration. Once cooking starts, the contents of the pot normally determine which ingredient to fetch next.

\begin{lstlisting}[style=generatedsnippet,firstline=125,lastline=129,firstnumber=125]
    wait = jnp.where(x < 5, 0, jnp.where(x > 5, 2, 4))
    action = jnp.where(ingredient >= 0, fetch, wait)
    # The partner can finish the pot without causing a surplus pickup.
    action = jnp.where((n == 2) & other_raw & (stalled < 6), wait, action)
    action = jnp.where((n >= 3) | (timer > 0) | (pot[1] > 0), plate, action)
\end{lstlisting}
When the pot already contains two ingredients and the partner is carrying another, the cook briefly waits instead of fetching a surplus ingredient. If the pot stays unchanged for six steps, this waiting condition expires. The policy anticipates a useful contribution without waiting indefinitely for it.

\clearpage
\subsection{Test-Time Adaptation}
\label{app:policy-test-time-adaptation}

\subsubsection{One-Sided Adaptation: Learning to Interpret a Fixed Sender}
\label{app:policy-test-time-wide}
\label{app:adaptation-one-sided}
\textbf{Claude Fable 5.1, Test-time Wide, generation seed 2.}
\begin{lstlisting}[style=generatedsnippet,firstline=398,lastline=408,firstnumber=398]
    # recipe 0: stand at (4,1) facing LEFT (turn is blocked by supply at (4,0))
    at0 = (my_r == 4) & (my_c == 1)
    act0 = jnp.where(at0 & (my_dir == LEFT), A_STAY,
                     jnp.where(at0, A_LEFT,
                               jnp.where(my_r == 5, A_UP, A_LEFT)))
    # recipe 1: stand at (4,3) facing RIGHT (arrive from (4,2) moving right)
    at1 = (my_r == 4) & (my_c == 3)
    act1 = jnp.where(at1 & (my_dir == RIGHT), A_STAY,
                     jnp.where(my_r == 5, A_UP,
                               jnp.where(my_c < 3, A_RIGHT, A_LEFT)))
    action = _i32(jnp.where(recipe == 1, act1, act0))
\end{lstlisting}
One agent can see the requested recipe; the other must cook without seeing it. The informed agent signals recipe 0 by standing on the left and facing left, and recipe 1 by standing on the right and facing right. This signalling rule stays fixed, regardless of how the cook responds.

\begin{lstlisting}[style=generatedsnippet,firstline=249,lastline=266,firstnumber=249]
    hf = jnp.clip(H_FEAT, 0, NF - 1)
    evt_h = apply_evt[hf] & (H_FEAT >= 0)
    val_h = f_val[hf]
    factor = jnp.where(evt_h[:, None], jnp.where(val_h[:, None] == H_PRED, LIKE_OK, LIKE_BAD), 1.0)
    # boundary change: the recipe changed right after my delivery d_last, so
    # that delivery was correct; this pins the polarity of feature phi.
    bchg = new_evt & (prev_anchor >= 0) & (f_val != prev_anchor)
    s_star = (prev_anchor ^ jnp.clip(mem['d_last'], 0, 1)) & 1
    kill = bchg[hf] & (H_FEAT >= 0) & (H_POL != s_star[hf])
    factor = jnp.where(kill[:, None], 0.05, factor)
    W2 = W * factor
    # features that never changed across a delivery carry no evidence: keep
    # their hypothesis mass frozen (only the pinned recipe belief updates).
    gated = (nchg[hf] == 0) & (H_FEAT >= 0)
    mass_old = W.sum(1)
    mass_new = jnp.maximum(W2.sum(1), 1e-9)
    W3 = jnp.where(gated[:, None], W2 * (mass_old / mass_new)[:, None], W2)
    W = W3 / jnp.maximum(W3.sum(), 1e-9)
\end{lstlisting}
The cook learns how to read its partner. Does standing on the left mean recipe 0 or recipe 1? Is facing direction more useful than position? It considers 17 possible explanations, including that the partner provides no useful signal. It updates these beliefs by watching how the signals change around deliveries, then uses them to choose a dish.

This is one-sided adaptation: the cook learns the signal's meaning, but the sender keeps using the same rule.

\clearpage
\subsubsection{Mutual Adaptation: Learning Signals and Responding to Corrections}
\label{app:adaptation-response-learning}
\textbf{GPT-6 Astra, Test-time Simple, generation seed 2.}
\begin{lstlisting}[style=generatedsnippet,firstline=64,lastline=86,firstnumber=64]
    counts = jnp.asarray([at(obs, 2, 4, r, c, x) for x in (31, 32, 33)])
    total = jnp.sum(counts)
    visible = c == 2
    new_pot = visible & (total > 0) & (previous_total == 0)
    learn = new_pot & (age >= 4) & (counts[2] == 0)
    reversal = learn & (response[symbol] >= 0) & (response[symbol] != jnp.argmax(counts))
    response = jnp.where(reversal, jnp.full_like(response, -1), response)
    trials = jnp.where(reversal, jnp.zeros_like(trials), trials)
    response = response.at[symbol].set(jnp.where(learn, jnp.argmax(counts), response[symbol]))
    trials = trials.at[symbol].add(learn.astype(jnp.int32))
    previous_total = jnp.where(visible, total, previous_total)
    idle = jnp.where((previous_total == 0) & (symbol == last_symbol), idle + 1, 0)
    stalled = (idle >= 32) & (age >= 20)
    response = response.at[symbol].set(jnp.where(stalled, -1, response[symbol]))
    trials = trials.at[symbol].add(stalled.astype(jnp.int32))
    idle = jnp.where(stalled, 0, idle)

    # Start with the spatially aligned code, but prefer a symbol whose
    # observed effect is correct over an untested convention.
    rank = jnp.where(recipe == 0, jnp.asarray([3, 0, 2, 1]), jnp.asarray([3, 1, 2, 0]))
    costs = jnp.where(response == recipe, rank,
                     jnp.where(response < 0, 10 + 4 * trials + rank, 100 + 4 * trials + rank))
    desired = jnp.argmin(costs)
\end{lstlisting}
The sender places a marker and watches which ingredient the cook adds to the pot. It remembers what each signal made the cook do, then reuses signals that worked. If the cook does nothing, it tries another signal. If the cook starts interpreting a signal differently, it resets what it has learned. The same recipe can therefore be communicated differently to different partners.

\begin{lstlisting}[style=generatedsnippet,firstline=181,lastline=196,firstnumber=181]
    # A partner can revise its cue after observing our first pot ingredient.
    # Beyond the ordinary post-delivery window, a settled replacement is
    # treated as a correction of this committed dish, not a new recipe draw.
    committed_type = jnp.argmax(counts)
    correction = settled & (total > 0) & (counts[2] == 0) & (pending_target < 0)
    correction = correction & (since_delivery > 24) & (stable != cycle_sig) & (cycle_sig >= 0)
    correction = correction & ((cycle_sig > 0) | (jnp.abs(meaning[0]) >= 32))
    correction = correction & ((stable > 0) | (streak >= 10))
    pending_target = jnp.where(correction, 1 - committed_type, pending_target)
    unchanged_run = jnp.where(correction, 0, unchanged_run)
    correction_value = jnp.where(committed_type == 0, 64, -64)
    meaning = meaning.at[new_index].set(jnp.where(correction, correction_value, meaning[new_index]))
    correction_opposite = jnp.where(new_index == 1, 4, 1)
    correction_spatial = correction & ((new_index == 1) | (new_index == 4))
    meaning = meaning.at[correction_opposite].set(jnp.where(correction_spatial,
                     -correction_value, meaning[correction_opposite]))
\end{lstlisting}
The cook also adjusts. It learns how long its partner takes to update a signal after delivery and waits accordingly (source lines 142--155). If the partner changes the signal after cooking has begun, the cook can treat this as a correction: ``You are making the wrong dish.'' It updates the signal's meaning and can store the unwanted dish for later instead of delivering it (source lines 211--224).

\clearpage
\subsubsection{Mutual Adaptation: Modelling the Partner's Decoder}
\label{app:adaptation-decoder-model}
\textbf{GPT-6 Astra, Test-time Simple, generation seed 4.}
\begin{lstlisting}[style=generatedsnippet,firstline=114,lastline=138,firstnumber=114]
    started = visible & (count > 0) & (memory['previous_count'] == 0)
    actual = jnp.argmax(pot[31:34])
    learned = started & (stable >= 6) & (actual < 2)
    counts = memory['counts'].at[signal,jnp.minimum(actual,1)].add(learned.astype(jnp.int32))
    idle = jnp.where(visible & (count == 0), memory['idle'] + 1, 0)
    stalled = (idle > 35) & (stable > 30)
    timeouts = memory['timeouts'].at[signal].set(jnp.where(stalled,1,memory['timeouts'][signal]))
    models,decoder_prior = decoder_models()
    response = models[:,signal]
    likelihood = jnp.where(response == actual,.98,jnp.where(response < 0,.35,.02))
    predict_decoder = jnp.where(learned,.85*memory['decoder_belief']+.15*decoder_prior,memory['decoder_belief'])
    posterior = predict_decoder * jnp.where(learned,likelihood,1.)
    timeout_event = stalled & (memory['timeouts'][signal] == 0)
    posterior *= jnp.where(timeout_event,jnp.where(response < 0,.98,.02),1.)
    posterior /= jnp.maximum(jnp.sum(posterior),1e-30)
    utility = jnp.where(models == recipe,1.,jnp.where(models < 0,-.25,-1.))
    score = jnp.sum(posterior[:,None]*utility,axis=0)
    patterns = jnp.arange(8)
    token_count = ((patterns&1)>0).astype(jnp.float32) + ((patterns&2)>0) + ((patterns&4)>0)
    score -= .012*token_count
    score += .012*(patterns == signal) + .008*(patterns == jnp.where(recipe == 0,1,4))
    target = jnp.argmax(score)
    # Finish a batch consistently before correcting the message; no additional
    # raw ingredients can enter after the pot reaches capacity.
    target = jnp.where((count > 0) & ((count < 3) | (actual == recipe)),signal,target)
\end{lstlisting}
This sender learns how its partner interprets different arrangements of markers. When the cook starts a dish, the sender uses the chosen ingredient to update its guesses about the partner's interpretation. If the cook does nothing, it considers that the message may be unrecognised. It then chooses the message most likely to make this partner cook the requested recipe.

\begin{lstlisting}[style=generatedsnippet,firstline=236,lastline=247,firstnumber=236]
    # A stable changed message before delivery cannot be recipe resampling.
    # It is evidence that the partner is correcting a misunderstanding.
    correction = (recipe >= 0) & (count > 0) & (age > 18) & (stable >= 6)
    correction &= (pattern > 0) & (pattern != accepted) & ~corrected & ~delivered
    wrong_type = jnp.argmax(pot[31:33])
    corrected_belief = condition_on_pattern(encoding_prior(),pattern)
    state_weight = jnp.where(jnp.arange(2) == wrong_type,.1,.9)
    corrected_belief = corrected_belief * state_weight[None,None,:]
    corrected_belief /= jnp.maximum(jnp.sum(corrected_belief),1e-12)
    belief = jnp.where(correction,corrected_belief,belief)
    accepted = jnp.where(correction,pattern,accepted)
    corrected |= correction
\end{lstlisting}
The cook can also recognise corrections. If the sender changes its message while food is already in the pot, and the new message persists, the cook treats this as evidence that it chose the wrong ingredient. It revises its interpretation of the message. Both sides can therefore adjust: the sender changes what it communicates, and the cook changes how it understands it.

\clearpage
\subsubsection{Changing an Encoding When Communication Fails}
\label{app:adaptation-encoding-repair}
\textbf{Claude Fable 5.1, Test-time Simple, generation seed 0.}
\begin{lstlisting}[style=generatedsnippet,firstline=318,lastline=328,firstnumber=318]
    rcell, _ = _read(obs, r, c, 2, 0)
    rn0, rn1 = rcell[36], rcell[37]
    rec = jnp.where(rn0 == 3, 0, jnp.where(rn1 == 3, 1, -1))
    # Adaptive recipe-1 encoding.  mode 0: {middle,bottom}; mode 1: {bottom}; mode 2: {middle}.
    # Recipe 0 is always {top}.  Switch mode when the cook clearly cannot read the signal
    # (pot stays empty for a long time while the signal is complete) or misreads it
    # (adds ingredient 0 to an empty pot while recipe 1 has been signalled for a while).
    mode = mem['pd']
    desired = jnp.stack([rec == 0,
                         (rec == 1) & (mode != 1),
                         (rec == 1) & (mode != 2)])
\end{lstlisting}
This sender can ask for recipe 1 in three ways: place markers on the middle and bottom counters, on the bottom counter only, or on the middle counter only. Recipe 0 always uses the top counter.

\begin{lstlisting}[style=generatedsnippet,firstline=333,lastline=352,firstnumber=333]

    outer = _update_outer(obs, r, c, mem['outer'], LOUT_R, LOUT_C)

    pot, pot_vis = _read(obs, r, c, 2, 4)
    pot_n0 = jnp.where(pot_vis, pot[31], mem['seen'] % 4)
    pot_tot = jnp.where(pot_vis, pot[31] + pot[32] + pot[33], mem['seen'] // 4)
    pot_busy = (pot_tot > 0) | (pot[30] > 0) | (pot[39] > 0)
    prev_n0 = mem['seen'] % 4
    complete = jnp.all(cur == desired)
    stable = jnp.where(complete, jnp.minimum(mem['since'] + 1, 1000), 0).astype(I32)
    pot_idle_steps = jnp.where(pot_vis & (~pot_busy), jnp.minimum(mem['empty'] + 1, 1000),
                               jnp.where(pot_vis, 0, mem['empty'])).astype(I32)
    idle_evidence = (rec == 1) & complete & (stable >= SIG_IDLE_STEPS) & (pot_idle_steps >= SIG_IDLE_STEPS)
    prev_tot = mem['seen'] // 4
    wrong_evidence = (rec == 1) & pot_vis & (prev_tot == 0) & (pot_n0 > prev_n0) & (mem['since'] >= SIG_WRONG_STABLE)
    nswitch = mem['flipw']
    switch = (idle_evidence | wrong_evidence) & (nswitch < SIG_MAX_SWITCH)
    mode_next = jnp.where(switch, (mode + 1) % 3, mode).astype(I32)
    stable = jnp.where(switch, 0, stable)
    pot_idle_steps = jnp.where(switch, 0, pot_idle_steps)
\end{lstlisting}
The sender watches whether its message works. It switches to another pattern if the cook leaves the pot empty for 20 steps despite a complete signal, or starts adding the wrong ingredient after the signal has been stable for 10 steps. It removes the old markers and places the new ones, with up to four switches per episode. This lets it try another way of communicating when the partner seems confused.

\clearpage
\subsubsection{Trying a Different Division of Labour}
\label{app:adaptation-plate-offer}
\textbf{Claude Fable 5.1, Test-time Wide, generation seed 4.}
\begin{lstlisting}[style=generatedsnippet,firstline=348,lastline=363,firstnumber=348]
    mode = m['mode']
    stage = m['stage']
    offer_t = m['offer_t']
    mode = jnp.where(read_event & (mode == 1), 0, mode)
    # the partner answered a plate offer with a dish we just delivered: offer again
    reoffer = delivered & (m['trusted'] == 1)
    start_offer = ((mode == 0) & (nread == 0) & (wait_t > WAIT_BEFORE_OFFER)) | reoffer
    mode = jnp.where(start_offer, 1, mode)
    stage = jnp.where(start_offer, 0, stage)
    stage = jnp.where(placed_offer, 1, stage)
    offer_t = jnp.where(placed_offer, t, offer_t)
    plate_on_counter = c32['plate_only'] | c34['plate_only']
    timeout = jnp.where(plate_on_counter, OFFER_TIMEOUT, OFFER_TAKEN_TIMEOUT)
    offer_done = (mode == 1) & (stage == 1) & (t - offer_t > timeout)
    mode = jnp.where(offer_done, 2, mode)
    offering = (mode == 1) & (stage == 1)
\end{lstlisting}
This policy can change who selects the dish. Normally, the cook reads its partner's signals and chooses what to serve. If it has food ready but has never received a usable signal, it can offer a plate on a shared counter. The partner, who can see the recipe, can then scoop the correct dish and pass it back for delivery.

\begin{lstlisting}[style=generatedsnippet,firstline=552,lastline=568,firstnumber=552]
    hold_t = jnp.where(hpl & (~potK_good) & (~potK_cooking), m['hold_t'] + 1, 0)

    conds = [
        hd,
        hpl & potK_good,
        hpl & (hold_t > HOLD_TIMEOUT),
        hpl,
        he & (c32['plate_only'] | c34['plate_only']) & (potK_good | potK_cooking),
    ]
    choices = [
        T_PUT,
        T_POTK,
        T_PUT,
        T_SIG,
        T_TAKE,
    ]
    task = jnp.select(conds, choices, T_SIG).astype(I32)
\end{lstlisting}
After delivering a dish received from the partner, the cook offers another plate. If nobody takes the plate, it eventually reclaims it. The partner's code above completes the exchange: take the plate, scoop the requested dish when ready, and put it on a shared counter. Otherwise, continue signalling. In the code, \texttt{T\_TAKE}, \texttt{T\_POTK}, \texttt{T\_PUT}, and \texttt{T\_SIG} are these four actions.

The code is written before play, but these choices are made during the episode. The policies can learn what a partner means, change how they communicate, respond to corrections, and try a different way of sharing the work.

\clearpage
\subsection{Overcooked V2: Coordination in the Novel Layouts}
\label{app:novel-layout-policy-examples}
\subsubsection{Reserving the Ingredients a Partner Is Already Carrying}
\textbf{GPT-6 Astra, Pipeline Cook, generation seed 4.}
\begin{lstlisting}[style=generatedsnippet,firstline=84,lastline=93,firstnumber=84]
    # Detect a likely server from actual plate/dish inventory, not seat number.
    other_serving=(t-partner_time<18)&(partner_inv[0]>0)
    full_present=jnp.any(full)
    need_recipe=(~known_recipe)|((t>=cooldown)&(recipe_time<cooldown))|(jnp.any(suspect_full)&(t-recipe_time>25))
    can_gather=~full_present&(t>=cooldown)&~need_recipe
    # A seen partner's ingredient is reserved when deciding what to fetch.
    reserved=jnp.where((t-partner_time<20)&(partner_inv[0]==0),partner_inv[2:],0)
    wanted=jnp.maximum(deficit-reserved,0)
    pending=(jnp.sum(deficit)>0)&(jnp.sum(wanted)==0)
    can_gather=can_gather&~pending
\end{lstlisting}
The agent treats a recently observed partner holding a plate as a likely server. It also subtracts ingredients in the partner's hands from what still needs collecting. If those ingredients would complete the batch, it pauses gathering. These assumptions expire when the observations become old, so an absent or delayed partner does not reserve the work forever.

\begin{lstlisting}[style=generatedsnippet,firstline=186,lastline=198,firstnumber=186]
    close_conflict=obstructed&(dist[pos,partner_idx]<=2)
    traffic=jnp.where(close_conflict,m['traffic']+1,0)
    my_priority=inv[1]*4+raw*2+plate
    their_priority=pinv[1]*4+((pinv[0]==0)&(jnp.sum(pinv[2:])>0))*2+pinv[0]
    should_yield=(my_priority<their_priority)|((my_priority==their_priority)&(seat==1))|(traffic>3)
    begin_yield=((close_conflict&should_yield)|stalled)&(t>=m['yield_until'])
    progress_time=jnp.where(begin_yield,t,progress_time)
    progress_dist=jnp.where(begin_yield,999,progress_dist)
    yield_until=jnp.where(begin_yield,t+jnp.where(stalled,dist[pos,bay]+6,4)+jax.random.randint(rng,(),0,3),m['yield_until'])
    yield_pos=jnp.where(begin_yield,bay,m['yield_pos'])
    yielding=t<yield_until
    routing=jnp.where(obstructed,jnp.min(jnp.where(goalmask[None,:],dist,999),axis=1),routing)
    routing=jnp.where(yielding,dist[:,yield_pos],routing)
\end{lstlisting}
When the agents block one another, the policy gives priority according to what they are carrying: a finished dish receives more priority than a raw ingredient or an empty plate. It can move into a side bay and wait for the partner to pass. A persistent conflict or lack of progress can also trigger yielding, even when the usual priority rule would let it continue.

\clearpage
\subsubsection{Accounting for Work Underway Elsewhere}
\textbf{Claude Opus 5.5, Block Grid, generation seed 0.}
\begin{lstlisting}[style=generatedsnippet,firstline=242,lastline=253,firstnumber=242]
    # if the current recipe is already cooking/ready somewhere, new pots cook
    # a different recipe as stock instead of a likely-stale duplicate
    dup = jnp.any((pcooking | pcooked) & jnp.all(cnt == recipe[None], axis=1)) & known
    ridx = jnp.argmax(jnp.all(J_RECIPES == recipe[None], axis=1))
    alt = J_RECIPES[(ridx + 1 + side) % 4]
    compat = known & jnp.all(cnt <= recipe[None], axis=1) & ~(dup & (csum == 0))
    sup_ok = jnp.all(cnt[:, None, :] <= J_RECIPES[None], axis=2)  # [4,4]
    has_sup = jnp.any(sup_ok, axis=1)
    sup_goal = J_RECIPES[jnp.argmax(sup_ok, axis=1)]
    sup_goal = jnp.where((dup & (csum == 0))[:, None], alt[None], sup_goal)
    fill0 = cnt + jnp.stack([3 - csum, jnp.zeros_like(csum), jnp.zeros_like(csum)], axis=1)
    goal = jnp.where(compat[:, None], recipe[None], jnp.where(has_sup[:, None], sup_goal, fill0))
\end{lstlisting}
The agent checks its remembered pot contents before starting another batch. If the requested recipe is already cooking or ready, it can prepare a different recipe as stock rather than duplicate work that may become obsolete after delivery. It also tries to extend partially filled pots into valid recipes.

\begin{lstlisting}[style=generatedsnippet,firstline=213,lastline=225,firstnumber=213]
    # partner seen with a dish earlier and now empty-handed: it may have
    # delivered (resampling the recipe), so our recipe knowledge is stale
    p_has_dish = jnp.sum(obs[:, :, 16]) > 0
    p_empty = p_vis & (jnp.sum(obs[:, :, 15:20]) == 0)
    # delivery-approach zone: column 9 below the hub and row 17 near the goal
    p_near_goal = ((partner[1] == GOAL[1]) & (partner[0] >= 12)) | \
        ((partner[0] == GOAL[0] - 1) & (jnp.abs(partner[1] - GOAL[1]) <= 2))
    p_moved = jnp.any(partner != mem['ppos'])
    trig_age = jnp.minimum(mem['trig_age'] + 1, 999)
    p_dropped = ((mem['p_dish'] > 0) | (p_near_goal & p_moved)) & p_empty
    trig_age = jnp.where(p_dropped, 0, trig_age)
    rage = jnp.where(p_dropped & ~r_seen, jnp.maximum(rage, VERIFY_AGE + 1), rage)
    p_dish = jnp.where(p_vis, p_has_dish, mem['p_dish'] > 0)
\end{lstlisting}
The partner may deliver outside the agent's view and thereby change the recipe. Seeing the partner become empty-handed, or move near the delivery station while empty-handed, makes the policy treat its old recipe information as potentially stale. This is a heuristic cue to check again, not certainty that delivery occurred.

\clearpage
\subsection{Hanabi: Interpreting and Choosing Hints without Supplied Conventions}
\label{app:hanabi-standard-policy-examples}
\textbf{Claude Fable 5.1, standard two-player Hanabi, generation seed 1.}
\subsubsection{Reducing Trust in an Interpretation After Failed Plays}
\begin{lstlisting}[style=generatedsnippet,firstline=175,lastline=178,firstnumber=175]
    self_removed = actor_self & (is_play | is_discard)
    partner_removed = actor_partner & (is_play | is_discard)
    bombed_marked = actor_self & is_play & (~success) & (markers[removed] > 0)
    marked_bombs = marked_bombs + bombed_marked.astype(jnp.int32)
\end{lstlisting}
The policy remembers cards that a partner's hint made it regard as likely playable. If it plays one of these cards and loses a life, it records the failure in \texttt{marked\_bombs}.

\begin{lstlisting}[style=generatedsnippet,firstline=218,lastline=224,firstnumber=218]
    # ---------------- play evaluation ----------------
    mark_f = markers.astype(F32) / 100.0
    prot_f = protect.astype(F32)
    boost = jnp.where(marked_bombs >= 2, 0.0, jnp.where(marked_bombs >= 1, BOOST_AFTER_BOMB, BOOST))
    p_eff = p_play + boost * mark_f * (1.0 - p_play)
    p_eff = jnp.where(p_play > 0, p_eff, 0.0)
    thr = jnp.where(lives >= 3, THR_L3, jnp.where(lives >= 2, THR_L2, 0.999))
\end{lstlisting}
The additional confidence attached to such hints decreases after one failed marked play and disappears after two. The policy still uses the literal information in the hints. It also demands greater confidence when fewer lives remain. This lets experience weaken an unreliable assumption about what the partner intended.

\subsubsection{Choosing a Hint That Prevents a Costly Discard}
\begin{lstlisting}[style=generatedsnippet,firstline=278,lastline=289,firstnumber=278]
    # predicted partner discard before / after each hint
    d_before = _discard_pref(pb_useless, pb_crit, pb_play, p_hinted, pm_f, pp_f, p_occ, SLOT_IDX)
    chop_b = jnp.argmax(d_before)
    hinted_after = jnp.maximum(p_hinted[None, :], touched)
    prot_after = jnp.maximum(pp_f[None, :], (tch & (pa_play < MARK_MIN_P)).astype(F32))
    d_after = _discard_pref(pa_useless, pa_crit, pa_play, hinted_after, mark_strength_after, prot_after,
                            p_occ[None, :], SLOT_IDX[None, :])
    chop_a = jnp.argmax(d_after, axis=1)                                 # [10]
    crit_b = truly_critval[chop_b] * (p_occ[chop_b] > 0)
    crit_a = truly_critval[chop_a] * (p_occ[chop_a] > 0)
    save_gain = SAVE_WEIGHT * (crit_b - crit_a)
    hint_scores = hint_scores + save_gain
\end{lstlisting}
For each possible hint, the agent predicts which card the partner would discard before and after receiving it. It rewards hints that move the predicted discard away from an important card. In the code, \texttt{chop\_b} and \texttt{chop\_a} are the predicted discard positions, and \texttt{save\_gain} measures the reduction in loss. This is an explicit model of the hint's effect on the partner, although the partner may not follow that model exactly.

\clearpage
\subsection{Hanabi: Using H-group Level 1 Conventions}
\label{app:hanabi-hgroup-policy-examples}
\textbf{GPT-6 Astra, two-player Hanabi with H-group Level 1 instructions, generation seed 4.}
\subsubsection{Keeping Both Play and Save Interpretations of a Clue}
\begin{lstlisting}[style=generatedsnippet,firstline=99,lastline=108,firstnumber=99]
    connected = jnp.concatenate([jnp.ones((5, 1), jnp.bool_), jnp.cumprod(accounted[:, :4], axis=1).astype(jnp.bool_)], axis=1).reshape(25)
    play_set = connected & useful
    critical_save = (counts == 1) & (ranks != 4)
    special_save = rank_clue & ((value == 1) | (value == 4))
    save_set = useful & (critical_save | special_save) & (focus == chop)
    meaning = play_set | save_set
    new_recv = intersect_or(recv, literal_after[receiver], literal_after[receiver])
    good = new_recv & useful[None, :]
    new_recv = jnp.where(new[:, None] & jnp.any(good, -1, keepdims=True), good, new_recv)
    focused = new_recv[focus] & meaning
\end{lstlisting}
A clue can mean that a card should eventually be played, or that it should be saved from discard. The policy retains both possibilities. A save interpretation is considered when the clue focuses on the card normally discarded next, called the \emph{chop}. The variables use zero-based ranks, so values 1 and 4 denote twos and fives. The inferred meaning is combined with the clue's literal information.

\subsubsection{Recognising That Another Card Must Be Played First}
\begin{lstlisting}[style=generatedsnippet,firstline=109,lastline=126,firstnumber=109]
    # A Level 1 self-prompt needs an already-clued connector; never use an
    # unclued finesse position. Prefer ordinary interpretations when available.
    predecessor = jnp.maximum(jnp.arange(25) - 1, 0)
    compatible = (new_recv[:, predecessor] & touched[receiver, :, None]
                  & (jnp.arange(5) != focus)[:, None])
    connector = jnp.max(jnp.where(compatible, jnp.arange(5)[:, None], -1), axis=0)
    prompt_set = (new_recv[focus] & useful & (ranks == heights[colors] + 1)
                  & (connector >= 0) & ~other_have[predecessor])
    need_prompt = ~jnp.any(focused) & jnp.any(prompt_set)
    chosen = jnp.max(jnp.where(prompt_set, connector, -1))
    same_connector = jnp.all(~prompt_set | (connector == chosen))
    use_prompt = need_prompt & same_connector
    single_connector = jnp.sum(jnp.any(compatible & prompt_set[None, :], axis=-1)) == 1
    focused = jnp.where(need_prompt, prompt_set, focused)
    new_recv = new_recv.at[focus].set(jnp.where(jnp.any(focused), focused, new_recv[focus]))
    connector_meaning = jnp.concatenate([prompt_set.reshape(5, 5)[:, 1:], jnp.zeros((5, 1), jnp.bool_)], axis=1).reshape(25)
    connector_note = new_recv[jnp.maximum(chosen, 0)] & connector_meaning
    new_recv = new_recv.at[jnp.maximum(chosen, 0)].set(jnp.where(use_prompt, connector_note, new_recv[jnp.maximum(chosen, 0)]))
\end{lstlisting}
If the ordinary interpretations do not fit, the policy checks for a self-prompt: the clue may ask the receiver to play an already-clued prerequisite card before the newly indicated card. For example, an already-clued red two may need to precede a red three. The code restricts this interpretation to already-clued cards and records the inferred prerequisite. It does not invent a prerequisite in an untouched card.

\clearpage
\subsubsection{Revising an Interpretation When Its Support Disappears}
\begin{lstlisting}[style=generatedsnippet,firstline=234,lastline=246,firstnumber=234]
    predecessor = jnp.maximum(jnp.arange(25) - 1, 0)
    still_possible = jnp.transpose(notes[:, :, predecessor], (0, 2, 1))
    wait = wait & still_possible[:, None, :, :]
    supported_by_board = jnp.tile(jnp.arange(5), 5) <= jnp.repeat(heights, 5)
    broken_delay = was_waiting & ~jnp.any(wait, -1) & ~supported_by_board[None, None, :]
    revised = notes & ~broken_delay
    notes = jnp.where(jnp.any(revised, -1, keepdims=True), revised, literal)
    wait = wait & ~supported_by_board[None, None, :, None]
    # A model of the partner's belief is uncertain; never treat a demonstrably
    # wrong predicted identity as public card evidence.
    partner_consistent = jnp.any(notes[1] & (hand > 0), axis=-1)
    notes = notes.at[1].set(jnp.where(partner_consistent[:, None], notes[1], literal[1]))
    wait = wait.at[1].set(wait[1] & partner_consistent[:, None, None])
\end{lstlisting}
The policy tracks which cards support a delayed-play interpretation. If those possibilities disappear without the required progress on the board, it removes the unsupported interpretation. It also checks its model of the partner's beliefs against the partner's visible hand. If they contradict one another, it falls back to literal clue information instead of continuing to rely on the mistaken convention inference.

\clearpage
\subsection{Yokai: Learning from a Partner's Card Movements}
\label{app:yokai-policy-examples}
\textbf{GPT-6 Astra, sixteen cards with perfect memory, generation seed 3.}
\subsubsection{Treating a Move as Tentative Evidence About Hidden Colours}
\begin{lstlisting}[style=generatedsnippet,firstline=151,lastline=158,firstnumber=151]
    moved = jnp.any(obs.positions != memory['positions'], axis=1) & ~reset
    adj = jnp.sum(jnp.abs(obs.positions[:, None] - obs.positions[None, :]), axis=-1) == 1
    oldadj = jnp.sum(jnp.abs(memory['positions'][:, None] - memory['positions'][None, :]), axis=-1) == 1
    partner_seen = features[:16, 11] == 1
    suggested = moved[:, None] & adj & ~oldadj & partner_seen[:, None] & partner_seen[None, :] & (features[0, 12] == 0)
    evidence = .9 * suggested / jnp.maximum(jnp.sum(suggested, axis=1, keepdims=True), 1)
    relations = jnp.maximum(memory['relations'] * adj, evidence + evidence.T)
    planning = motion_beliefs(p, relations)
\end{lstlisting}
When the partner places two previously inspected cards next to one another, the policy treats this as evidence that their colours may match. It knows which cards the partner inspected, but not what colours the partner saw. This tentative evidence affects planning without overwriting colours learned by direct observation.

\begin{lstlisting}[style=generatedsnippet,firstline=164,lastline=168,firstnumber=164]
    # Gather fresh evidence, particularly about recent partner relocations.
    observe_score = uncertainty * (10. + 2. * moved + .2 * (1 - features[:16, 11]))
    observe_score += .8 * info_value * (~certain)
    observe_score += .01 * jnp.sum(obs.adjacency_matrix[:16, :16], axis=1)
    observe_action = 1 + jnp.argmax(jnp.where(obs.legal_actions[1:17], observe_score, -1e6))
\end{lstlisting}
Uncertain cards moved by the partner receive extra priority for inspection. The agent therefore uses the partner's action both as a clue and as a reason to gather more direct evidence. It also values observations that could improve its next move.

\subsubsection{Sharing Useful Information and Checking the Stopping Condition}
\begin{lstlisting}[style=generatedsnippet,firstline=191,lastline=205,firstnumber=191]
    reveal = obs.legal_actions[1617:1624]
    hint_size = jnp.sum(hintset, axis=1)[:, None]
    precise = (hint_size == 1) & (settled[None, :] >= 1.)
    useful_pair = (hint_size == 2) & (settled[None, :] >= 2.) & (features[:16, 11][None, :] == 0) & (split_anchor[None, :] == 0)
    inform = (precise | useful_pair) & (match > .999) & (need_move[None, :] < 6.)
    inform_scores = jnp.full((7, 100), -1e6).at[:, at].set(jnp.where(inform, hint_card_score, -1e6))
    inform_scores = jnp.where(obs.legal_actions[1624:2324].reshape(7, 100), inform_scores, -1e6)
    can_inform = jnp.max(inform_scores) > -1e5
    hint_action = jnp.where(can_inform, 1624 + jnp.argmax(inform_scores), jnp.where(jnp.any(reveal), 1617 + jnp.argmax(reveal), place_action))
    # A connected graph of each known color certifies success when all are known.
    solved = jnp.all(certain) & jnp.all((same < .999) | reach)
    substep = features[0, 13]
    action = jnp.where(substep < 2, observe_action, jnp.where(substep == 2, move_action, hint_action))
    action = jnp.where((substep == 0) & solved, 2324, action)
    action = jnp.where(features[0, 12] == 1, action, 0).astype(jnp.int32)
\end{lstlisting}
The agent prefers informative hints on settled groups, including two-colour hints for cards the partner has not inspected. It ends the game only when every colour is certain and each colour forms a connected group. This check uses evidence-based beliefs, excluding the speculative colour inferences from partner movements.

\clearpage
\section{Policy Complexity and Refinement Effort}
\label{app:policy-complexity}
Beyond evaluation scores, we quantify the complexity of the generated policies using adaptations of \ac{cc} \citep{mccabe1976complexity} and \ac{hv} \citep{halstead1977elements}. We compute these metrics on JAXPR, JAX's intermediate representation, using JAX's native tracing functionality to capture policy operations and control flow expressed through JAX primitives. Classical \acs{cc} measures the number of linearly independent paths through a program's control flow. Our JAXPR adaptation also counts conditional selections such as \texttt{where}, which select between computed values rather than branching between execution paths. \Ac{hv} combines the total number of operator and operand occurrences with the number of distinct operators and operands to estimate a program's information content in bits. We apply this measure to JAXPR primitive operations and their operands after dead-code elimination.

We also report the number of completed official evaluations, which counts how many times the model receives feedback from the official evaluator during policy generation. Models may additionally implement custom rollout functions and use their results for refinement; these unofficial evaluations are excluded from the reported counts.

% Exact values and bootstrap provenance: policy_complexity_sources.json.
\begin{table}[!htbp]
    \centering
    \caption{\textbf{Policy complexity and refinement effort.} Each cell reports the mean across 11 environments, with a 95\% confidence interval.}
    \label{tab:policy-complexity}
    \begingroup
    \small
    \setlength{\tabcolsep}{1pt}
    \renewcommand{\arraystretch}{1.1}
    \newcommand{\pcvalue}[3]{\shortstack[c]{{\footnotesize #1}\\[-1pt]{\fontsize{6}{7}\selectfont\color{black!70}[#2,\,#3]}}}
    \begin{tabular*}{\linewidth}{@{\extracolsep{\fill}}lccccccc@{}}
    \toprule
    Model & \shortstack[c]{{\acs{sp}}\\{rank \ensuremath{(\downarrow)}}} & \shortstack[c]{{\acs{ixp}}\\{rank \ensuremath{(\downarrow)}}} & \shortstack[c]{{Inter-model}\\{\acs{xp} rank \ensuremath{(\downarrow)}}} & \acs{cc} & \acs{hv} & \acs{loc} & \shortstack{Official\\evaluations} \\
    \midrule
    GPT-5.6 Luna & \pcvalue{6.9}{6.7}{7.0} & \pcvalue{6.8}{6.6}{7.0} & \pcvalue{7.0}{6.7}{7.0} & \pcvalue{51.2}{43.1}{59.4} & \pcvalue{9,548}{7,698}{11,385} & \pcvalue{132.6}{121.5}{143.5} & \pcvalue{8.0}{7.1}{8.8} \\[1pt]
    Claude Sonnet 5 & \pcvalue{5.7}{5.4}{6.1} & \pcvalue{5.5}{5.0}{5.6} & \pcvalue{5.5}{5.2}{5.6} & \pcvalue{52.4}{44.9}{61.7} & \pcvalue{9,580}{7,921}{11,588} & \pcvalue{184.1}{174.6}{193.9} & \pcvalue{14.0}{12.0}{16.4} \\[1pt]
    GPT-5.5 & \pcvalue{5.3}{4.7}{5.4} & \pcvalue{5.5}{4.9}{5.7} & \pcvalue{5.1}{4.9}{5.5} & \pcvalue{96.2}{69.8}{134.7} & \pcvalue{30,951}{15,284}{52,844} & \pcvalue{201.3}{191.0}{212.9} & \pcvalue{5.8}{5.4}{6.2} \\[1pt]
    GPT-5.6 Sol & \pcvalue{2.5}{2.3}{3.0} & \pcvalue{3.5}{2.8}{4.0} & \pcvalue{3.3}{2.9}{3.7} & \pcvalue{118.4}{104.1}{137.3} & \pcvalue{36,866}{29,945}{45,678} & \pcvalue{264.8}{250.5}{279.6} & \pcvalue{24.6}{22.4}{26.9} \\[1pt]
    Claude Opus 5 & \pcvalue{3.2}{2.8}{3.5} & \pcvalue{3.1}{2.7}{3.5} & \pcvalue{3.2}{2.6}{3.6} & \pcvalue{96.0}{85.8}{107.4} & \pcvalue{20,914}{18,531}{23,779} & \pcvalue{297.9}{287.4}{308.7} & \pcvalue{11.2}{8.6}{14.3} \\[1pt]
    GPT-6 Astra & \pcvalue{2.5}{2.1}{2.8} & \pcvalue{1.9}{1.6}{2.4} & \pcvalue{2.1}{1.7}{2.5} & \pcvalue{134.5}{120.1}{148.8} & \pcvalue{31,329}{26,993}{35,856} & \pcvalue{181.9}{171.3}{192.5} & \pcvalue{13.0}{11.4}{14.5} \\[1pt]
    Claude Fable 5.1 & \pcvalue{1.8}{1.6}{2.4} & \pcvalue{1.7}{1.5}{2.4} & \pcvalue{1.9}{1.5}{2.3} & \pcvalue{159.7}{137.4}{185.2} & \pcvalue{34,678}{30,147}{39,857} & \pcvalue{316.2}{302.7}{329.7} & \pcvalue{7.8}{7.1}{8.6} \\[1pt]
    \bottomrule
    \end{tabular*}
    \endgroup
\end{table}

Table~\ref{tab:policy-complexity} suggests that stronger coordination performance is generally associated with greater policy complexity, but not necessarily longer code or more refinement. Astra and Fable achieve the best mean \acs{ixp} ranks and generate policies with the highest mean \ac{cc}. However, Astra's policies are substantially shorter than those of Sol and Opus, showing that code length alone does not capture structural complexity. \Ac{hv} also does not consistently track performance: GPT-5.5 produces policies with a volume comparable to Astra's despite substantially weaker coordination. Refinement effort varies considerably across models. Fable achieves the best mean \acs{ixp} rank with 7.8 official evaluations on average, compared with Sol's 24.6, suggesting that stronger models are more sample-efficient.

\clearpage
\section{Learning Curves}

\begin{figure}[!ht]
    \input{Figures/learning_curves}
\end{figure}

\clearpage
\section{Examples of Experiment-Ledger Entries}
\label{app:experiment-ledgers}

\begin{figure}[!ht]
    \input{Figures/experiment_ledger_examples}
\end{figure}

\clearpage
\section{Full Generated Policy Listings}
\label{app:full-generated-policies}
We reproduce three complete, unchanged evaluated policy files. Focused excerpts and explanations appear in Appendix~\ref{app:generated-policies}. Source line numbers match those excerpts.

\subsection{GPT-6 Astra on Overcooked V2: Grounded Coordination (Simple)}
\label{app:policy-grounded-coord-simple}
\noindent Adapts recipe-signal timing and uses displayed or inferred recipes to guide cooking.

\begin{lstlisting}[style=generatedpolicy]
"""Grounded recipe communication and local, observation-driven cooking."""

import jax.numpy as jnp

POLICY_API_VERSION = 1
POLICY_METADATA = {'name': 'grounded_cook', 'adapter': 'overcooked_v2', 'adapter_version': 1, 'max_height': 5, 'max_width': 8, 'num_ingredients': 3}


def locate(obs, channel, global_row, global_col):
    index = jnp.argmax(obs[:, :, channel].reshape(-1))
    return global_row + 2 - index // 5, global_col + 2 - index % 5


def approach(row, col, facing, target_row, target_col, direction):
    move = jnp.where(row < target_row, 1, jnp.where(row > target_row, 3,
           jnp.where(col < target_col, 0, jnp.where(col > target_col, 2, 4))))
    turn = jnp.asarray([3, 1, 0, 2], jnp.int32)[direction]
    return jnp.where((row == target_row) & (col == target_col),
                     jnp.where(facing == direction, 5, turn), move).astype(jnp.int32)


def init_agent_0():
    return jnp.asarray([0, 0, 100, 0, 0, -1, 0, 6, 0], jnp.int32)


def agent_0_act(obs, reset, memory, rng):
    memory = jnp.where(reset, init_agent_0(), memory)
    time, old_count, since_signal, empty_age, retries, old_recipe, last_fill, fill_gap, warned = memory
    row, col = locate(obs, 23, 2, 0)
    facing = jnp.argmax(obs[2, 2, 1:5])
    pot = jnp.sum(obs[:, :, 29:34] * obs[:, :, 22:23], axis=(0, 1))
    count = jnp.sum(pot[2:])
    cleared = (old_count > 0) & (count == 0) & (col == 2)
    inserted = (count > old_count) & (col == 2)
    fill_gap = jnp.where(inserted & (old_count > 0), jnp.clip(time - last_fill, 6, 24), fill_gap)
    last_fill = jnp.where(inserted, time, last_fill)
    warned = jnp.where(cleared, 0, warned)
    empty_age = jnp.where(cleared, 0, empty_age + 1)
    since_signal = jnp.where(cleared, 100, since_signal + 1)
    retries = jnp.where(cleared, 0, retries)
    recipe = jnp.argmax(jnp.sum(obs[:, :, 36:39] * obs[:, :, 23:24], axis=(0, 1)))
    changed = (old_recipe >= 0) & (recipe != old_recipe)
    display_active = jnp.sum(obs[:, :, 36:39] * obs[:, :, 24:25]) > 0
    retry_base = jnp.maximum(20, (3 * fill_gap) // 2)
    retry_interval = retry_base * jnp.left_shift(jnp.asarray(1, jnp.int32), jnp.minimum(jnp.maximum(retries - 1, 0), 3))
    signal_delay = jnp.maximum(3, (2 * fill_gap) // 3 - 5)
    next_cycle_steps = 20 + (11 * fill_gap) // 3
    cycle_signal = (count == 0) & (((empty_age >= signal_delay) & (since_signal >= retry_interval)) | changed)
    correction = (count > 0) & (jnp.argmax(pot[2:]) != recipe) & (warned == 0) & (since_signal >= 10)
    ready = (cycle_signal | correction) & ~display_active & (time + next_cycle_steps < 400)
    action = approach(row, col, facing, 2, 2, 2)
    at_button = (row == 2) & (col == 2) & (facing == 2)
    action = jnp.where(at_button & ~ready, 4, action).astype(jnp.int32)
    activated = at_button & ready
    since_signal = jnp.where(activated, 0, since_signal)
    warned = jnp.where(activated & correction, 1, warned)
    retries = retries + activated.astype(jnp.int32)
    memory = jnp.stack([time + 1, jnp.where(col == 2, count, old_count), since_signal, empty_age, retries, recipe, last_fill, fill_gap, warned]).astype(jnp.int32)
    return action, memory, {'pot_count': count, 'activated': activated, 'empty_age': empty_age, 'fill_gap': fill_gap}


def counter_options(obs, row, col, facing, recipe, source):
    # All four usable counters are directly observable from the middle row.
    counter_rows = jnp.asarray([0, 0, 4, 4], jnp.int32)
    counter_cols = jnp.asarray([5, 6, 5, 6], jnp.int32)
    local_rows, local_cols = counter_rows - row + 2, counter_cols - col + 2
    seen = (local_rows >= 0) & (local_rows < 5) & (local_cols >= 0) & (local_cols < 5)
    items = obs[jnp.clip(local_rows, 0, 4), jnp.clip(local_cols, 0, 4), 29:34]
    standing_rows = jnp.where(counter_rows == 0, 1, 3)
    directions = jnp.where(counter_rows == 0, 0, 1)
    distance = jnp.abs(row - standing_rows) + jnp.abs(col - counter_cols)
    free = seen & (jnp.sum(items, axis=1) == 0)
    soup = seen & (items[:, 1] > 0) & (items[:, 2 + recipe] == 3)
    raw = seen & (items[:, 0] == 0) & (items[:, 1] == 0) & (items[:, 2 + source] == 1)
    def select(mask):
        index = jnp.argmin(jnp.where(mask, distance, 100))
        action = approach(row, col, facing, standing_rows[index], counter_cols[index], directions[index])
        return jnp.any(mask), action
    return select(free), select(soup), select(raw)


def init_agent_1():
    # time, recipe, confidence (0 unknown / 1 displayed / 2 inferred),
    # previous goal interaction, wait count, ever informed, silent cycles
    return jnp.asarray([0, 0, 0, 0, 0, 0, 0], jnp.int32)


def agent_1_act(obs, reset, memory, rng):
    memory = jnp.where(reset, init_agent_1(), memory)
    time, recipe, valid, old_goal, waiting, informed, silent_cycles = memory
    row, col = locate(obs, 22, 2, 4)
    facing = jnp.argmax(obs[2, 2, 1:5])
    inv = obs[2, 2, 5:10]
    dish, plate = inv[1] > 0, inv[0] > 0
    raw = (jnp.sum(inv[2:]) > 0) & ~dish
    empty = jnp.sum(inv) == 0
    delivered = (old_goal > 0) & ~dish
    valid = jnp.where(delivered, 0, valid)
    waiting = jnp.where(delivered, 0, waiting)

    displayed = jnp.sum(obs[:, :, 36:39], axis=(0, 1))
    visible = jnp.sum(displayed) > 0
    recipe = jnp.where(visible, jnp.argmax(displayed), recipe)
    valid = jnp.where(visible, 1, valid)
    informed = jnp.maximum(informed, visible.astype(jnp.int32))
    silent_cycles = jnp.where(visible, 0, silent_cycles)
    waiting = jnp.where((valid == 0) & empty, waiting + 1, waiting)
    wait_limit = jnp.where(time < 40, 20, jnp.where((informed == 0) | (silent_cycles >= 3), 4, 10))
    fallback = (valid == 0) & (waiting >= wait_limit)
    silent_cycles = silent_cycles + fallback.astype(jnp.int32)
    recipe = jnp.where(fallback & ((informed == 0) | (silent_cycles >= 3)), 1 - recipe, recipe)
    valid = jnp.where(fallback, 2, valid)

    pot = jnp.sum(obs[:, :, 29:34] * obs[:, :, 22:23], axis=(0, 1))
    count = jnp.sum(pot[2:])
    cooked = pot[1] > 0
    # A partly filled pot cannot be undone: finish it homogeneously, then
    # store it if a later grounded display says the dish would be wrong.
    source = jnp.where(count > 0, jnp.argmax(pot[2:]), recipe)
    source_row = jnp.where(source == 0, 1, 3)
    get_raw = approach(row, col, facing, source_row, 6, 2)
    pot_action = approach(row, col, facing, 2, 5, 3)
    get_plate = approach(row, col, facing, jnp.where(row == 3, 3, 1), 5, 3)
    deliver = approach(row, col, facing, 2, 6, 2)
    wait_display = jnp.where(pot_action == 5, 4, pot_action)

    (free_counter, store_action), (stored_soup, soup_action), (stored_raw, raw_action) = counter_options(obs, row, col, facing, recipe, source)
    get_raw = jnp.where(stored_raw, raw_action, get_raw)
    acquire = jnp.where((count == 0) & stored_soup, soup_action, get_raw)
    mismatched = jnp.argmax(inv[2:]) != recipe
    park_dish = dish & mismatched & (valid == 1) & free_counter
    park_raw = raw & mismatched & (valid == 1) & (count == 0) & free_counter

    action = jnp.where(dish, jnp.where(park_dish, store_action, deliver),
             jnp.where(plate, jnp.where(cooked, pot_action, jnp.where(pot_action == 5, 4, pot_action)),
             jnp.where(raw, jnp.where(park_raw, store_action, pot_action),
             jnp.where(count >= 3, get_plate,
             jnp.where((valid > 0) | (count > 0), acquire, wait_display)))))
    goal_attempt = dish & (row == 2) & (col == 6) & (facing == 2) & (action == 5)
    memory = jnp.stack([time + 1, recipe, valid, goal_attempt.astype(jnp.int32), waiting, informed, silent_cycles]).astype(jnp.int32)
    trace = {'recipe': recipe, 'valid': valid, 'pot_count': count,
             'waiting': waiting, 'delivered': delivered, 'row': row, 'col': col,
             'silent_cycles': silent_cycles, 'park_dish': park_dish,
             'park_raw': park_raw, 'stored_soup': stored_soup}
    return action.astype(jnp.int32), memory, trace
\end{lstlisting}

\clearpage
\subsection{GPT-6 Astra on Yokai: Sixteen Cards, Perfect Memory}
\label{app:policy-yokai-2p-16c}
\noindent Updates beliefs from partner relocations to guide card observations and moves.

\begin{lstlisting}[style=generatedpolicy]
"""Local evidence and geometric color grouping for cooperative Yokai."""

import jax
import jax.numpy as jnp

POLICY_API_VERSION = 1
POLICY_METADATA = {'name': 'bare_candidate', 'adapter': 'yokai', 'adapter_version': 1, 'num_agents': 2, 'num_colors': 4, 'deck_size': 16, 'grid_size': 10, 'num_hints': 7, 'num_nodes': 23, 'node_features': 15, 'num_actions': 2325, 'policy_structure': 'shared_local'}


def init_agent():
    return {"colors": jnp.zeros((16, 4), jnp.float32),
            "positions": jnp.zeros((16, 2), jnp.int32),
            "relations": jnp.zeros((16, 16), jnp.float32),
            "future": jnp.zeros((16,), jnp.bool_)}


def beliefs(obs, seen):
    hints = obs.node_features[16:, 4:8].astype(jnp.float32)
    attached = obs.adjacency_matrix[:16, 16:].astype(jnp.float32)
    support = attached @ hints
    allowed = jnp.where(jnp.sum(support, axis=1, keepdims=True) > 0, support, 1.)
    known = jnp.sum(seen, axis=1, keepdims=True) > 0
    domains = jnp.where(known, seen > 0, allowed > 0)
    def force(_, d):
        singleton = jnp.sum(d, axis=1, keepdims=True) == 1
        fixed_count = jnp.sum(d & singleton, axis=0, keepdims=True)
        possible_count = jnp.sum(d, axis=0, keepdims=True)
        d = d & ((fixed_count < 4) | singleton)
        forced = d & (possible_count == 4)
        d = jnp.where(jnp.any(forced, axis=1, keepdims=True), forced, d)
        return d
    domains = jax.lax.fori_loop(0, 16, force, domains)
    # Fall back to own observations if an independently supplied hint conflicts.
    domains = jnp.where(jnp.any(domains, axis=1, keepdims=True), domains, jnp.where(known, seen > 0, True))
    known = jnp.sum(domains, axis=1, keepdims=True) == 1
    seen = domains.astype(jnp.float32)
    p = seen / jnp.sum(seen, axis=1, keepdims=True)
    def balance(_, p):
        remain = jnp.maximum(4. - jnp.sum(jnp.where(known, seen, 0.), axis=0), 0.)
        u = jnp.where(known, 0., p)
        u = u * remain / jnp.maximum(jnp.sum(u, axis=0), 1e-6)
        u = u / jnp.maximum(jnp.sum(u, axis=1, keepdims=True), 1e-6)
        return jnp.where(known, seen, u)
    return jax.lax.fori_loop(0, 8, balance, p)


def removed_components(edge):
    ids = jnp.arange(16)
    present = ids[None, :] != ids[:, None]
    edges = edge[None] & present[:, :, None] & present[:, None, :]
    labels = jnp.broadcast_to(ids, (16, 16))
    def propagate(_, lab):
        return jnp.minimum(lab, jnp.min(jnp.where(edges, lab[:, None, :], 16), axis=-1))
    labels = jax.lax.fori_loop(0, 15, propagate, labels)
    reach = (labels[:, :, None] == labels[:, None, :]) & present[:, :, None] & present[:, None, :]
    reps = (labels == ids[None, :]) & present
    return reach, reps


def motion_beliefs(p, relations):
    fixed = jnp.max(p, axis=1, keepdims=True) > .999
    remaining = jnp.maximum(4. - jnp.sum(jnp.where(fixed, p, 0.), axis=0), 0.)
    def update(_, q):
        q = p * jnp.exp(3. * (relations @ q))
        q = q / jnp.maximum(jnp.sum(q, axis=1, keepdims=True), 1e-6)
        u = jnp.where(fixed, 0., q)
        u = u * remaining / jnp.maximum(jnp.sum(u, axis=0), 1e-6)
        u = u / jnp.maximum(jnp.sum(u, axis=1, keepdims=True), 1e-6)
        return jnp.where(fixed, p, u)
    return jax.lax.fori_loop(0, 8, update, p)


def geometry(obs, p, relations, positions=None):
    ids = jnp.arange(16)
    present = ids[None, :] != ids[:, None]
    pos = obs.positions if positions is None else positions
    dist = jnp.sum(jnp.abs(pos[:, None] - pos[None, :]), axis=-1)
    certain = jnp.max(p, axis=1) > .999
    color = jnp.argmax(p, axis=1)
    edge = (dist == 1) & (color[:, None] == color[None, :]) & certain[:, None] & certain[None, :]
    # Compute the color components after removing each possible moving card.
    reach, reps = removed_components(edge)
    cells = jnp.stack((jnp.arange(100) // 10, jnp.arange(100) % 10), axis=1)
    cd = jnp.sum(jnp.abs(cells[:, None, :] - pos[None, :, :]), axis=-1)
    near = (cd == 1).astype(jnp.float32)
    touched = jnp.einsum('dj,ijk->idk', near, reach.astype(jnp.float32)) > 0
    oldtouched = jnp.einsum('ij,ijk->ik', (dist == 1).astype(jnp.float32), reach.astype(jnp.float32)) > 0
    same = p @ p.T
    confidence = jnp.where(certain, 1., .12 + .6 * jnp.maximum((jnp.max(p, axis=1) - .25) / .75, 0.))
    weights = same * confidence[:, None] * confidence[None, :]
    weights = jnp.where(certain[:, None] & certain[None, :], weights, jnp.maximum(weights, relations))
    comp_gain = jnp.sum((touched.astype(jnp.float32) - oldtouched[:, None, :]) * reps[:, None, :] * weights[:, None, :], axis=-1)
    edge_gain = jnp.einsum('dj,ij->id', near, weights * present) - jnp.sum((dist == 1) * weights, axis=1)[:, None]
    distance_gain = jnp.sum((dist[:, None, :] - cd[None]) * weights[:, None, :] * present[:, None, :], axis=-1) / jnp.maximum(jnp.sum(weights * present, axis=-1)[:, None], 1.)
    degree = jnp.sum(near, axis=1)[None, :] - near.T
    scores = 12. * comp_gain + .65 * edge_gain + .7 * distance_gain + .025 * degree
    fullreach, fullreps = removed_components(dist == 1)
    touch_count = jnp.einsum('dj,ijk->idk', near, fullreach.astype(jnp.float32))
    fulltouch = touch_count > 0
    vacant = ~jnp.any(cd == 0, axis=1)
    legal = vacant[None, :] & jnp.all(fulltouch | ~fullreps[:, None, :], axis=-1) & (obs.node_features[:16, 10:11] == 0)
    unrestricted = jnp.max(jnp.where(vacant[None, :], scores, -1e6), axis=1)
    constrained = jnp.max(jnp.where(legal, scores, -1e6), axis=1)
    blocked = (jnp.sum(fullreps, axis=1) > 1) & certain & (obs.node_features[:16, 10] == 0) & (constrained < 6.) & (unrestricted > 6.)
    bridge_all = jnp.all(fulltouch | ~fullreps[:, None, :], axis=-1)
    comp_size = jnp.sum(fullreach, axis=1)
    loses_contact = (near[None, :, :] > 0) & (touch_count == 1) & (comp_size[:, None, :] > 1)
    bridge = bridge_all[:, :, None] & ~loses_contact & present[:, None, :]
    bridge_value = jnp.max(bridge * (blocked * jnp.minimum(unrestricted - constrained, 12.))[:, None, None], axis=0).T
    raw_scores = jnp.where(legal, scores, -1e6)
    scores = jnp.where(legal, scores + .35 * bridge_value, -1e6)
    # Expected value of learning the moving card's color before choosing its destination.
    target = p * confidence[:, None]
    cg = jnp.einsum('idk,kc->idc', (touched.astype(jnp.float32) - oldtouched[:, None, :]) * reps[:, None, :], target)
    eg = jnp.einsum('idk,kc->idc', (near[None] - (dist == 1)[:, None, :]) * present[:, None, :], target)
    dg = jnp.einsum('idk,kc->idc', (dist[:, None, :] - cd[None]) * present[:, None, :], target)
    dg /= jnp.maximum(jnp.einsum('ik,kc->ic', present, target)[:, None, :], 1.)
    conditional = 12. * cg + .65 * eg + .7 * dg + .025 * degree[:, :, None]
    conditional = jnp.where(legal[:, :, None], conditional, -1e6)
    best_now = jnp.maximum(jnp.max(scores), 0.)
    info_value = jnp.sum(p * jnp.maximum(jnp.max(conditional, axis=1) - best_now, 0.), axis=1)
    return scores, edge, same, certain, info_value, raw_scores


def prepare_move(obs, p, relations, scores, raw_scores):
    _, first = jax.lax.top_k(scores.reshape(-1), 8)
    reveals = jnp.sum(jnp.any(obs.node_features[16:, 4:8] > 0, axis=1))
    placements = jnp.sum(obs.node_features[16:, 10])
    elapsed_turns = reveals + placements
    def follow(action):
        card, cell = action // 100, action % 100
        pos = obs.positions.at[card].set(jnp.stack((cell // 10, cell % 10)))
        adj = jnp.sum(jnp.abs(pos[:, None] - pos[None, :]), axis=-1) == 1
        future = geometry(obs, p, relations * adj, pos)[5]
        future = jnp.where((obs.node_features[:16, 11:12] == 1) | (elapsed_turns < 12), future, -1e6)
        best = jnp.argmax(future)
        return jnp.max(future), best // 100
    values, movers = jax.vmap(follow)(first)
    total = raw_scores.reshape(-1)[first] + .9 * values
    choice = jnp.argmax(total)
    useful = (elapsed_turns < 13) & (jnp.max(scores) < 8.) & (jnp.max(total) > jnp.max(raw_scores) + 4.) & (values[choice] > 6.)
    return first[choice], movers[choice], useful


def agent_act(obs, reset, memory, rng):
    memory = jax.tree.map(lambda x: jnp.where(reset, jnp.zeros_like(x), x), memory)
    features = obs.node_features
    visible = features[:16, :4].astype(jnp.float32)
    seen = jnp.where(jnp.sum(visible, axis=1, keepdims=True) > 0, visible, memory['colors'])
    p = beliefs(obs, seen)
    moved = jnp.any(obs.positions != memory['positions'], axis=1) & ~reset
    adj = jnp.sum(jnp.abs(obs.positions[:, None] - obs.positions[None, :]), axis=-1) == 1
    oldadj = jnp.sum(jnp.abs(memory['positions'][:, None] - memory['positions'][None, :]), axis=-1) == 1
    partner_seen = features[:16, 11] == 1
    suggested = moved[:, None] & adj & ~oldadj & partner_seen[:, None] & partner_seen[None, :] & (features[0, 12] == 0)
    evidence = .9 * suggested / jnp.maximum(jnp.sum(suggested, axis=1, keepdims=True), 1)
    relations = jnp.maximum(memory['relations'] * adj, evidence + evidence.T)
    planning = motion_beliefs(p, relations)
    scores, _, _, _, info_value, raw_scores = geometry(obs, planning, relations)
    certain = jnp.max(p, axis=1) > .999
    same = p @ p.T
    edge = adj & (same > .999) & certain[:, None] & certain[None, :]
    uncertainty = 1. - jnp.max(p, axis=1)
    # Gather fresh evidence, particularly about recent partner relocations.
    observe_score = uncertainty * (10. + 2. * moved + .2 * (1 - features[:16, 11]))
    observe_score += .8 * info_value * (~certain)
    observe_score += .01 * jnp.sum(obs.adjacency_matrix[:16, :16], axis=1)
    observe_action = 1 + jnp.argmax(jnp.where(obs.legal_actions[1:17], observe_score, -1e6))
    move_action = 17 + jnp.argmax(jnp.where(obs.legal_actions[17:1617].reshape(16, 100), scores, -1e6))
    planned_action, planned_mover, useful_plan = prepare_move(obs, planning, relations, scores, raw_scores)
    use_plan = useful_plan & (jnp.sum(certain) >= 12) & (features[0, 12] == 1) & (features[0, 13] == 2)
    move_action = jnp.where(use_plan, 17 + planned_action, move_action)
    future = jnp.where((features[0, 12] == 1) & (features[0, 13] == 2), (jnp.arange(16) == planned_mover) & use_plan, memory['future'])
    hintset = features[16:, 4:8].astype(jnp.float32)
    match = hintset @ p.T
    reach = edge | jnp.eye(16, dtype=jnp.bool_)
    def closure(_, r):
        return (r.astype(jnp.int32) @ r.astype(jnp.int32)) > 0
    reach = jax.lax.fori_loop(0, 4, closure, reach)
    settled = jnp.sum(edge, axis=1).astype(jnp.float32)
    need_move = jnp.maximum(jnp.max(scores, axis=1), 0.)
    split_anchor = jnp.sum((same > .999) & ~reach & (features[:16, 10] == 1)[None, :], axis=1)
    hint_card_score = 100. * match + 3. * settled[None, :] - .5 * need_move[None, :] - 30. * split_anchor[None, :] - 30. * future[None, :]
    # Favor precise truthful public evidence when choosing among available hints.
    hint_card_score += (4. - jnp.sum(hintset, axis=1))[:, None] * 7.
    hint_card_score += (1. - features[:16, 11])[None, :] * (4. - jnp.sum(hintset, axis=1))[:, None]
    at = obs.positions[:, 0] * 10 + obs.positions[:, 1]
    place_scores = jnp.full((7, 100), -1e6).at[:, at].set(hint_card_score)
    place_scores = jnp.where(obs.legal_actions[1624:2324].reshape(7, 100), place_scores, -1e6)
    place_action = 1624 + jnp.argmax(place_scores)
    reveal = obs.legal_actions[1617:1624]
    hint_size = jnp.sum(hintset, axis=1)[:, None]
    precise = (hint_size == 1) & (settled[None, :] >= 1.)
    useful_pair = (hint_size == 2) & (settled[None, :] >= 2.) & (features[:16, 11][None, :] == 0) & (split_anchor[None, :] == 0)
    inform = (precise | useful_pair) & (match > .999) & (need_move[None, :] < 6.)
    inform_scores = jnp.full((7, 100), -1e6).at[:, at].set(jnp.where(inform, hint_card_score, -1e6))
    inform_scores = jnp.where(obs.legal_actions[1624:2324].reshape(7, 100), inform_scores, -1e6)
    can_inform = jnp.max(inform_scores) > -1e5
    hint_action = jnp.where(can_inform, 1624 + jnp.argmax(inform_scores), jnp.where(jnp.any(reveal), 1617 + jnp.argmax(reveal), place_action))
    # A connected graph of each known color certifies success when all are known.
    solved = jnp.all(certain) & jnp.all((same < .999) | reach)
    substep = features[0, 13]
    action = jnp.where(substep < 2, observe_action, jnp.where(substep == 2, move_action, hint_action))
    action = jnp.where((substep == 0) & solved, 2324, action)
    action = jnp.where(features[0, 12] == 1, action, 0).astype(jnp.int32)
    future = jnp.where((features[0, 12] == 1) & (features[0, 13] == 3), False, future)
    memory = {'colors': seen, 'positions': obs.positions, 'relations': relations, 'future': future}
    return action, memory, {'known': jnp.sum(certain), 'solved': solved, 'move_value': jnp.max(scores), 'belief': p, 'planning': planning, 'positions': obs.positions, 'locked': features[:16, 10]}
\end{lstlisting}

\clearpage
\subsection{GPT-6 Astra on Hanabi: H-group Level 1}
\label{app:policy-hanabi-2p-h-group-level-1}
\noindent Interprets clues using play, save, and self-prompt alternatives.

\lstinputlisting[style=generatedpolicy]{Policies/hanabi_2p_h_group_level_1/policy.py}

\end{document}